# GuavaNet: A deep neural network architecture for automatic sensory evaluation to predict degree of acceptability for Guava by a consumer

Vipul Mehra
vm8176@rit.edu

# GuavaNet: A deep neural network architecture for automatic sensory evaluation to predict degree of acceptability for Guava by a consumer

by

**Vipul Mehra**

A Thesis Submitted
in
Partial Fulfillment of the
Requirements for the Degree of
Master of Science in Information Sciences and Technologies
Supervised by

Dr. Michael McQuaid

School of Information

B. Thomas Golisano College of Computing and Information Sciences
Rochester Institute of Technology
Rochester, New York

March 2021

The thesis "GuavaNet: A deep neural network architecture for automatic sensory evaluation to predict degree of acceptability for Guava by a consumer" by Vipul Mehra has been examined and approved by the following Examination Committee:

Dr. Michael McQuaid
Senior Lecturer
Thesis Committee Chair

Edward Holden
Associate Professor

Dr. Erik Golen
Senior Lecturer

# Dedication

I dedicate my research work to my family and friends. A special thank to my loving sweet niece Ruhi, whose smile is always a relief and encouragement during my stressful moments.

# Acknowledgments

This research work is a part of the Master of Science in Information Sciences and Technologies at the School of Information, B. Thomas Golisano College of Computing and Information Sciences, Rochester Institute of Technology, Rochester, New York. I would like to express my gratitude to my committee chair Dr. Michael McQuaid, who always showed confidence in me and guided me throughout this project. I would also like to thank Edward Holden, Associate Professor and Dr. Erik Golen, Senior Lecturer, who are my other two committee members for their guidance and support. I would also like to thank my friends and family who supported me and offered their deep insight into the research work.

I wish to acknowledge the help provided by the technical and support staff, in the School of Information, B. Thomas Golisano College of Computing and Information Sciences, Rochester Institute of Technology, Rochester, New York. I would also like to show my deep appreciation to Dr. Qi Yu, our Program Director, who helped me in every aspect of research needed for the timely completion of thesis work. Special thanks to Dr. Tanupriya Chaudhary (Associate Professor Department of Informatics, University of Petroleum and Energy Studies, Dehradun, India and Honorary Secretary International Association of Professional and Fellow Engineers, Delaware, USA) for continuous encouragement and Dr. Tanmay Sarkar (Lecturer - Department of Food Processing Technology, West Bengal State Council of Technical Education, West Bengal, Malda-732102, India) for his expert opinion on the hedonic ratings given to guavas.

# Abstract

## GuavaNet: A deep neural network architecture for automatic sensory evaluation to predict degree of acceptability for Guava by a consumer

**Vipul Mehra**

**Supervising Professor: Dr. Michael McQuaid**

This thesis is divided into two parts: **Part I: Analysis of Fruits, Vegetables, Cheese and Fish based on Image Processing using Computer Vision and Deep Learning: A Review**. It consists of a comprehensive review of image processing, computer vision and deep learning techniques applied to carry out analysis of fruits, vegetables, cheese and fish. This part also serves as a literature review for Part II.

**Part II: GuavaNet: A deep neural network architecture for automatic sensory evaluation to predict degree of acceptability for Guava by a consumer**. This part introduces to an end-to-end deep neural network architecture that can predict the degree of acceptability by the consumer for a guava based on sensory evaluation.

**Keywords:** YOLOv4, CNN, VGG-16, ResNet-18, ResNet-50, Transfer Learning

# Contents

# List of Tables

# List of Figures

# Part I

# Analysis of Fruits, Vegetables, Cheese and Fish based on Image Processing using Computer Vision and Deep Learning: A Review

## 1 Abstract

This paper focuses on a detailed literature review on analysis of Fruits, Vegetables, Cheese and Fish using techniques of image processing, computer vision and deep learning in the food sector. Where the automation facilities are not available the food sorting and grading is done manually by humans but that is inconsistent, time consuming, expensive and can be easily influenced by locals. The food sector automation improves food grading based on quality, which is very important for exports to other countries and adds to economic growth and productivity of a nation and promotes healthy living among the consumers. An important sensory feature of food items is their appearance, which influences their market value and is to the liking of consumers. This paper provides a detailed overview of various image processing, computer vision and deep learning techniques used to address food quality, and also makes critical comparison of the various algorithms proposed by researchers for classification and quality inspection of fruits, vegetables, fish and cheese.

## 2 Introduction

Food is the primary source of energy that helps in growth, provides energy for activities, repair and other body functions. Most of us tend to repeatedly buy high quality food brands that meet our expectations. Similarly, even a small incident of compromised food quality can damage brand image, and companys profits may collapse. Thus appropriate quality control measures necessary for food brand management. Quality control through effective inspection, and control of production processes and operations helps food companies greatly reduce production costs by eliminating inferior products and waste production, which helps in greatly reducing production costs and better companys goodwill in consumers hearts. The company will also gain a good brand reputation increasing its chances of surviving in a highly competitive market. This is of great help in attracting more customers to use their products, and increase their sales. Therefore, for the well-being and safety of public consumers and the relationship between food quality and brand awareness, food safety 10 and quality inspections are essential. The precise detection applications of modern technology can help companies provide quality assurance consistently, because these technologies capture what is impossible for the human eye and conventional detection media to do. Although modern food inspection technologies are very effective in solving public health and safety issues, correct and successful inspection is also a key element in establishing and improving a companys brand image. The types of food products reviewed in this paper are natural products like fruits and vegetables and processed or man made food products such as cheese. The analysis of their being fresh or not fresh along with their nutritional content and the process employed in picking, storage or their making (processed food) help in determining whether the food is good for health or not. Therefore, the need for automatic detection of freshness of food is the need of the hour and Deep Learning techniques for image processing can provide the best techniques for food classification and quality determination.

# 3 Motivation

I always wanted to do something which satisfied my personal goal of opting for a thesis during my MS program, does my personal growth and is also for the common good. The works related to common good like general welfare, liberty, justice, common defense etc. are generally the responsibilities of governments. Although safe water and safe food also fall in this category, not enough work is carried out in this area related to common people. Therefore, I chose to do the thesis on something related to safe food, and chose some of the daily used food categories like fruits, vegetables, cheese, and fish specifically for the review and studies to understand the concepts used in the food industry, the reason this review paper came into being. The review of the works done by the researchers and food technologists, helped me gain insight into the safe food practices being followed in general. The knowledge gathered on food, thus paved the way for my current research work on guava.

# 4 Theoretical Background

This paper reviews studies conducted on digital image based classification of food products, identification of food diseases and freshness for natural agro-products fruit and vegetables, natural aquatic food fish and processed dairy food - cheese. Computer Vision System based methods are used for the identification of diseases in fruits and vegetables and their classification by analyses of their images, and also to assess quality and freshness of cheese and fish. Majority of studies on fruit and vegetable recognition or detection of food diseases have taken into account the sensory features such as color and texture of the classification. Full-scale research work on fruit recognition, classification, and their diseases and defects investigates the complete growth cycle of fruits on trees, starting with flowers and continuing until the fruits are ripe and ready to be picked. The same applies to vegetables. However, reviews of various research papers limit the classification methods used, and disease detection is performed for one disease at a time. The review includes a brief introduction of Computer Vision System (CVS) methods including AI/ML deep learning, CIElab color space etc. used in the food industry and supermarkets before jumping on to typical five system methods of computer vision systems related to food image analysis viz Image acquisition, Preprocessing, Image segmentation, Feature extraction, and Classification. The success of Classification, which is the final high level processing and decision making task, depends on the quality of input data prepared by the first four steps for the final step of classification used for sorting and quality grading of fruits, vegetables, cheese, and fish. The review sequence is therefore subdivided into parts, which are given below:

1. Introduction to Computer Vision
2. Five typical system methods of computer vision systems related to food image analysis
    - Image acquisition (Task 1)
    - Preprocessing (Task 2)
    - Image segmentation (Task 3)
    - Feature extraction (Task 4)
    - Classification (Task 5, the final task )

Hereunder we are discussing each of the above parts sequentially.

## 4.1 Introduction to Computer Vision (CVS)

The main aim of computer vision in digital image analysis is to make computers gain best possible understanding from acquired digital images data and automate the tasks that the human visual system can perform. Computer vision tasks include methods for acquiring digital images (or converting to digital images if captured image is analog), processing, analyzing and transforming those digital images (whether a single or a sequence of images) into high level numerical useful information, which when used as an input to a computer vision system developed for a specific task, can help in appropriate decision making. For example, In our review it is used for object detection, which can be a particular type of fruit, vegetable, cheese type or fish type (Huang, 1996). The traditional CVS came into use in the late 1960s and is widely being used in various industries and other crucial applications like industrial automation, security inspections, medical imaging, military applications, robotic guidance, food quality and safety inspections, etc.

Recent researches on use of computer vision in food classification and grading revived feature-based methods used in combination with ML techniques and complex optimization frameworks, which automatically evaluates and inspect texture, shape, color and size based features and defects of food items. Moreover new developments in deep learning have put some more teeth into the field of computer vision and with the use of deep learning algorithms in computer vision datasets the accuracy achieved is much better than previous methods (Forsyth and Ponce, 2003). Computer vision systems used in the food industry are made up of a combination of hardware devices which include optical, electromagnetic, digital, and software systems and algorithms used for food identification, freshness and quality based grading. In short these systems are formed by integrating various mechanical systems with optical and electronic devices and softwares to handle all these devices and algorithms (Morris, 2004).

Moreover it is contactless, non destructive, which are essential requirements for handling food products and online inspection, and also being cost effective, is widely used all over the food industry. Images are captured by camera under proper lighting conditions. The image is captured in the visible spectrum of light and the density of photons reflected depends on the surface of the object being photographed, which can be partially reflective at some points and partially absorptive at other portions of the object surface. The camera lens basically being a light collector, gathers the scattered photons and converts those signals into an image. Once the image is captured, it is digitized, that is converted the images into numerical data, through a process called digitization. If the camera used is analog then the analog image captured is converted into pixels using a digitizer. The digitizer is also known as frame grabber. Since nowadays high quality digital cameras and scanners are available, this process of Digitization is not required. Digitization of images breaks the image information into a two-dimensional grid consisting of small regions, which contain very small sized picture elements, known as pixels. The pixel information is then stored in the form of matrices. Refer to figure 1 below

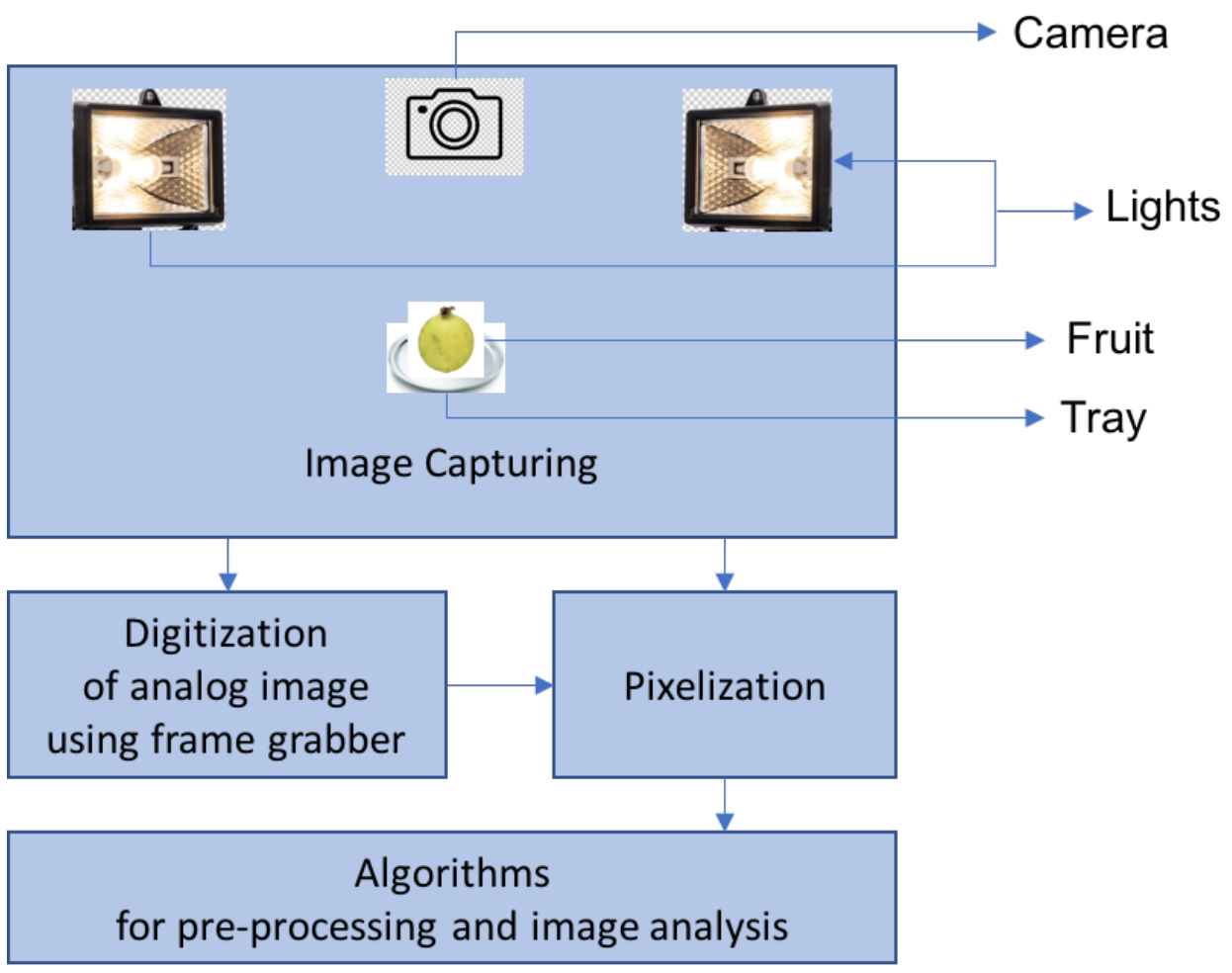

Figure 1. Block diagram of Image capturing and digitization

Thus computer vision systems as a technique are able to quantify the external features of products from the digitized images so obtained, and thus a trustworthy technique for recognizing and classifying the objects. Presently the computer vision systems are mainly used in the food industry for carrying out automated inspection, which facilitates the producers/manufacturers of food products to keep a check on quality (Stockman and Shapiro,

2001).

## 4.2 Computer vision system methods related to food image analysis

The configuration of a computer vision system depends on the application. For our review, it could be a stand alone system if used in small groceries stores, whereas if it is for supermarkets, it could be a sub-system of a larger system such as for sorting and grading of food items, it may have to control mechanical actuators, and also have other subsystems and information databases for grading of food items, and some other man-machine interfaces. For the image based classification of food items, a typical computer vision system need to perform five typical tasks, refer to figure 2 below showing the flow diagram of the five tasks, and thereafter the explanation of the functionality of each task follows:

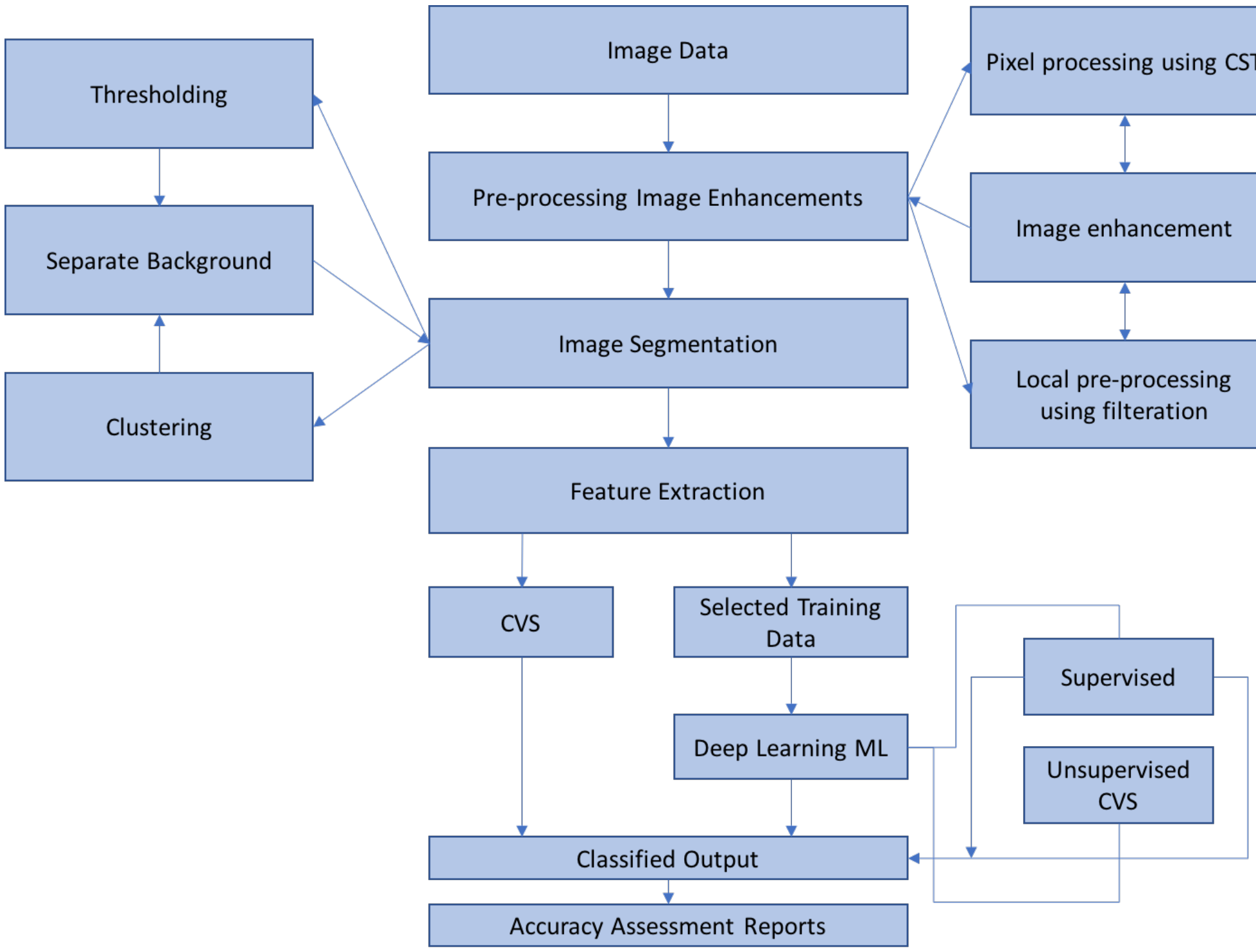

Figure 2. Block Diagram of typical functions of a computer vision system

### 4.2.1 Image acquisition (Task 1)

Digital images are produced by image sensors, which may include distance sensors, tomography devices, radar, ultrasound cameras, etc., in addition to various types of photosensitive cameras to produce image data, which could be an ordinary 2D image or a complex 3D or sometimes an image sequence as well, all depends on the sensor used. The pixel values of the captured digital images can be in the spectral bands of gray or color images depending on the light intensity, but there may be exceptions in which these pixel values may relate to physical measures, such as depth, absorption or reflectance of sonic or electromagnetic waves.

A typical CVS contains five basic components, refer to figure 3 below: motor, image capture board (digitizer or frame grabber), illumination, camera, and computer hardware. When analyzing the product, the illumination structure provides front and back lighting. Front lighting required to check surface quality attributes such as color, texture and skin defects. Backlight checks boundary quality attributes such as size and shape. Traditional, multispectral and hyperspectral computer vision systems are widely used for quality analysis of processed food like cheese, aquatic food items like fish and the natural agricultural products like fruits and vegetables (Sonka et al., 2008).

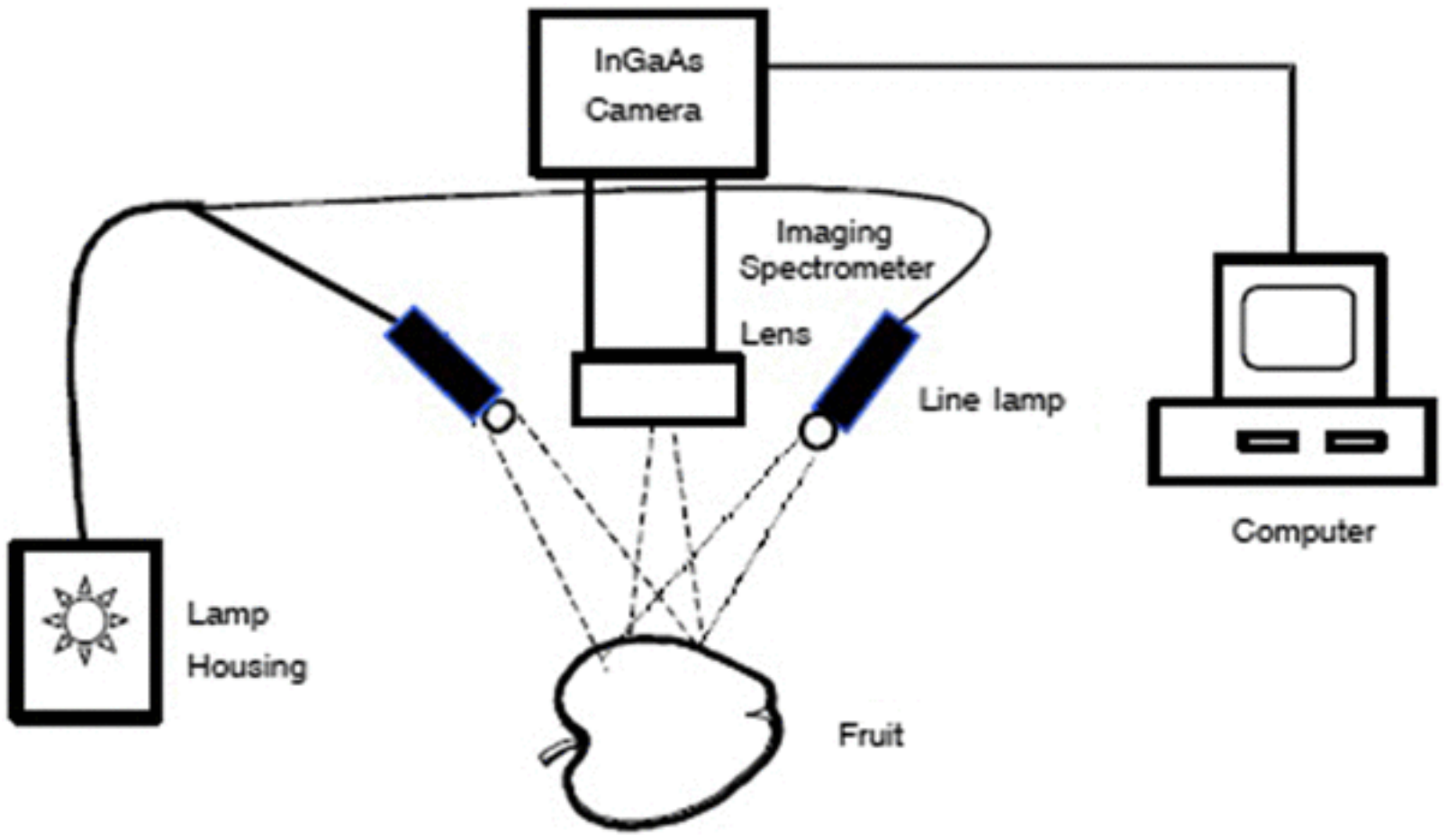

Figure 3. Components of a Computer Vision System, from Baneh et al., 2018

### 4.2.2 Preprocessing (Task 2)

Raw images taken by camera or acquired through various other means may have multiple noises, and therefore fail to deliver the best results when used for image analysis in computer vision applications. Therefore, before applying computer vision methods to image data for extracting features or specific information about regions of interest, the data required to be preprocessed to ensure that

- Marginal distortions are removed by correcting the image coordinate system.
- Essential features of image expanded by reducing noise, and establishing a certain image (degraded form), which is more deterministic than the original captured image.
- Contrast enhancement.
- Converting image into numerical dataset for mathematical normalization and further image enhancement of the image structure at local suitable scale.

The most common method of pixel preprocessing is color space transformation (CST), which is used to evaluate food quality. Most CST applications depend on the Hue, Saturation, and Intensity (HSI), where saturation results in monochromatic images, which vividly impart texture to food images. Local preprocessing (Filtration) uses a small part of the neighborhood of pixels in the input image to generate new brightness values in the output image. It uses a simple filter (to reduce noise), a median filter (to reduce peak noise) and an improved unfrosted filter (to identify cracks in eggs).

The preprocessing techniques in use are as follows: Refer to Table 1

1. Pixel Brightness Transformation (PBT)
2. Geometric transformation
3. Image filtering

#### 4.2.2.1 Pixel Brightness Transform (PBT) operations:

PBT is used to control and transform pixel brightness and contrast values; the output pixel's value is a function of the corresponding input pixel value.

As shown in Table 1, the three most used brightness transform operations are

1. Gamma correction
2. Sigmoid stretching
3. Histogram equalization

There are two point processes viz. multiplication and addition, which use a constant g(x)=$\alpha$f(x)+$\beta$ where $\alpha>0$ is the gain parameters to control the contrast, and $\beta$ is the bias parameter that controls brightness. The image brightness and contrast varies according to the values of parameters alpha and beta respectively.

**Gamma Correction** Gamma Correction carries out a non-linear operation on the individual pixel values of the source image, and can cause saturation of the image being altered, whereas linear operations like scalar multiplication and addition/subtraction, are carried out on individual pixels during image normalization.

$$O = \left(\frac{I}{255}\right)^{\gamma} \times 255$$

The above formula shows the relation between output image and gamma correction, see the plot showing the non linear relationship (GreatLearningTeam, 2020).

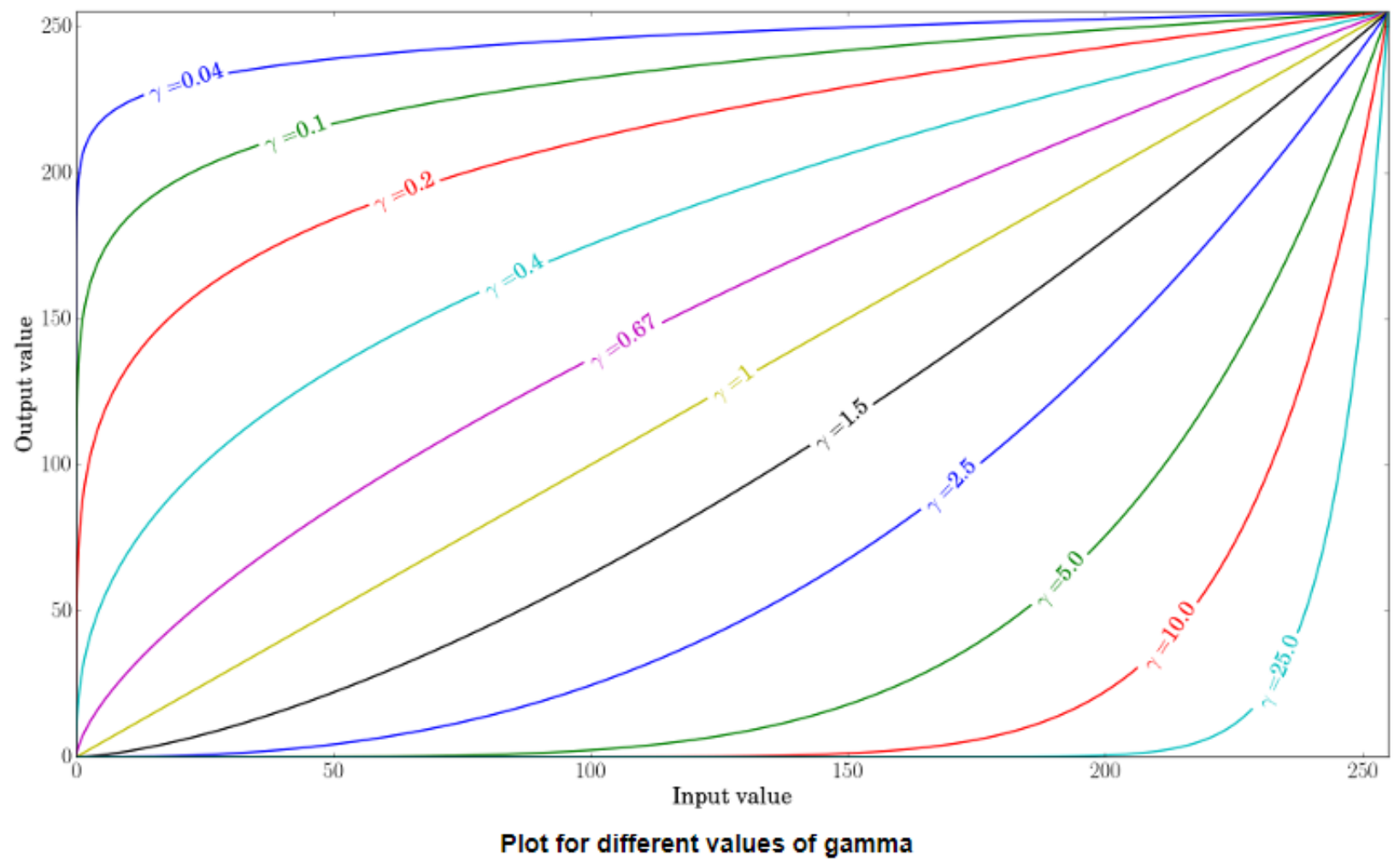

Figure 4. Plot for different values of Gamma, from GreatLearningTeam, 2020

**Histogram equalization** It is a technique for contrast enhancement by transforming an image so that its intensity histogram gets the desired shape. The normalized histogram. P(n) = number of pixels with intensity n / total number of pixels (GreatLearningTeam, 2020).

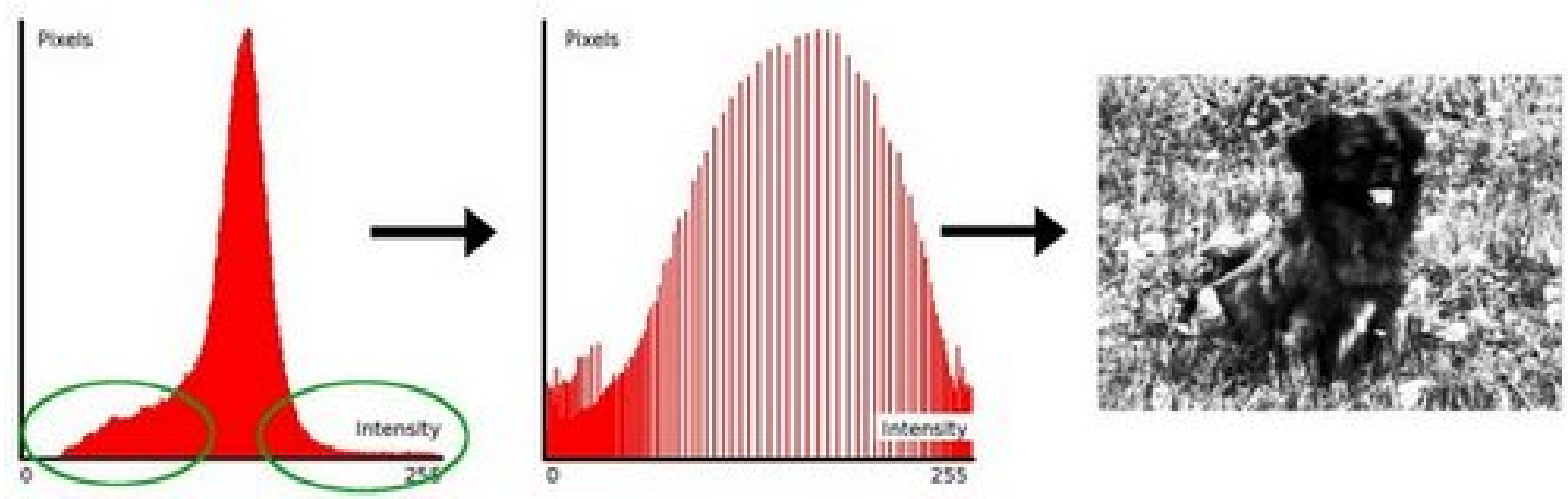

Figure 5. Histogram Equalization, from Bradski, 2000

**Sigmoid stretching** Sigmoid stretching is a continuous nonlinear activation function. Being an "S" shaped function it got its name, sigmoid.

$$S(x) = \frac{1}{1+e^{-tx}} \qquad f(x,y) = \frac{1}{1+e^{\left(c*\left(th-g_s(x,y)\right)\right)}} \tag{1}$$

Where f(x,y) is an enhanced pixel value, c - contrast factor, th - Threshold value, gs(x,y)- original image. By changing the value of 'c' and threshold value of contrast can be controlled (GreatLearningTeam, 2020).

#### 4.2.2.2 Geometric Transformations

It modify positions of pixels in an image, eliminate geometric distortions during image capturing, but keep the colours intact/unchanged. Rotation, scaling and distortion of images are part of the normal Geometric transformation operations (Bradski, 2000). Geometric transformations perform two basic steps , the first step does physical rearrangement of pixels in the image, also known as 'Spatial transformation', and the second step assigns grey levels to the transformed image, and is called 'Grey level interpolation' (GreatLearningTeam, 2020). Geometric Transformation Operations include:

1. **Scaling**, which is just resizing of image

$$\begin{array}{l} x' = x \cdot S_x \\ y' = y \cdot S_y \end{array} \Rightarrow \begin{bmatrix} x' \\ y' \end{bmatrix} = \begin{bmatrix} S_x & 0 \\ 0 & S_y \end{bmatrix} \cdot \begin{bmatrix} x \\ y \end{bmatrix} \tag{2}$$

2. **Translation** which shifts location of the object

$$\begin{array}{l} x' = x + \Delta x \\ y' = y + \Delta y \end{array} \Rightarrow \begin{bmatrix} x' \\ y' \end{bmatrix} = \begin{bmatrix} x \\ y \end{bmatrix} + \begin{bmatrix} \Delta x \\ \Delta y \end{bmatrix} \tag{3}$$

3. **Rotation** will be rotates an object with theta degrees

$$\begin{array}{l} x' = x \cdot \cos\phi - y \cdot \sin\phi \\ y' = x \cdot \sin\phi + y \cdot \cos\phi \end{array} \Rightarrow \begin{bmatrix} x' \\ y' \end{bmatrix} = \begin{bmatrix} \cos\phi & -\sin\phi \\ \sin\phi & \cos\phi \end{bmatrix} \cdot \begin{bmatrix} x \\ y \end{bmatrix} \tag{4}$$

4. **Shearing**, does horizontal shifting of pixels.

$$\begin{array}{l} x' = x + y \cdot J_x \\ y' = x \cdot J_y + y \end{array} \Rightarrow \begin{bmatrix} x' \\ y' \end{bmatrix} = \begin{bmatrix} 1 & J_x \\ J_y & 1 \end{bmatrix} \cdot \begin{bmatrix} x \\ y \end{bmatrix} \tag{5}$$

5. **Affine Transformation** is the combination of four transformations including the scale factors, the shearing factors and the rotation angle, which are merged into one matrix.

$$\begin{array}{l} x' = a_1 \cdot x + a_2 \cdot y + a_3 \\ y' = b_1 \cdot x + b_2 \cdot y + b_3 \end{array} \Rightarrow \begin{bmatrix} x' \\ y' \end{bmatrix} = \begin{bmatrix} a_1 & a_2 \\ b_1 & b_2 \end{bmatrix} \cdot \begin{bmatrix} x \\ y \end{bmatrix} + \begin{bmatrix} a_3 \\ b_3 \end{bmatrix} \tag{6}$$

6. **Perspective Transformation**, as the name suggests, this transformation alters the perspective of an image for better insight into the information around a point on the image, which you want to gather by changing the perspective.

**Interpolation** Interpolation is needed when the new point coordinates (x,y) obtained by using transformation methods do not fit properly into the discrete raster of the output image. The use of interpolation operations helps to obtain the individual pixels value in the output image raster. The below expression is for the brightness value of the pixel (x,y) in the output image, x, y belongs to the discrete raster (GreatLearningTeam, 2020).

**Interpolation Types:**

1. Nearest neighbor interpolation re samples the pixel values in the input matrix.

$$s(x, y) = f_s(\text{round}(x),\ \text{round}\ (y)) \tag{7}$$

2. Linear interpolation: According to the assumption made by linear interpolation, the four points in the neighborhood of point (x,y) are explored by the the brightness function, which is also considered linear in the neighborhood of point (x,y).

3. Bicubic interpolation uses sixteen neighboring points for interpolation, approximating locally by a bicubic polynomial surface, to improve the brightness function model.

$$h_3 = \begin{cases} 1 - 2|x|^2 + |x|^3 & \text{for } 0 < |x| < 1 \\ 4 - 8|x| + 5|x|^2 - |x|^3 & \text{for } 1 < |x| < 2 \\ 0 & \text{otherwise} \end{cases} \tag{8}$$

#### 4.2.2.3 Image Filtering

A filter is defined by the kernel - a small array of numerical data for each pixel and pixels in its neighborhood in an image. Filtering modifies the image, enhancing its properties, also useful for extraction of valuable information about the edges, corners, and blobs in the images.
The review paper discussed some popular basic filtering techniques used by researchers in pre-processing process for betterment of image quality, which include:

1. Low Pass Filtering is used for smoothing of images by minimizing disparity in nearby pixels, taking the mean of pixel values.

2. High pass filters are used for sharper images and for detection of edges. The convolution kernel used in this technique is different from that of low pass filtering.

3. Directional Filtering computes the first derivatives (gradient), thus is a useful method for edge detection. Wherever a significant difference is noticed in adjacent pixel values of an image, generally an indication of presence of edges, direction filtering is used to compute slopes to confirm the presence of edges.

4. Laplacian Filtering computes the second derivatives of an image to measure the rate at which the gradient changes. This confirms the presence or absence of edges, thus also acts as an edge detector. The kernels used in Laplacian filtering operations usually have negative values in a cross pattern in the middle of the data array of images, the center value could be positive or negative, whereas the corner values are zero or positive (GreatLearningTeam, 2020).

Table 1. Summary of Image preprocessing techniques

| Sno | Technique | Goals | Steps | Transform Operation Used |
|---|---|---|---|---|
| 1. | Pixel Brightness Transformation (PBT) | 1. Modify pixel brightness, 2. Grayscale transformation | Brightness/contrast adjustment, color correction and transformations | 1. Gamma correction 2. Sigmoid stretching 3. Histogram equalization |
| 2. | Geometric transformation | 1. Modify positions of pixels 2. Keep colors unchanged | 1. Rearranging pixel's positions (spatial transformation) 2. Assign grey levels Grey level interpolation) | Transformations:1. scaling 2. Translation 3. Rotation 4. Shearing 5. Affine 6. Perspective Interpolation methods: 1. Nearest neighbor 2. Linear interpolation 3. Bicubic |
| 3. | Image filtering | 1. Minimise pixel disparity 2. Sharper images, 3. Identify Edges. 4. Confirm pixels of edges | 1. Smoothing, 2. Edge detection, sharpening, 3. compute slope-1st derivative 4. Compute second derivative | 1.Smoothing-low pass filters 2. High pass filters-Edge detection/sharpening 3. Direction filtering- for first derivative 4. Laplace filtering for 2nd derivative |

### 4.2.3 Image segmentation (Task 3)

There are occasions when we need to decide on relevant image points of interest or regions with an object of interest before carrying on the further processing of images. Therefore, after preprocessing, image segmentation is required to further improve the image quality for the purpose of its analysis to divide the digital image into different regions. The issues involved are:

- Extraction of more than one feature needed for recognizing the produce as there may be variations in shape, texture and color depending on ripeness.
- The problem of hue and reflections when food is in plastic.
- Use of binary classifiers in a multiclass scenario.
- Reducing the background for clarity of image.
- Precise defect segmentation for the disease recognition system.

- Better performance of systems with less training data.

Therefore the main function is to separate the background to deal with important areas during object evaluation. The image is segmented into multiple parts based on commonalities in pixels characteristics, called regions of interest (ROI). The most important functionalities of 'Image segmentation' methods are separating image foreground from background, and clustering regions of interests based on similarities in color and/or shape of pixels.

#### 4.2.3.1 Image segmentation techniques types:

- Non-contextual segmentation: The most commonly used example is 'Thresholding', deploying a single threshold, is useful in transforming greyscale or color images into binary images, also known as 'binary region map'. Depending on the value of threshold, the binary map classify two distinct types of regions of interest based on input data values lower than threshold or above the threshold. Thresholding could be 'Simple', 'Adaptive, and Colour thresholding.

- Contextual segmentation is better than the non contextual in separating individual pixels because it takes into consideration the closeness of pixels, which is not done in 'Non-contextual thresholding' operations. Two basic approaches to contextual segmentation uses signal discontinuity, and similarity as two basic approaches to define ROIs (Regions of interest). Assuming discontinuity or abrupt signal changes is an indication of boundary change, signal continuity technique marks boundaries for uniform regions. Whereas use of certain similarity criteria are used to define uniform regions by grouping together connected pixels that satisfy the similarity criteria used for 'Similarity' based techniques. The types of Contextual segmentation are 'Pixel connectivity', 'Region similarity', 'Region growing', and 'Split-and-merge segmentation'.

- Texture Segmentation : Texture is another important feature in many food image analysis applications. The texture problems are handled based on the four subcategories, which are 'structural approach', 'statistical approach', 'model based approach', and 'filter based approach' (Bhargava and Bansal, 2018).

#### 4.2.3.2 Fourier Transform

It is used to decompose an image into sine and cosine components as an output of the transformation, which is known as Fourier domain or Fourier transform. As we know that input is a spatial domain image, and particular frequency related to each point in this correspond to a unique frequency in Fourier domain image. The Fourier Transforms are useful in applications, which include image filtering, reconstruction and image compression.

A sampled Fourier transform technique mostly used is 'Discrete Fourier Transform' (DFT). This technique uses a set of samples, large enough to describe the spatial domain image fully, and the pixel count is the same in the spatial domain image and the obtained Fourier domain (GreatLearningTeam, 2020).

The two-dimensional DFT is represented by the formula given below, for a square image of size NxN.

$$F(k,l) = \sum_{i=0}^{N-1} \sum_{j=0}^{N-1} f(i,j) e^{-\imath 2\pi \left( \frac{ki}{N} + \frac{lj}{N} \right)} \tag{9}$$

And the Inverse Fourier Transform is represented by the equation

$$f(a,b) = \frac{1}{N^2} \sum_{k=0}^{N-1} \sum_{l=0}^{N-1} F(k,l) e^{\imath 2\pi \left( \frac{ka}{N} + \frac{lb}{N} \right)} \tag{10}$$

Table 2. Summary of Image Segmentation techniques

| Sno | Technique | Goals | Steps | Transform Operation Used |
|---|---|---|---|---|
| 1. | Image segmentation | 1. Separating background 2. clustering similar pixels | 1. Non Contextual-Thresholding 2. Contextual 3. Texture | ThresholdingTypes: Simple/Adaptive/ Color Contextual Segmentation types: Pixel connectivity, Region similarity, Region growing, Split-merge Texture Segmentation |
| 2. | Fourier transform and image restoration | Image reconstruction, restoration, compression, Filtering, and analysis | 1. Decompose image into sine and cosine components 2. Transform Input (spatial domain) image into equivalent Fourier domain output | Discrete Fourier Transform (DFT) |

Correct segmentation is critical to further development of image analysis, and incorrect segmentation will reduce the performance of the classifier. The two widely used segmentation techniques are thresholding and clustering. Threshold processing divides each pixel in the image into two categories, namely the interest area and the background area. Pixels with specific gray levels belong to the attention category area, and the pixels with equal gray level belong to the background category.

### 4.2.4 Feature Extraction (Task 4)

Next step after image segmentation is feature extraction. These functions are fundamental factors in CVS because they contain valid data for image perception, interpretation, and object classification. In this process extracted features form a feature vector classified as recognition input. These feature vectors uniquely and accurately define the shape of the object. Color, texture and morphological features are often used to analyze the defects and maturity of food items. Table 3 below summarises the color and the texture features for fruits and vegetables classification (Dubey and Jalal, 2015).

Table 3. Summary of color and texture features

| Feature Name | Feature Covered | Description |
|---|---|---|
| Global color Histogram (GCH) | Color | Encodes the image information related to each distinct color, a set of ordered values, which represent the probability of a pixel of the same color. Uniform normalization and quantization employed to reduce the numbers of distinct colors, and to avoid scaling bias. |
| Color Coherence Vector (CCV) | Color | CCV is the degree to which image pixels of the same color belong to a coherent region, that is a region with homogeneous color. Only coherent pixels are considered part of that contiguous region, and not the incoherent pixels. The methods employed for computation of CCVs eliminate variations between neighboring pixels by blurring and discretizing the color space of the image. Next it classifies pixels as coherent or incoherent, afterwards two color histograms for coherent and incoherent pixels are computed, and both are then stored as a single histogram. |

Continued on next page...

| Feature Name | Feature Covered | Description |
|---|---|---|
| Border/Interior Classification (BIC) | Color | If the pixel surrounded by its four neighbor pixels at top, bottom, left, and right, is an interior pixel, otherwise the pixel is classified as a border pixel. Afterwards two color histograms for border and interior pixels are computed. |
| Local BinaryPattern (LBP) | Texture | By thresholding the neighborhood of each pixel, the image pixel is labeled as a binary number. It's robust to monotonic gray-scale changes due to lighting intensity variations. It can analyze images in challenging real-time environments. |
| Completed Local Binary Pattern (CLBP) | Texture | The local region of CLBP is defined by a center pixel and LDSMT(Local Difference Sign - Magnitude Transform). The center pixels representing the image gray level are converted by global thresholding into a binary code CLBP-Center (CLBP_C), and the LDSMT decomposes the image local differences into the signs and the magnitudes, the two complementary components, and CLBP-Sign (CLBP_S) and CLBP-Magnitude (CLBP_M), the two operators, are proposed to code them. CLBP_S is the equivalent of LBP, with more local structure information than CLBP_M, this is because the LBP is well suited to extract the texture features.A combination of CLBP_S, CLBP_M, and CLBP_C can improve rotation invariant texture classification significantly. |

Continued on next page...

| Feature Name | Feature Covered | Description |
|---|---|---|
| Unsers Feature (UNSER) | Texture | The first step considers displacement (d1, d2) of an image to measure the sum and difference of intensity values, the calculations done for sum and difference histograms, which are then stored as a single histogram. |
| Improved Sum and Difference Histogram (ISDH) | Texture | ISDH encodes a sum and difference histogram based on the neighborhood information of an image pixel.The calculations are carried out for the x and y directions neighboring pixels separately, and the outputs simulated in y-direction by calculating sum and difference in the y-direction itself, thus making the algorithm very efficient. |

The inspection for quality of fruits and vegetables is done most of the time using RGB(Red, Green, Blue), HSI(Hue, Saturation, Intensity) and CIELab color spaces. You can specify a color space to extract color features from an image. Images acquired through the common RGB color model are based on the original colors red (R), green (G), and blue (B). This color model divides the image into red, green and blue planes and determines all color moments (Mustafa et al., 2011). In an image, for the same pixel, different RGB devices will produce different RGB values. In order to standardize these values, several conversion techniques are used. Because RGB is non-linear in human visual inspection, it is impossible to analyze the sensory characteristics of food. To overcome this problem, HSI was proposed and developed, which is the main tool for developing image processing algorithms based on colors commonly and accepted by humans. However, HSI is similar to RGB and is not sensitive to small color changes. Therefore, these are not recommended for evaluating product color changes during processing. CIELAB color space, which can characterize all colors clearly visible to the human eye, and is designed as a device-dependent model for reference, where "L" is a measure of brightness, and "a" and "b" change red/green and green/blue balance. It is perceptually unified, so the color difference perceived by humans is the same as the Euclidean distance in CIELAB space. Since the color obtained from the CIELab color space can be used to easily analyze the color measured by computer vision, it provides a feasible method to evaluate the color performance of the object. Refer to table 4 below:

Table 4. CIELab color space summary

| Color parameter | Parameter represent | Color value | Value meaning |
|---|---|---|---|
| L* or Lstar | Perceptual lightness value. | Range 0 to 100 | '0' is black, and '100' is white |
| a* axis | Represent relative concentration of green-red opponent colors. | Range, -128 to 127 | Negative value means more towards green and Positive value means more towards red. |
| b* axis | Represent relative concentration of blue-yellow opponent colors. | Range, -128 to 127 | Negative value means more towards blue, Positive value means more towards yellow. |

Deep learning techniques are becoming very popular in the food industry due to its ease in feature extraction leading to classification and grading of food items. Analysis of food images using CNN(Convolutional Neural Networks) is the most common methodology of Deep Learning in food recognition and classification. CNN(Convolution Neural Network) are being used widely in agro-based industries for food classifications based on category discrimination, and identification of ingredients. Most popular image processing CNN architectures are AlexNet (Alex Krizhevsky, 2012), a Visual Geometry Group (VGG) network that uses iterative units, ResNet (Residual Neural Network), and parallel data channels such as VGG (Karen Simonyan, 2014) and GoogLeNet (Szegedy et al., 2015) and those constructed by residual blocks (He et al., 2016). All these network architectures, with pretrained weights like ImageNet (Deng et al., 2009), can be downloaded from 'Model Zoo'. As all these pretrained models can extract image features like color, texture, high-level abstract representations etc,. researchers can use their specific image datasets for transfer learning by FineTuning, i.e retraining with their weights for final classification, keeping the existing weights of the convolution layer or adjusting the weights of the whole network slightly. This methodology shortened the training time and provided more accurate results (Dubey and Jalal, 2015).

#### 4.2.4.1 Color features

One of the major factor that affects customers rejection or choice of food items, whether natural foods or processed foods is 'color'.

#### 4.2.4.2 Size and shape feature extraction for fruits and vegetables:

Size and shape are known as morphological features, which are used frequently in the agriculture industry for classification of fruits and vegetables. Since the price relates to the size of fruits and vegetables, therefore fruits and vegetables are graded in groups according to their size. The feature ‘size’ is quantified by using measures of length, width, projected area, perimeter, major and minor axes, and is helpful in automated sorting in industries. The actual number of pixels in a region are calculated from the ‘area’, and the distance between two neighboring pixels. Once an object is segmented, measures of its area and perimeter are stable irrespective of its shape or orientation.

The quantum of produce is usually assessed from the measure of length and width of the produce. Usually during processing, the product shape changes, thus it is necessary to restore in time the direction of the calculated length and width. The major axis is the longest line between two boundary pixels across the object and the longest line perpendicular to the major axis is the minor axis. Shape is another key visual feature of image description, but it is not possible to determine similarity of shapes in fruits and vegetables as content regarding shape can’t be measured accurately. There are two types of shape descriptors, a) region-based, which consider the overall area of the object and b) contour-based, which use local features for boundary segmentation.

| | |
|---|---|
| Centre of gravity | $g = (\frac{1}{n}\sum_{i=1}^{n} x_i, \frac{1}{n}\sum_{i=1}^{n} y_i)$ |
| Radial distance | $\rho_i = \| p_i - g \|_2$ |
| Average bending energy | $E_b = \frac{1}{n}\sum_{s=0}^{n-1} k(S)^2$ |
| Circularity area ratio | $\zeta_A = \frac{A_{shape}}{A_{circle}} = \frac{4\pi A_{shape}}{P_{sphere}}$ |
| Circularity perimeter | $\zeta_P = \frac{A_{shape}}{P^2}$ |
| Circle Variance | $\zeta_\rho = \frac{\phi_p}{\mu_p}$ |
| Rectangularity | $\zeta_R = \frac{A_{shape}}{A_{box}}$ |
| Convexity | $\zeta_C = \frac{P_{hull}}{P_{shape}}$ |
| Solidity | $\zeta_S = \frac{A_{shape}}{A_{hull}}$ |
| Hole area ratio | $\zeta_H = \frac{A_{hole}}{A_{shape}}$ |
| Eccentricity | $\zeta_e = \frac{\lambda_1}{\lambda_2}$ |
| Ellipse Variance | $d = \sqrt{\rho_i^T M^{-1} \rho_i}$ |
| Profile | $\phi_x(i) = \forall_{x=i} : y_{max} - y_{min}$ |

Figure 6. Formulae for Shape Descriptors, from Hameed et al., 2018

#### 4.2.4.3 Texture features extraction

A large number of food images are recognized/interpreted using the human visual system, texture is useful for suitable classification of food items. The texture is measured using the pixel group, which represents the distribution of elements and the appearance of the food surface, and used in machine vision to predict the quality of the food surface in the form of roughness, contrast, entropy, direction, etc. The texture feature is also useful in determining maturity and sugar content (the internal quality) of fruits. Texture also segregates different image patterns by extracting the intensity values between pixels, and their quantitative and qualitative analysis. Quantitative analysis helps classification according to six texture characteristics, namely contrast, roughness, linearity, directionality, roughness and regularity. According to (Bagri and Johari, 2015) qualitative analysis can determine four features, which are contrast, correlation, entropy and energy. The different types of Texture features are broadly classified as:

1. Statistical texture extracts the gray level matrices such as co-occurrence matrix, pixel run length matrix, and neighboring gray level dependence matrix, based on intensity values of pixels.
2. Model-based texture includes fractal, random field, and autoregressive models.
3. Structural texture includes lines, edges constructed by pixels intensity.
4. Transform-based texture extracted spatial domain images.

The commonly used texture type is statistical texture because of its lower computational cost and higher accuracy.

### 4.2.5 High-level processing and decision making or Classification (Task 5)

The food products brought to many markets often have different properties and may be delivered immature, wrinkled, damaged or contain rotten material. Such food items may be disastrous for health, and also fetch low prices causing loss to sellers. The damaged and malformed food items need to be categorized and healthy food items should be graded according to size, weight, shape, color, maturity etc. In the case of agro - products, especially fruits and vegetables, the post harvest grading operation plays a very important role in removing unwanted or foreign matter from the harvested crop. Grading allows classifying fruits and vegetables into different grades according to size, shape, and color. Thus, the sensory features such as color, size, shape, and texture are the most important parameters for food classification algorithms and are chosen to form a training set, and then a classification algorithm is applied to create the knowledge used to identify unknown cases. Many CVSystems methods based on NN(Neural Network), SVM(Support Vector Machine), ANN(Artificial Neural Network), CNN(Convolutional Neural Network), are developed and are in use for food classification based on quality. For example the KNN(k-Nearest Neighbor) based system spots similarity in samples through distance metrics. It first selects k neighbors, and then counts the number of data points based on the Euclidean distance in each category, and then selects a new point to repeat count for the new point. In the SVM based classification algorithm both linear and non-linear data are classified. Kernel function maps data non-linearly to a high-dimensional space. SVM, when used for two-class problems, finds the linear optimal hyperplane to maximize the distance between extreme pole points of both the classes, called support vectors. The development of deep learning and CNN is very effective for food classification. Deep learning learns image features, extracts contextual details and global features, which will help significantly reduce errors. The classification is therefore a two-step process. Since the computer vision system now has the features as input for final analysis of the images to obtain the desired output in the form of food objects detected and sorted and then graded based on the quality, therefore in the first step the different varieties are sorted and separated, and the second step grade the separated varieties according to the quality. And then come the final step to determine how our computer vision system performed, whether our automated detection application is successful, or didn't perform as per the expectations, and needs a rework. The two steps of the classification process follows:

#### 4.2.5.1 Sorting to separate varieties of food items

This is the first type of classification, where the combined produce in the supermarkets needs to be separated into individual food items, for example to separate apples from a mixed lot of fruits and vegetables. As mentioned above this can be done manually or by using automated sorting machines controlled by image processing systems.

#### 4.2.5.2 Defects/Diseases and quality grading of Food

A disease outbreak can result in significant loss in yield and quality of produce. Therefore, there is a need to inspect the produce on a regular basis 24x7. There are some peculiar diseases with which if a fruit on a tree gets infected, can also spread to other parts like branches, leaves and other fruits on the tree. Some examples of common diseases of apple fruit are apple scab, apple rot and apple blotch (Hartman, 2010). Apple scab will cause gray or brown to olive-green spots that enlarge to become almost circular. Apple rot will cause slight depressions, round brown or black spots, which are sometimes covered by a red halo. In order to find out what measures are required to be taken for disease control to avoid the similar losses during the next harvest season, there is an urgent and essential requirement to recognize the symptoms of food diseases, and there is a need for automatic detection as soon as the onset of disease starts on the crop, trees, or in storages of food items, so that the losses can be minimized with immediate remedial measures. Due to huge agricultural fields, automated monitoring is a very tough task. Moreover huge difference in defect and disease types, detection is still a difficult task (Unay and Gosselin, 2006). Though CVS provides automated, cost-effective and non-destructive technology, it's still not near to perfection and also insufficient to meet the ever rising demand for quality food.

Grading is usually based on physical properties such as weight, size, color, shape, specific gravity, diseases and defects. For fresh marketing of food, the known grading methods are manual grading or machine grading. With either method, the produce is graded based on size. Fruits and vegetables are usually graded based on state, federal, and international standards. All countries set their own standards for different grades according to market requirements. However, in the international market, three general grades are considered: Extra Class, Class I and Class II.

# 5 Review

## 5.1 Grading

According to (Pathare et al., 2013) color for fruits and vegetables, is an indirect measure of their quality, which is an indicator of their freshness, maturity and safety, which depends on physical and chemical changes, internal biochemistry, microorganisms that occur during maturation, growth and harvest post-processing and handling stages. The color function has many advantages, such as high eciency, easy to extract color information from the image, independent of size and direction, powerful in expressing the visual content of the image, robust to background complexity, separation of images from each other. In an image, for the same pixel, different RGB(Red, Green, Blue) devices will produce different RGB values. In order to standardize these values, several conversion techniques are used. Because RGB is non-linear in human visual inspection, it is impossible to analyze the sensory characteristics of food. To overcome this problem, HSI(Hue, Saturation and Intensity) was proposed and developed, which is the main tool for developing image processing algorithms based on colors commonly and accepted by humans. However, HSI is similar to RGB and is not sensitive to small color changes. Therefore, these are not recommended for evaluating product color changes during processing. CIELAB color space, which can characterize all colors clearly visible to the human eye and is designed as a device-dependent model for reference, where "L" is a measure of brightness, and "a" and "b" change red/green and green/blue balance. It is perceptually unified, so the color difference perceived by humans is the same as the Euclidean distance in CIELAB space. Since the color obtained from the CIELAB color space can be used to easily analyze the color measured by computer vision, it provides a feasible method to evaluate the color performance of the object.

(Pereira et al., 2018) worked on papaya fruit ripening. They used digital imaging and random forest techniques for predicting the ripening of papaya by analysing the color of its peel. Therefore, the color features of the papaya are extracted and worked on to define whether the papaya is ripe or not. This technique is cost effective and provided a high accuracy result of 94.30.

In the case of dairy products like processed cheese, there are many factors which affect their color, such as their milk composition, the food additives, manufacturing technology used, natural milk flora activity, and mature technology used in the manufacturing process. The color measurement can detect certain abnormalities or defects that may exist in the food so produced. Consumers requirements for quality of dairy products continue to increase, requiring faster, objective and accurate food color evaluation. Since visual sensory evaluation is laborious, time-consuming, expensive and tedious, therefore, color measurement is commonly used in quality control and product development to evaluate the color of curds and cheese. The natural milk flora activities, the technological processes, and the maturation technologies used, can change the color of cheese. The color of cheese is also related to the cow's diet, as well as pigments used, and the cheese varieties. Recent studies have also emphasized the potential role of colorimetry in assessing cheese smear maturity and measuring defects in cheese maturation (such as browning) (Carreira et al., 2002) (Dufossé et al., 2005) (Olson et al., 2007).

(Carreira et al., 2002) determined these factors that ultimately result in the cheese browning and specifically probed on the browning of Camembert cheese.

(Dufossé et al., 2005) provided the qualitative analysis of red-smear soft cheese by using Spectrocolorimetry.

(Rohm and Jaros, 1996) and (Rohm and Jaros, 1997) studied the relationship between cheese color and maturation time, and the results showed that the value of L* ("L" is a measure of brightness) decreased, and the values of a* and b* ("a" and "b" change red/green and green/blue balance) increased during the maturation of Emmental cheese.

(Ginzinger et al., 1999) did research on yellowness index and found it is highly correlated with b*(green/blue balance), and as the cheese ages yellowness increases. Whereas uniform goat cheese is of good quality and white in color, otherwise there is defect in manufacturing.

(Buffa et al., 2001) researched on cheese color changes, which is prepared from goat milk that is pasteurized, but raw and pressurised. A Hunter Lab spectrophotometer was used to measure the color with L*( a measure of brightness), and the a* and b* values which change red/green and green/blue balance. It turns out that there are significant color differences between various cheese types due to the milk processing and maturation time. There is no evidence of change in value of a*(red/green balance) with the aging of cheese. Whereas the value of L*, and that of b* (green/blue balance) increases.

(Marchesini et al., 2009) reported contradictions to this theory, and found that the values of L* (which is a measure of brightness), a* (which defines red/green balance), and b* (defining green/blue balance) decreases when cheese ages. After maturation time, the increase rate of water-soluble nitrogen in raw milk cheese is very high. The cheese surface develops more openings with low uniformity but protein becomes dense and brightness reduced. This is because of higher nitrogen content, but less moisture. Manufacturers add food color to make the look of cheese attractive. For example, food colors are added in cheddar cheese to make it look orange and to hide the effect of seasonal change in color of cheese (Kang et al., 2010).

(Pótorak et al., 2015) researched the relationship between fat type used for cheese making and changes in color of some selected cheese types, as the cheese sold in the Polish market used vegetable oil replacing the milk fat. The research used a colorimeter to measure the color attributes L*(which is a measure of brightness), a* (value of red/green balance),

b*(value of green/blue balance), h (value of hue), and C* (value of chroma). The values of C* (chroma) and b* (green/blue balance) was lower in cheese using vegetable oil, made with full-fat milk. Similarly, the ratio of cheese b* (value of green/blue balance), made with rapeseed oil instead of milk fat, was signicantly reduced. This is due to the fact that vegetable oil does not significantly affect the yellow intensity in cheese.

(Quevedo et al., 2008) described a method of recording the surface texture of bananas using a CVS(Computer Vision System). The texture fractal Fourier analysis determined the increase in fractal value to identify and segregate overripe bananas.

(Razak[1] et al., 2012) proposed an algorithm based method to determine the mango production grade by digital blur image processing. The accuracy achieved was over 80.00%.

(Moallem et al., 2017) proposed an Apple grading algorithm using Multi-Layer Perceptron (MLP) neural network for defect segmentation and extraction of statistical, texture, and geometric features, and achieved accuracy of 89.20% for defective apples and 92.50% for healthy apples.

(Sahu and Potdar, 2017) used an algorithm for grading mangoes based on quality. Quality Ratio (QR) threshold is used to identify defects and maturity of mangoes. The mango is rotten when QR is more than the threshold value, and good when QR value is lower than the threshold.

(Naik and Patel, 2017) used CIELAB color space algorithm and thermal images, weight, eccentricity and area of mango fruit to calculate mango size and maturity of mangoes. Accuracy achieved 89% with 2.3s intensity.

(Radojevi et al., 2011) did apple fruit analysis combining 256 grey images and using parameterization algorithms, a reliable fruit sorting methodology based on digital pattern recognition, linear fitting and numerical integration.

(Unay et al., 2011) proposed multi-category classification for the grading of apple fruit. Specific segmentation of the stem/calyx region is done using multispectral images to extract texture, geometric and statistical features from the segmented region. The accuracy reached 93.50%.

(Gopal et al., 2012) used PDF(Probability Density Function) median to classify and histogram intersection to avoid score mismatch while grading and classifying apple fruit.

(Cavallo et al., 2018) assessed the quality of packaged fresh-cut lettuce and then did classification based on nearest 3 neighbors methodology, using CIELAB color space. Minimum color distortions recorded on a selected region using a CNN(convolutional neural network) deep learning approach.

(Dale et al., 2013) studied the use of NIR-HSI(Near Infrared- Hyperspectral Imaging) in agriculture and used it for the purpose of agricultural product quality control.

(Lorente et al., 2012) investigated the literature on using HSI(Hyperspectral Imaging) to examine fruits and vegetables quality and explained different methods of acquiring images and using those to inspect the internal and external features of agricultural food items.

According to (Peri, 2006), cheese features such as physical, sensory, nutritional, cooking ability, safety and chemical are used by the computer vision system for cheese quality evaluation. The cheese texture features, which define the surface characteristics of cheese, and include (appearance, sharpness and uniformity, degree of curvature of the exposed slices, the degree of dryness or cracking, and the opacity) are most important for quality based classification of cheese. Good shredability cheese is free owing, and low shredability cheese sticks are used to form cheese balls. Low moisture and proteolysis rate, and high protein and high calcium-casein ratio give better shredability to cheese. Whereas cheese with higher water content, higher proteolysis, and low calcium, low protein to fat ratio, and longer shreds and free of fat have lower shredability.
(Ni and Gunasekaran, 1998) used computer vision methods to determine the length of shredded mozzarella cheese (Apostolopoulos and Marshall, 1994).
Image processing has developed an algorithm that does not require manual analysis of shredded cheese images to separate them and quantify morphological features to characterize the length of each shredded cheese. The developed method can successfully identify individual fragments even when fragments touch or overlap. The current fragmentation assessment method is manual, which is a very time-consuming process.
(Ni and Guansekaran, 2004) calculated the length of the shredded cheese using XY scan algorithm. The accuracy has reached 99%. XY scan works well for all shredded shapes and can use XY scan measurement data to calculate specific quality indicators and indicate the extent of free ow and the degree of debris disappearance.
Another property which determines the quality of cheese is 'Meltability' and is dened as the percentage of cheese melted and diffused in an unheated sample. The ideal meltability allows uniform melting of processed cheese (Arnott et al., 1957).
(Everard et al., 2005) studied the influence of inorganic salts, moisture, fat ratio and aging on the melting properties of cheddar cheese to assess its quality, by using computer vision methods.

Fish is considered good quality if it is fresh in appearance, and not smelling foul. These parameters can be classied according to the general appearance, smell, pigmentation, stiffness, coloring, etc. of the studied fish samples (Huss et al., 1995). The appearance has been used as a parameter for evaluating the freshness of fish. Since diagnosis is based on image processing, visually perceptible changes in parameters should be considered for analysis. Also, different body parts of fish can be used to determine its freshness. For fish freshness the segmentation method using Lab space and k-means clustering is an effective and accurate freshness classification method. The wavelet transform coefficient of the segmented ROI(Region of Interest) using the Haar filter gives the discrimination coefficient of the freshness range. In the past few years, spectral sensing and machine intelligence have been widely used for rapid non-destructive quality inspection of aquatic products (for example, prediction of the chemical properties of fish muscle) (Cheng and Sun, 2017). Data analysis algorithms such as PLSR(Partial least square regression), SVM(Support Vector Machine), LR(Linear Regression), etc., can be used as a powerful tool for quality classification, based on the freshness and nutrition prediction of sample spectral data. In terms of fish quality analysis, the combination of deep learning technology and spectroscopy technology can perform excellent quality inspections of the internal and external characteristics of fish based on the inspection results.

(Nandi et al., 2016) did grading of mangoes by using fuzzy incremental learning algorithms. Firstly features extracted, then support vector regression performed and finally grading done, accuracy close to 87% achieved.
(Arakeri et al., 2016) developed required hardware and software in two phases to grade tomatoes. Hardware captures the image to automate the movement of the fruit to the appropriate canister, and then software grades tomatoes considering their ripening and defects. Accuracy of 96.47% achieved in quality based grading of tomatoes.
(Si et al., 2017) proposed a process to measure length/width ratio of potato tubers. The comparison with manual calipers showed 96.00% accuracy. The automated system developed by (Ali and Thai, 2017) used mechanical/electrical parts and graded fruits using external quality features like decay and surface defects, thus saving on effort, time and achieving improved effciency.

## 5.2 Classification

(Mustafa et al., 2011) used Artificial Neural Network (ANN) and Digital Image Processing techniques for designing a system that will be used for sorting fruits. Color features were primarily used for feature extraction. The inspection for quality of fruits and vegetables is done most of the time using RGB (Red, Green, Blue), HSI (Hue, Saturation, Intensity) and CIELab color spaces. You can specify a color space to extract color features from an image. Images acquired through the common RGB color model are based on the original colors Red (R), Green (G), and Blue (B). This color model divides the image into red, green and blue planes and determines all color moments.
(Kondo, 2009) used a combination of maximum length, respiration, and diameter as measurements to classify apples. (Blasco et al., 2003) did that for dates. (Riyadi et al., 2007) chose eggplant and (Kondo et al., 2007) used diameters for sorting strawberries, lemon and citrus.
(Al Ohali, 2011) developed a prototype CVS for sorting dates and then grading those accordingly. As a consequence the RGB images are thus used to classify date fruit into three categories. The BPNN (back propagation neural network) classifier, when used, could accurately sort out 80.00% dates.
(Hassankhani and Navid, 2012) and (Xiaobo et al., 2007) used the same technique to sort food items such as tomatoes, potatoes, strawberries, and mangoes.
According to (Cubero et al., 2011) the food items with a peculiar or of uneven shapes are generally low priced or difficult to dispose off. Hence the shape of a food item is an essential feature to classify/grade quality fruits and vegetables. Food item convexity, roundness, tightness, length, width, elongation, boundary coding, fourier descriptor, aspect ratio and invariant moments are the most commonly used shape features used for quality analysis of produce by the food industry.
As per (D. Zhang et al., 2010) and (Sadrnia et al., 2007) when both size and shape features are used together to classify potatoes, citrus, peaches, apples, eggplants, etc, makes classication highly rened and more reliable.

(D. G. Kim et al., 2009) created a model for identification of citrus peel diseases using color texture analysis. They transformed HSI images by making use of color co-occurrence. Texture features extracted by Stepwise discriminant analysis. The discriminant function based on the generalized squared distance is used to downsize texture features.
(Zhao et al., 2009) worked to detect citrus peel diseases by making use of color and texture analysis. HSI texture features reduction by stepwise discrimination analysis provided 95% classification accuracy.
(Khojastehnazhand et al., 2010) developed an algorithm to sort tangerine using two methods. The idea was to develop a system so as to calculate tangerine volume. Two major image processing approaches were used in this paper.
(Savakar, 2012) performed visual classification of five different fruits in two steps. First is by extracting the features and then applying classification. They extracted color and texture features using an algorithm which used three different types of features viz. color, texture, and a combination of color and texture, which are then classied by a BPNN(Back Propagation Neural Network).
(D. Li et al., 2017) proposed a method that combines color, shape and texture features to solve the problem of target and background segmentation of green apple picking robots in complex backgrounds and used grayscale to extract texture features.
(Sa et al., 2016) used a DNN(Deep Neural Network) based system for classication of seven fruits.
(Y. Zhang et al., 2014) proposed a hybrid classification method using FSCBC(Fitness Scale Chaotic Articial Bee Colony) algorithm and FNN (Fully Connected Network) for fruit detection. PCA used to reduce images. Accuracy 89.10% for feature classification.
(S. Wang et al., 2015) developed a system that uses WE(Wavelet Entropy), PCA(Principal Component Analysis), FNN(Fully Connected Network) trained by FSCBC( Fitness Scale Chaotic Artificial Bee Colony) and BBO (Biogeography Based Optimization) for fruit classification. Achieved classication accuracy of 89.50% for both (WE+PCA+FSCBC-FNN) and (WE+PCA+BBO-FNN) systems.
(Dubey and Jalal, 2016) used a combination of color coherence vector, complete local binary pattern and Zernike moment to achieve 95.94% accuracy for the classification of apple fruit as diseased or healthy.
(Y. Zhang et al., 2016) used a classication method based on (Biogeography Based Optimization)BBO, FNN(Fully Connected Network), and the fine fold stratified cross-validation for fruit classification, and reported an accuracy of 89.11%.
(Zaborowicz et al., 2017) proposed an algorithm to assess the quality of greenhouse tomatoes. Use of ANN model and two digital images one of stem and another of front of tomato provided correct classification.
(Y. Zhang and Wu, 2012) used a multi-class kernel Support Vector Machine (SVM) to classify fruits. The background of images removed by using split and merge algorithms and obtaining histograms that contain the data information of extracted color, texture and shape features. They developed three types of SVMs, which were directed acyclic graph SVM, maximum winning vote SVM, and winner-takes-all SVM, and also three kernels, which were Gaussian radial basis, homogeneous polynomial, and linear. Accuracy achieved was 88.2%.
(Bulanon et al., 2009) used different images combining methods for fruit detection, namely fuzzy logic and LPT(Laplace Pyramid Transform) were used for analysis. The findings showed that the performance of fuzzy logic method was better than LPT, and fusion images by both methods improved the identification ability as compared to use of thermal images alone. SID (Spectral Information Divergence) classification performed on the hyperspectral images of grapefruits to identify cankers in reference to normal grapefruits and also on other citrus surfaces by using predetermined canker reference spectra to quantify spectral similarity. Accuracy 96.2% for SID threshold of 0.008.
(Bulanon et al., 2008) experimented with the temperature changes in the citrus canopy choosing different time durations of the day, for identifying oranges. Attributes like surface temperature, ambient temperature and relative humidity were measured over a 24 hour cycle duration. The temperature curves (from the afternoon 16:00 to midnight) of the canopy and the fruit showed a large temperature gradient.

## 5.3 Defect/Disease Detection

(Amato et al., 2012) studied the effect of different bacteria's and microorganisms on red smear cheese that is packed and stored in foil. The result showed that the defect on the cheese is not caused by a single or a microbial group of microorganisms rather it can be caused by contribution of various groups of organisms.
(Zhu et al., 2007) used a kernel PCA (Principal Component Analysis) that combines the Gabor wavelet representation of apple images with the nuclear PCA method of apple quality inspection based on near-infrared imaging. Based on geometrical features using Computer Vision apples are classified as healthy or blemished. By using the Gabor kernel PCA, the need for local feature segmentation is eliminated, and a 90.60% recognition is obtained.
(Rong et al., 2017) made use of a sliding comparison window local segmentation algorithm to segment various surface defects such as thrips scars, injury by insects, copper burns, wind scarrings etc. They were able to detect accurately 97.00% of defective oranges.
(Unay and Gosselin, 2006) proposed defect segmentation of 'Jonagold Apples'. Supervised classifiers performed better as compared to unsupervised. Several methods based on threshold classification are used for pixel segmentation of "Jonagold" apple surface defects. (Unay and Gosselin, 2006) pixel neighborhoods technique improved segmentation accuracy.
(Vijayarekha, 2008) used MIA(Multivariate Image Analysis) based on multi-directional PCA(Principal Component Analysis) method to segment defects on apple fruit. MIA grouped all pixels with the same spectral features into a cluster to easily identify the external defects.
(Dubey and Jalal, 2012) used LBP (Local Binary Pattern) for extracting texture and color features to classify and defect detection in apples. The accuracy for the classification is 93%. (Dubey and Jalal, 2012) and (Dubey and Jalal, 2013) proposed a method that used the technique of image segmentation, which was based on K-means clustering to extract features from the segmented area. A MSVM(Multi-Class Support Vector Machine) made to use those features for training and classification purposes specifically for apple fruit disease detection.

(Xing et al., 2005) used PCA to identify bruise marks using hyperspectral images on 'Golden Delicious Apple'. The 2nd and 3rd principal components of images are used to identify bruises. Accuracy of identifying non-bruised apples was 93.00%.

(Jhuria et al., 2013) detected scab and root diseases in apples using texture, color and morphological features. The best ever 90% accuracy achieved for morphological features extraction.

(Jawale and Deshmukh, 2017) made use of thermal cameras and modernized image processing for automatic detection of fruit diseases by Bruise Detection System. Use of ANN(Artificial Neural Network) and real time assessment gives sufficient speed and accuracy.

(Blasco et al., 2007) assumed that a good quality fruit surface has a sound peel. The proposed algorithm for defect detection used region oriented segmentation to detect defects in citrus fruits, and achieved 95% accuracy.

(Xiao-bo et al., 2010) used color cameras to detect defected peel from the surface of fruits and vegetables. Error for defective apples lowered to 4.2% from 21.80% using a three camera system.

(Hu et al., 2014) used K-means clustering to segment ripening bananas. The first clustering segmented banana contours, and then the banana surface damaged lesions and spots were quantified.

(J. Li et al., 2013) applied the combination of both image processing and chemo metric tools to detect defects in Oranges using spatial features extraction from hyperspectral images.

(Bennedsen et al., 2005) worked on eight varieties of apples to classify those using ANN(Artificial Neural Network) based CVS(Computer Vision System). The accuracy achieved in detecting individual surface defects was 77%-91%. Similarly, these routines were able to measure 78%-92.7% of the defective area in totality.

Bennedsen and Peterson, 2005 took multiple pictures of a rotating apple, keeping that in front of the camera and was successful in effciently eliminating the dark areas on the surface of the apple.

(Pydipati et al., 2006) referred to uninfected fruits, identified four types of citrus diseases by using methods like color co-occurrence and the generalized square distances in HSV (Hue, Saturation, Value) color space and reached an accuracy gure of more than 95%.

(D. G. Kim et al., 2009) did classification of grapefruit peel diseases by using a data set of six known grapefruit peel diseases in reference to uninfected grapefruit. To counter those diseases strong texture grapefruit is developed and discriminant analysis was used to classify those. Accuracy achieved was 96%.

(López-Garca et al., 2010) used an unsupervised approach based on the strategy of analyzing multivariate images by using PCA for detecting skin defects in citrus fruits for the generation of a reference eigenspace from a matrix deduced by evolving data related to the space and color from a defect free peel sample. In addition, the concept of multi-resolution was also introduced to fasten the process. 120 samples of citrus tested the success rate of detecting defects on an individual basis was 91.5%, and the success rate of detecting damaged samples and sound samples was 94.2%.
(Crowe and Delwiche, 1996b) used fusion of two images (NIR and pipeline). The information available using structured lighting arrangement helped to distinguish defects related to concavities. They classified defects based on the numbers of defective pixels found, and also estimated the defect area for each fruit.
(Crowe and Delwiche, 1996a) proposed use of three cameras that can perceive the reflectance in the visible spectral region and also in the narrow bands in the near-infrared spectral region in order to perform color evaluation and find out fruit defects simultaneously. The visible spectral region was used for color grading, while the narrow band centered at 780 nm was used for detection of concavity using structured illumination and the second band with center 750 nm was used to identify dark spots under complex lighting arrangements.
(Leiva-Valenzuela and Aguilera, 2013) tested different algorithms to automatically recognize blueberries damages and stem and calyx ends damages. It was found that SVM(Support Vector Machine) and LDA(Linear Discriminant Analysis) are the best classifiers. By using these classifiers, the orientation of blueberries was successfully determined in 96.8% of the cases taken up for testing. According to reports, the average outcome for mechanically damaged, shrunken and fungal rotted blueberries were 86%, 93.3% and 97%, respectively.
Fecal contamination of apples required immediate resolve as that raised concerns about food safety (M. Kim, Lefcourt, Chao, et al., 2002). A combination of PCA (Principal Component Analysis) and HRI (Hyperspectral Reflectance Imaging) was used to detect apples with fecal contamination. A MIS (Multispectral Imaging System) which could implement three visible light wavelengths and two NIR(Near-Infrared) wavelengths, used to identify apples with fecal contamination. (M. Kim, Lefcourt, Chen, et al., 2002) showed use of multispectral fluorescence to recognize fecal contaminated apples.
(Pujari, Yakkundimath, and Byadgi, 2013) and (Pujari, Yakkundimath, and Byadgi, 2013) used BPNN(Back propagation neural network) classifiers based on color and texture features in RGB and YCbCr(Luminance, Chroma Blue, Chroma Red) color spaces to identify diseased fruits, which were classified according to grades determined in reference to normal fruits. A success rate of approximately 88% achieved using this approach.
(Kleynen et al., 2005) created a MVS (Multispectral Vision System) with four bands in the visible/NIR range. The defects in fruits classified into four categories: mild, more serious, leading to rejection and recent bruising. They made use of pattern matching algorithms to detect defects of stem ends and calyxes. Pixel classification methodologies were used, and Bayes theorem based non-parametric models used to segregate defective and non-defective fruits. A good classification rate achieved for apples having serious defects and recently bruised surfaced apples.
(Leemans et al., 1998) proposed a method to detect defects in "Golden Delicious" apples based on their surface color information data. The first step was generating a model based on the variation from normal color, next, a comparison is made between each pixel of the image with the pixels of the model for segmentation of defects. A pixel is considered healthy tissue if that matches with the model pixel.
(Leemans et al., 1999) proposed a Bayesian classification based segmentation method, which used the information contained in the two-color apple image. This process could divide most of the defects (such as bruises, pits, fungal attack, scar tissue, frost damage, scab and insect attack) into several parts.
(R. Lu, 2003) explored NIR-HSI (spectral region 900 to 1700 nm) technique to detect apple bruises. The use of the NIR-HSI system made it possible to detect bruises, old and new, on the surface of apples. It was found that the combination of the ratio method and the R and G components, coupled with the large area and slender area removal algorithm, can be used to effectively distinguish stem ends from the defects.
(Mehl et al., 2002) used a multispectral approach and HSI to nd defects in three apple varieties: The technologies involved (a) MIS(Multispectral Imaging System) from HSI analysis to recognize the spectral characteristics of apples for distinct selected filters, and (b) multispectral imaging to quickly detect contaminated apples. Using these techniques, it was found that the separation effect between the normal and contaminated Golden Delicious and Gala varieties is very distinct, while the separation effect of the Red Delicious variety is limited.
(Q. Li et al., 2002) came out with an experimental hardware system, which used computer imaging technology to sort out defective apple surfaces. The hardware system could simultaneously check each apple from the four sides on the sorting line. Techniques also developed for removal of background from an image, defect segmentation, and stem end and calyx region recognition.
(Lopez et al., 2011) proposed a CVS that can detect defects and can also classify the types of defects in the citrus family. Results were promising.
(Ouyang et al., 2012) designed a synthetic segmentation algorithm for real-time online segmentation of greenhouse strawberry diseases. Eigenvalues are extracted and normalized and then used to train SVMs, BPNNs using eigenvectors of the sample. The results show higher recognition accuracy for SVMs than BPNNs.
(Panli, 2012) created a method using mathematical morphology to segment stems. The method is based on the phase of Fourier transform and applies the attention selection model to extract the fruit saliency map to detect defects on the surface of the fruit. Also, examination of the still X-ray images on the computer screen is an acceptable method of identication of infected apples: 50% of defective apples were identified, and 5% of good apples were classified as defective apples. The cheese gets defective because of ageing, bad quality of milk used in manufacturing or high moisture content or due to formation of excessive eye holes and slits etc. In the case of cheese, factors such as cheese aging, production of CO2 and N2 gases form surface openings called ‘eye holes’ or ‘stomata’, which are usually round in shape and shiny in appearance. Their size, number, distribution and shape are considered in classification and quality assessment of cheese. Clostridium or excessive CO2 production can cause negative effects on cheese quality.
(Le Bourhis et al., 2007) reported that clostridium causes poor eyeball formation, white spots and the rancid smell in Swiss cheese. In the late stage of maturation (after a warm room), Swiss-type cheese produces excessive eyes, cracks and produces excessive CO2, which is called late or secondary fermentation. Undesirable gas generation can cause cuts, cracks, or a loose packaging appearance, which is a problem with cheddar cheese (Mullan, 2000).

(Fox et al., 2017) pointed out that one of the least controlled aws in round-eyed cheese is the kerf in the cold storage after the cheese is taken out of the warm room.

(Caccamo et al., 2004) measured cheese slice area, thickness, and stomata percent present in the total area of the image to reason out abnormal Emmental cheese structure.

(Melilli et al., 2004) worked on salted pasta-filata and Ragusano cheese to detect early defects due to gas. In cheese made with raw milk in brine, early formation of gas is usually due to poor milk quality, poor hygiene during the cheese making process, slow acid production during the production process, and slow absorption of salt from the brine. Coliform bacteria are the main cause of early gas production in raw milk cheese (Bintsis and Papademas, 2002), as raw milk and ambient gas enter the cheese during cheese making, while during curing of cheese salt gets absorbed slowly causing many small holes due to early gas formation. While coliform bacteria are responsible for early gas production defects, Clostridium tyrosines bacterias cause late gas production (Thylin et al., 1995).

(Melilli et al., 2004) experimented on Tofu pre-curing in presence of brine solution (18% brine vs saturated brine) at three different temperatures of 12°C, 15°C and 18°C, to study effect on early gas formation and changes in number of coliforms responsible for Ragusano cheese defects. The results showed that the three-way interaction between salt, stretching temperature and curd pH can greatly reduce the survival rate of coliform bacteria, thus reducing their numbers significantly in pre-cured Ragusano cheese. Cheese gap defect is basically due to formation of slits in cheese. It causes cracks in the main body of cheese that may be longer than 3.5 cm. Trapped air, cold storage, and abnormal gas production are the major reasons for slit formation. The air remaining during the compression of the curd is believed to be responsible for the open texture. Slits may also appear during refrigeration. Calcium and lactate ions supersaturation in the serum phase of cheese forms crystals of calcium lactate (Rajbhandari and Kindstedt, 2005) (Rajbhandari and Kindstedt, 2008). These calcium lactate crystals will appear as white spots or mists on the surface of cheddar cheese. Though harmless, they are a major quality problem for cheddar producers (Swearingen et al., 2004).

(Zabaleta et al., 2016) did extensive study on defects like eyes, paste color, peel, flavor, texture and shape on semi-hard raw goat milk cheese, and raw sheep milk cheese. One of the cheese defects investigated in their paper is that the peel is too halo, which is described as too dark or too wide mushy borders. The results obtained show that the medium and long-distance cheese exhibits a higher percentage of the skin halo, and the paste-like color, animal flavor and skin traces are deeper. The browning characteristic is an assessment of the overall color of the cheese after baking the pizza. Desirable characteristic is mild browning of cheese, but if browning is excessive then it is a defect, which is not acceptable to the consumer. Browning is considered a defect of processed cheese, whereas it is considered a desirable characteristic when used as cheese filling (Matzdorf et al., 1994).

According to (Kindstedt and Rippe, 1990) and (Rudan and Barbano, 1998) the property of cheese due to which it removes oil on heating is called ‘oiling off’. Cheese oiliness, which gives the pizza a shiny appearance, can be a quality topping, whereas that oiliness when found in processed cheese is considered undesirable, an unacceptable defect, though a moderate release of free oil from the cheese on heating during most cooking applications is considered ideal.
(Gowen et al., 2009) reported that starch, citric acid and salicylic acid are the impurities which are mixed in different types of fresh cheese. They used HSI to detect these impurities (Burger and Geladi, 2006b) (Burger and Geladi, 2006a).

# 6 Experiments and results

Table 5. Comparison of different features for quality analysis of food

| S.No. | Author(s) | Food Type | Parameters | Technique | Accuracy/Findings |
|---|---|---|---|---|---|
| 1. | Pereira et al., 2018 | Papaya | Ripening by color | RGB | 94.30% |
| 2. | Fox et al., 2017 | Cheddar cheese | gas defect | CVS | texture gas holes |
| 3. | Zhang et al. (2014a,b,c,d) | Potato | Irregularity evaluation | Fourier descriptors | 98.10% |
| 4. | Pótorak et al., 2015 | Cheese | effect of fat on color | Colorimetry | |
| 5. | Zhang et al. (2012) | | Shape grading | Fourier descriptors | 95.24% |
| 6. | Y. Zhang and Wu, 2012 | Pear | Physical properties | Depends on size | 88.20% |
| 7. | Al Ohali, 2011 | Date | Grading based on external quality | Fourier descriptor | 80.00% |
| 8. | Xiao-bo et al., 2010 | Apple | Grading by color | RGB and HSI | _ |
| 9. | Marchesini et al., 2009 | Asiago cheese | Ageing effect on color | Colorimetry | _ |
| 10. | Blasco et al.,2009 | Pomegranate | Grading by color | RGB | 90.00% |
| 11. | Gowen et al., 2009 | cheese | quality control | HSI | _ |
| 12. | Kondo, 2009 | | Grading by external quality | Deformability, Complexity,Roundness | _ |
| 13. | Le Bourhis et al., 2007 | Emmental cheese | butyric fermentation | CVS | _ |
| 14. | Blasco et al., 2007 | Citrus | Inspection | Maximum/Mini wavelet | 98.00% |
| 15. | Sadrnia et al., 2007 | Watermelon | Classification based on shape | Length to width ratio and fruit area to background area | _ |

| S.No. | Author(s) | Food Type | Parameters | Technique | Accuracy/Findings |
|---|---|---|---|---|---|
| 16. | Everard et al., 2005 | cheddar cheese | factors influencing melting | CVS | _ |
| 17. | Caccamo et al., 2004 | swiss cheese | measurement of area of gas holes | RGB channel | _ |
| 18. | Melilli et al., 2004 | filata cheese | early gas formation | image analysis poor milk quality | _ |
| 19. | Ni and Gunasekaran.,2004 | shredded cheese | length measured | X-Y scan | 99% |
| 20. | Zhu et al., 2007 | apple | quality grading | kernel PCA. Gabor wavelet images nuclear PCA method near-infrared imaging | 90.60% |
| 21. | QQuevedo et al., 2008 | banana | overripe recognition | surface texture CVS | _ |
| 22. | D. G. Kim et al., 2009 | | transformed HSI images Texture Features extracted | color co-occurrence Stepwise discriminant analysis generalized squared distance used. | _ |
| 23. | Zhao et al., 2009 | citrus | peel diseases classification | color and texture analysis.HSI texture features reduction.Stepwise discrimination analysis | 95% |

| S.No. | Author(s) | Food Type | Parameters | Technique | Accuracy/Findings |
|---|---|---|---|---|---|
| 24. | Khojastehnazhand et al., 2010 | | classify and to sort | size and color | 94.04% |
| 25. | Savakar, 2012 | | extract color and texture features | BPNN(Back Propagation Neural Network) digital blur image processing | 80% |
| 26. | Eazak et al.,2012 | mango | Grading | digital blur image processing | 80% |
| 27. | D. Li et al., 2017 | Apple | extract texture features for robot picking | color, shape and texture features arget and background segmentations | 92.5% |
| 28. | Moallem et al., 2017 | Apple | Grading | Multi-Layer Perceptron (MLP) neural network | _ |
| 29. | Sahu and Potdar, 2017 | Mango | Grading | Quality ratio threshold | _ |
| 30. | Naik and Patel, 2017 | Mango | Maturity | Quality ratio threshold | 89% |
| 31. | Rong et al., 2017 | oranges | grading defects | Segmentation | 97% |
| 32. | Unay and Gosselin, 2006 | Apple | Feature extraction | ANN base Segmentation | _ |
| 33. | Radojevi et al., 2011 | Apple | Analysis classification | parameterization algorithms digital pattern recognition,DNN | _ |
| 34. | InKyuSa et al., (2016) | 7 fruits | Classification | DNN | _ |

| S.No. | Author(s) | Food Type | Parameters | Technique | Accuracy/Findings |
|---|---|---|---|---|---|
| 35. | Unay et al., 2011 | Apples | multicategory classification | Segmentation of calyx | 93.5% |
| 36. | gopal2012classification | Fruits | Classification mismatch | PDF median-classify. Histogram intersection | _ |
| 37. | Dubey and Jalal, 2012 | Apples | Classification | Local Binary Pattern | _ |
| 38. | Y. Zhang et al., 2014 | | Feature Classification | FSCBC, FNN, PCA | 89.1% |
| 39. | S. Wang et al., 2015 | | Classification | Wavelet Entropy | 89.5% |
| 40. | Dubey and Jalal, 2015 | | Classification | CCV and CLBP and Zernike moment | 95.94% |
| 41. | Y. Zhang et al., 2016 | | Classification | BBO, FNN | 89.11% |
| 42. | Zaborowicz et al., 2017 | Tomatoes | Quality classification | ANN, images stem and front | _ |
| 43. | Cavallo et al., 2018 | packaged lettuce | quality classification | CIElab, CNN | _ |
| 44. | Dubey and Jalal, 2012 | | | training classification | segmentation, K-means, Clustering, SVM |
| 45. | Unay and Gosselin, 2006 | Jonagold apple | threshold classification | pixel neighborhoods, Segmentation Technique, 3SVMs+3kernels | 88.2% |
| 46. | Y. Zhang and Wu, 2012 | | background removal | 3 SVMs and 3 kernels | 88.2% |

| S.No. | Author(s) | Food Type | Parameters | Technique | Accuracy/Findings |
|---|---|---|---|---|---|
| 47. | Bulanon et al., 2009 | grapefruits | classification | SID(Spectral Information Divergence) Threshold .008 | 96.2% |
| 48. | Bulanon et al., 2008 | orange | identifying | Temperature gradient | _ |
| 49. | Dale et al., 2013 | agro products | quality control | NIR-HSI | _ |
| 50. | Lorente et al., 2012 | agro products | inspect internal/external features | HSI | _ |
| 51. | Peri, 2006 | cheese | shredability | moisture analysis | _ |
| 52. | Ni and Gunasekaran, 1998 | Mozzarella cheese | length of shreds | computer vision | _ |
| 53. | Apostolopoulos and Marshall, 1994 | shredded cheese | separating shreds | algorithm | _ |
| 54. | Ni and Guansekaran, 2004 | shredded cheese | length of shreds | XY scan | 99% |
| 55. | Everard et al., 2005 | cheddar cheese | effect on meltability | computer vision | _ |
| 56. | Huss et al., 1995 | fish | freshness | CVS to study | _ |
| 57. | Cheng and Sun, 2017 | fish | quality inspection | spectral sensing | _ |
| 58. | (Xing et al., 2005 | apple | identify non-bruised | 2nd, 3rd PCA | 93% |
| 59. | Nandi et al., 2016 | mangoes | quality grading | fuzzy incremental learning algorithms. | 87% |
| 60. | Jawale and Deshmukh, 2017 | fruits | disease detection | ANN- bruise detection | _ |
| 61. | Blasco et al., 2007 | citrus fruits | detect peel defects | region oriented segmentation | 95% |

| S.No. | Author(s) | Food Type | Parameters | Technique | Accuracy/Findings |
|---|---|---|---|---|---|
| 62. | Xiaobo et al., 2007 | apples | defective error | 3 camera system | error 4.2% |
| 63. | Hu et al., 2014 | banana | ripening- spot detection | clustering segmentation | _ |
| 64. | Arakeri et al., 2016 | tomatoes | quality grading | custom h/w, s/w | 96.47% |
| 65. | Si et al., 2017 | | tubers length/width ratio | customised process | 96% |
| 66. | Bennedsen et al., 2005 | apples-8 types | surface defects | ANN based CVS | 77% to 91% |
| 67. | Pydipati et al., 2006 | citrus fruits | diseases | color co-occurrence and generalized square distances in HSV color spaces | _ |
| 68. | D. G. Kim et al., 2009 | grapefruit | peel disease based classification | discrimination analysis | 96% |
| 69. | López-Garca et al., 2010 | citrus fruits | peel disease defects | PCA for multivariate images | 91.5% |
| 70. | Leiva-Valenzuela and Aguilera, 2013 | blueberries | stem/calyx end damages | SVM, LDA classifiers | 96.8% |
| 71. | Le Bourhis et al., 2007 | swiss cheese | excessive eyes | clostridium study for excessive gas production | _ |
| 72. | Fröhlich-Wyder et al., 2002 | cheddar cheese | cuts/cracks | undesirable gas generation study | _ |
| 73. | Caccamo et al., 2004 | Emmental cheese | abnormal structure | area measurement of stomata | _ |

| S.No. | Author(s) | Food Type | Parameters | Technique | Accuracy/Findings |
|---|---|---|---|---|---|
| 74. | Bintsis and Papademas, 2002 | cheese | early gas production | study on Coliform bacteria for gas defect | – |
| 75. | Rajbhandari and Kindstedt, 2005 | cheese | calcium lactate | study of supersaturation | – |
| 76. | Gowen et al., 2009 | fresh cheese | detection of impurities | HSI for detection | – |

# 7 Conclusion

This review paper focuses on the use of Image Processing Analysis, Computer Vision Technology and Deep Learning System (Neural Networks, AI/ML) for classification and grading of natural agricultural products such as fruits and vegetables, the aquatic food fish and the processed food cheese. The most important quality characteristics of agricultural products are their sensory features of appearance such as size, color, shape, texture and their diseases and defects. As an alternative to manual food inspection, the Computer Vision Systems, and Deep Learning based techniques and algorithms provide genuine, impartial and non-destructive assessment of food items. In this review paper a sincere attempt has been made to explore and compare various methods and algorithms proposed by different researchers. Although a number of researchers have proposed various methods for food quality inspection, grading and freshness, there is a need for a robust and versatile automated system, which provides improved performance, free from regional bias and able to grade/sort multiple food items based on quality, freshness and also able to detect the defects.

# Part II

# GuavaNet: A deep neural network architecture for automatic sensory evaluation to predict degree of acceptability for guava by a consumer.

## 8 Abstract

The thesis introduces to a deep neural network architecture based on transfer learning approach. This deep neural network architecture serves as an end-to-end framework that can predict the degree of acceptability by the consumer for a guava based on sensory evaluations. The architecture presented in this thesis consists of an object detector based on a YOLOv4 pre-trained on COCO dataset (Lin et al., 2015) that can localize guavas in a picture and the output from this deep neural network serves as an input to another deep neural network architecture which consists of a VGG-16 network as a base and custom Fully Connected Network (FCN) layer fine-tuned using a hybrid transfer learning approach by utilizing the architecture as feature extractor and then fine-tuning (Takhirov, 2017). Model selection is based on comparison of the performance of architectures such as VGG-16 as base and custom FCN layer, ResNet-18 as a base and custom FCN layer and ResNet-50 as a base and custom FCN layer trained as a feature extractor.
**Keywords:** YOLOv4,VGG-16, ResNet-18, ResNet-50

## 9 Introduction

Since ancient times, the people started showing utmost concern for healthy eating habits. One of the main constituents of which was eating fresh foods. With the evolution of human societies the concept of a healthy diet came into vogue, people started looking for foods which can supplement essential nutrients like minerals, vitamins, fiber etc., required for keeping a good health. Research done on various food items found that a diet rich in fruits and vegetables is most healthy and helps in reducing the risk of cancer and many other chronic diseases. This led to increased consumption of fresh fruits and vegetables, as they are a rich source of essential nutrients. Food is the primary source of energy that helps in growth, provides energy for activities, repair and other body functions. Most of us tend to repeatedly buy high quality food brands that meet our expectations. Similarly, even a small incident of compromised food quality can damage brand image, and company's profits may collapse. Thus appropriate quality control measures are necessary for food brand management. Quality control through effective inspection, and control of production processes and operations eliminate inferior goods and wastage reducing the cost of production and rise in goodwill among consumers. The company will also gain a good brand reputation increasing its chances of surviving in a highly competitive market. This is a great help in attracting more customers to use their products, and increasing their sales. Therefore, for the well-being and safety of public consumers and the relationship between food quality and brand awareness, food safety and quality inspections is essential. The precise detection applications of modern technology can help companies provide quality assurance consistently, because these technologies capture what is impossible for the human eye and conventional detection media to do. Although modern food inspection technologies are very effective in solving public health and safety issues, correct and successful inspection is also a key element in establishing and improving a company's brand image. The analysis of their being fresh or not fresh along with their nutritional content and the process employed in picking, storage or their making (processed food) help in determining whether the food is good for health or not. Therefore, the need for automatic prediction of consumer likability and freshness linked to consumer's likability and acceptability of food is the need of an hour and 'Deep Learning' techniques for image processing can provide the best techniques for food classification and quality determination.

Having said that, in this thesis we would be concentrating on Guava detection and grading the consumer likeability of the Guava based on sensory evaluation. Specifically, we would be focusing on guava, scientifically known as *Psidium Guajava*. Reason for choosing guava for this research work was due to the fact that guava remained unpopular and neglected as the majority preferred fruits like apples, bananas, oranges etc, even researchers neglected the research work on this fruit. As of late it gets the recognition of 'Super-food' due to its nutritional richness and enormous health benefits, considering it highly relevant to develop a system to determine its quality, freshness and consumer likeability. Guava is majorly grown or produced in tropical and subtropical regions of the world. It is said to have originated from Central America or Mexico and is profusely grown in China, India, Bangladesh, Nigeria, Thailand and Philippines due to its health benefits. The surface of the fruit can be hard and green or pale yellow with the white or pink inner core. The fruit can be used in salads or can be eaten by itself. Not only is the fruit flavorful but also carries great nutritional values. According to (Mariya et al., 2020) guava is highly rich in antioxidants, potassium, vitamin c and fibre. In fact vitamin c content in guava is almost double the vitamin c content of a regular orange. Since vitamin c is extremely beneficial for immune health therefore Guava is considered very important in terms of providing health benefits (Hoyt, 2019).

According to (Molidor, 2019) almost 40 percent of the food which has been produced is wasted in the United States and the giant retailers and restaurants account for this wastage.

Moreover, primary places of fruits and vegetables purchase include the giant retail supermarkets like Walmart, Whole Foods etc. as reported by (Quest, 2019), if we focus on food waste Statistics of Retail then it shows that almost 43 billion pounds or 10 percent of total food present in the grocery will never get picked up. Apart from this almost 30% of trash in grocery stores comprises food waste.

Since consumers prefer to choose fruits and vegetables from fresh supplies and from a large variety of sizes and qualities according to their personal preference, good condition and of desirable flavor merely by casual examination. Their experience is the most reliable guideline for them in choosing the fresh and quality product, with the balance between price and good quality. Generally consumers do not buy produce that is either bruised or damaged as bacteria attacks those places and then spreads rapidly to spoil the rest of the produce. Therefore food quality depends on the consumer acceptability of the product, which according to the common people is generally decided on the basis of sensory characteristics of food, which are texture, taste, aroma, and appearance. For them the appearance has to be the first attraction, and color feature is the most important feature for appearance. The consumers get attracted initially because of the color and prefer vibrant colored fruits, thus the very first sensory feature that affects customers' rejection or choice of fruit is color. It is an indirect measurement of quality attributes like freshness, desirability and diversity, maturity and safety (Pathare et al., 2013). Many consumers, who are price conscious also select vibrant color fruits with clean texture (no blemishes, no marks, no cuts on surface) based on the size to save on money and make economical purchases without compromising on quality, so the goal for them is to get more quantity in less price. So with this thought process of the consumer, retailers constantly face the liability of loss from wastage and spoilage of perishable commodities. A deep neural network architecture such as proposed GuavaNet, by predicting freshness and consumer likeability based on sensory evaluations will help retailers in reducing these losses and also satisfy the quality needs of the consumer at reasonable costs (Fruit and Branch, 1952) (First, 2016).

# 10 Motivation

"As Aristotle realized, there is a difference between the pleasures of the moment (hedonia), and the satisfaction that comes from constantly developing and living ones life to the fullest (eudaimonia)." This quote and the greater sense of purpose and personal growth associated with eudaimonia, which correlate with lower cortisol levels, better immune function, and more efficient sleep, turned out to be the goal of my research work, that is to do something 'Novel', that too with the intention of common good.

I gave very serious thought on various topics for research ranging from finance to technology to ...., but most of the ideas faltered, I failed to compromise the motive of"greater sense of purpose with personal growth goal", because the missing link was common good. The onset of the pandemic last year strengthened further my 'common good' attribute, I wanted to include in my research work. The pandemic made me learn the importance of a healthy lifestyle. As we know that we all have encounters with food every day, as it is the essential constituent for life. Therefore the idea of healthy food struck me. The next step was to search for a novelty. After reviewing many researches on food, I found that till date no work is done on acceptability index of food, specially no research has been conducted on a common man's fruit guava. I found this idea matching with the above motive of "greater sense of purpose and my personal growth", hence I chose this as my research topic.

# 11 Related Work

This section gives an overview of the Deep Learning methods by which we can achieve state-of-the-art results on challenging computer vision problems of image based classification, object detection such as fruit recognition and classification as the human can do with visual inspection. Computer Vision seeks to understand and classify digital images by means of computer. The main idea is to automate the job that human vision does. For this process to take over, computer vision tends to understand, process, analyze the digital images data from the real world to generate information. We will explore state-of-the art neural networks which can be trained with visual wisdom for computer vision tasks, and can help in finding possible solutions to define the fruit freshness and grade consumer acceptance of fruit, the way humans do in real life through sensory evaluation.

The proposed work's aim is to realize a method for embedding human knowledge into deep neural networks. Food recognition and classification is an important task to help human beings in selecting fresh and high quality food products. Images of food are one of the most important pieces of information to reflect the characteristics of food. Moreover, image sensing is a relatively easy and low-cost information acquisition tool for food appearance analysis. For natural products like fruits, the large variations in shape, texture, and color make food recognition a challenging task. Various backgrounds and layouts of food stuff also introduce variations for fruit recognition and classification.

At present, due to the common use of CNN, image analysis has been the most commonly used pattern in food recognition and classification. CNN (Convolutional Neural Networks) are being used widely in agro based industries for food classifications based on category discrimination and identification of ingredients. Most popular image processing CNN architectures are AlexNet (Alex Krizhevsky, 2012), a Visual Geometry Group (VGG) network that uses iterative units, ResNet (Residual Neural Network), and parallel data channels such as VGG (Karen Simonyan, 2014) and GoogLeNet (Szegedy et al., 2015) and those constructed by residual blocks like ResNet (He et al., 2016). All these network architectures, with pre-trained weights like ImageNet (Deng et al., 2009), can be downloaded from 'Model Zoo'. As all these pre-trained models can extract image features like color, texture, high-level abstract representations etc., thus can be used with own created specific image datasets for transfer learning by FineTuning, i.e retraining with their weights for final classification, keeping the existing weights of the convolution layer or adjusting the weights of the whole network slightly. This methodology shortened the training time and provided more accurate results. Deep learning can effectively solve the food/nonfood discrimination, a binary classification issue.
Food-5K, a database created by (Ashutosh Singla, 2016) contains 2500 food images (which are selected from three image sets viz. Food-101, UECFood-100 and UECFood-256) and 2500 nonfood images. When fine-tuned with GoogLeNet model, an accuracy of 99.2% is achieved. (X. Zhang et al., 2018) could achieve 98.7% for binary classification, and (McAllister et al., 2018) highest were 99.4% for validation dataset and 98.8% for evaluation dataset through RBF (Radial Basis FunctionRBF), a kernel-based SVM with ResNet-152. (Ragusa et al., 2016) achieved 94.86% by coupling fine-tuned AlexNet with a binary SVM classifier using a database of 3583 food images (from UNICT-FD889) and 8005 nonfood images (from Flickr).

After discriminating food from non food items, the problem reduced to a multi classification problem. Most of the researches used freely accessible different categories based large food images sets like UECFood-256 (Kawano and Yanai, 2014), Food-101 (Bossard et al., 2014), UECFood-100 (Matsuda et al., 2012) etc. to train a classifier and evaluate the trained model like DNN(Deep Neural Network) model for food recognition, while in some of researches experiments done on image datasets created from public datasets. The Food-101 database created by (Bossard et al., 2014) includes 101 food classes with 1000 images in each class. 50.76% classification accuracy achieved by using traditional ML methods. The two most used evaluation indicators in food classification are:
a) Top-1 accuracy i.e. Top-1%: This takes the largest classified group as the predicted result. If the most probable classified item and the predicted result matches, the prediction is correct, if not then wrong. b) Top-5 accuracy i.e. Top-5%.: It takes top five classifications as the largest probability vectors. The prediction is true if correct probability occurs, if not, then it is wrong. (Tatsuma and Aono, 2016) used covariances of features of trained CNN as representation of the images for classification, and achieved mean accuracy 58.65%. (Kawano and Yanai, 2014) used fine-tuned AlexNet, accuracy 70.41% for Top-1. (C. Liu et al., 2016) presented the ‘Deep Food’ network approach and achieved accuracy of 77.40% for Top-1% and 93.70% for Top-5%. (Fu et al., 2017) used a fine-tuned deep 50-layer ResNet and had accuracy 78.5% for Top-1% and 94.1% for Top-5%. Thus results of CNN models using the Food-101 dataset were better than the traditional approach. (Ciocca et al., 2017) contributed a large data set Food-527 and (Ciocca et al., 2018) also provided Food-475, they used the ResNet-50 model for classification. Best achievement on Food-527, Food-475, Food-50 and VIREO. They analysed that functions learned on Food-475, the largest food image database, give better results as compared to other smaller datasets. Thus concluded that a data set having more food images is better for food recognition algorithms.

(Martinel et al., 2018) to improve the accuracy of food identification did not consider specific characteristics of food images, but developed a slice convolution unit to extract common features of food, then added deep residual blocks to determine classification scores. Results Top-1 89.58%, Top-5 99.23% for Food-101, for UECFood-256 (90.27%, 98.71%), and for UECFood-100 (83.15%, 95.45%) respectively. (Heravi et al., 2017) designed AlexNet, used a data set containing 1316 images of 13 food categories, accuracy 95%. (Mezgec and Koroui Seljak, 2017) developed NutriNet to identify food and beverages. The CNN model used was trained on a training dataset having 225953 food and beverage images based on AlexNet architecture, and tested on a detection dataset having 130517 images. Training set accuracy 86.72% and test set accuracy 94.47% respectively.
(Fu et al., 2017) created ChinFood1000 database, used ResNet, Accuracy top-1 44.10% and top-5 68.40%. (Herruzo et al., 2016) evaluated CNN-based classifiers for Catalan food based on Catalan cuisine, a dataset 'FoodCAT', another database used is Food-101. Experiments have shown that GoogLeNet tuned on Food-101 and FoodCAT and processed with ultra-high resolution methods, recognized dishes (68.07% of the top 1 and 89.53% of top 5), and food category recognition (Top-1 72.29% and Top-5 97.07%). (Pandey et al., 2017) developed a multi-layer CNN to identify food, used two image databases - Food-101 and Indian Food Database with 50 categories, 100 images per category. The algorithm uses AlexNet as the baseline for deep CNN, and a multi-layer CNN pipeline to combine three different subnets. Excellent prediction results: For Food-101, Top-1-72.12%, Top-5-91.61%, Top-10- 95.95%. For Indian food database 94.40%, 97.60%. and 73.50% for all levels in two databases. The proposed integrated net is better than the CNN model. (Heravi et al., 2017) provided a new idea to transfer knowledge from a compressed GoogLeNet architecture as a trainer to a simple model trainee CNN with fewer parameters than the trainer CNN, and ensuring that it works faster than the trainer CNN. The trainee CNN is expected to provide the same classification scores as the trainer CNN. Therefore, this is a function approximation problem, and not a classification problem. The trainees were trained to approximate trainers with unlabeled non-food images, and then fine-tuned using the labeled food database for food classification. The proposed method achieves 62% of Top-1% on UECFood-256. Although performance is not very good, this shows that knowledge transfer methods can train a simple network with lower memory consumption.

Evaluating fruit freshness using different methodologies are carried out in past. One such Literature review is by (Moallem et al., 2017) wherein grading of golden delicious apple using K-NN and SVM classifiers is conducted. Initially the apple grading involved classifying apple as healthy or defected. Then the rank was given based on quality like first class, second class and rejected. SVM classifiers provided the best results. Also (Dubey and Jalal, 2016) classification of apple was based on three features color, texture and shape. K-means clustering was used to to detect the part of the apple which is infected and then shape, color and texture is evaluated to classify it as infected or healthy using MSVM(Multi-Class Support Vector Machine). (N.-V. Lu et al., 2020) performed a sensory evaluation on Saigon Beer to determine the beer quality during its production process . The features selection was based on correlation and was then integrated with Machine Learning techniques like Linear Regression, random forest, SVM and multi-layer perceptron. According to (Munkevik et al., 2007) when sensory evaluation of meals using computer vision technique was done, the ANN(Artificial Neural Network) classifiers provided the best result. These results were then compared to those results earlier given by human vision sensory panel which included 72 members and it was observed the new system performed significantly well.

Due to the need for deep learning on massive amounts of data, food images from the Internet or open-access food databases are always the first choice for training models. Next step is image processing (such as normalization, resizing) to reduce the interference caused by uneven lighting, inconsistent resolution etc. If the data set is not large enough, data expansion should be performed to enlarge the data set by random cropping, rotation, and flipping to simulate shooting from different angles. Generally, experiments based on large open data sets do not consider data expansion. Then, the prepared data set is divided into a training (or calibration) set for training the network, a validation set for fitting hyper-parameters, and an evaluation (or test) set for confirming the predictive ability of the model. After the data set is ready, the training task can be executed as the second step. Most researchers used pre-trained CNN to directly fine-tune their model for classification. Among the research discussed in this section, the method of modifying the model structure mainly focuses on the function of the combined mode. The feature combinations of different data sets extracted from different architectures are generated into the final feature map for classification. Top-1% and Top-5% are the most commonly used indicators for performance evaluation. Le used 75% of the database for training and 25% for testing. In future, more sensory information about food such as smell and weight can be considered as supplements to further improve the recognition accuracy.

# 12 Artificial Neural Networks

## 12.1 Feed-Forward Neural Networks

### 12.1.1 Introduction

In this chapter we will introduce with some key concepts about Feed-Forward Neural Networks and provide an overview of its functioning. As our proposed model is around a Feed-Forward Neural Network, it is important for us to understand it before we further delve into literature of the research.

### 12.1.2 Simple Perceptron

Just like brain has neurons, every neural network also has neurons. The neurons can exist in one layer or in multiple layers. In an artificial neural network a perceptron is a simple model of a neuron and is either a single unit or simply neuron which performs all the computations.

The perceptron can accept multiple inputs and provides a single set of outputs after several calculations consisting of simple matrix multiplication of the inputs and their corresponding weights. Example: When classifying two classes, it returns a binary value as output. The values so obtained are summed up and a bias $b$ is added to that value as shown in the figure below.

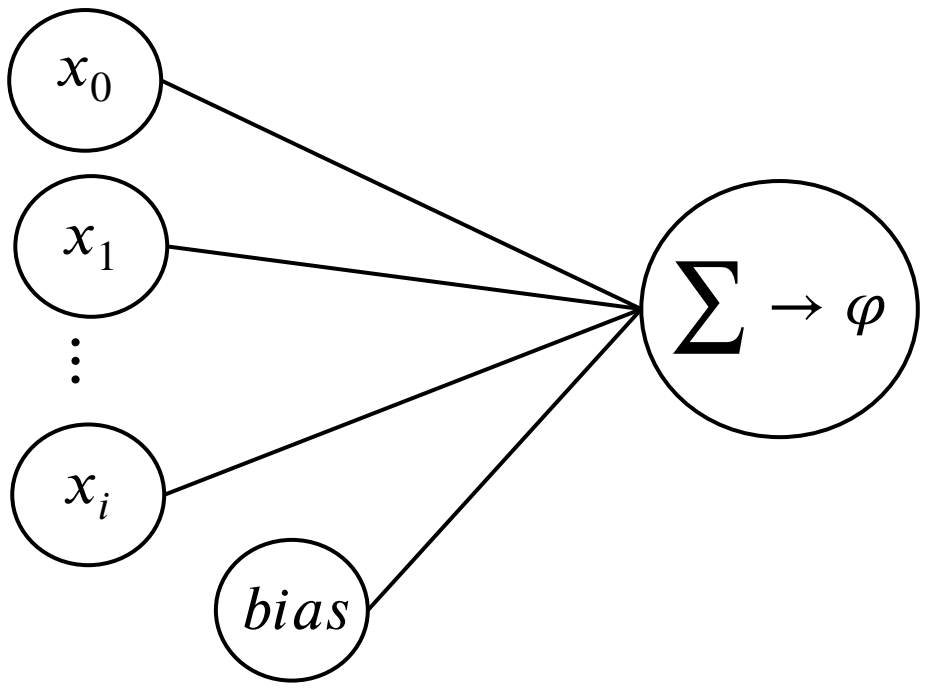

Figure 7. A Simple Perceptron

$$\sum_j w_j x_j + b \tag{11}$$

An activation function ($\varphi$) then determines the final output of the perceptron unit in the forward pass.

$$\varphi(x) = \begin{cases} 1 & \text{if } x < 0.5 \\ 0 & \text{if } x \geq 0.5 \end{cases} \tag{12}$$

Random initialization based on the multiplication value of parameters shown in bracket ((Actual output - Desired output) x learning rate), is done for the weights and bias after each epoch, which are then updated consequently to be used in backward pass. This methodology results in improved output results. Since a 'Unit Step Function' is used as an activation function, the output will get activated whenever it is above a set threshold, thus providing a 0 (zero), 1 (one) binary classifier (Den Bakker, 2017).

### 12.1.3 Single Layer Neural Network

For all practical purposes a single layer neural network can be considered equivalent to a single layer perceptron, thus can also work as a linear binary classifier. However, in the case of a single layer neural network, neurons are involved in the computations, which in fact are not single like a perceptron, but multiple units, that is therefore the only difference between a single layer neural network and a perceptron. By adding more units we will be able to solve more problems. The units are stacked in single layer, also called as the hidden layer. Each unit perform computations separately. In a single-layer neural network, the units are stacked in a hidden layer, hence the name hidden unit. Therefore, these hidden units in the hidden layer are considered as inputs to the single layer neural network, and the output layer acts as a perceptron. These are the reasons why single-layer neural networks do not require advanced algorithms like backpropagation for training. Rather, these can be trained by "stepping toward the error step by step" specified by the learning rate (Den Bakker, 2017).

### 12.1.4 Deep Neural Network

A feed-forward neural network is a term used for a multilayer perceptron. The information in this network flows in one direction as suggested by the name. The number of hidden layers are multiple ( more than one layer), thus as expected, power of this network increases and it can now learn complex and non-linear patterns as well.

### 12.1.5 What Activation Function To Use?

Although neural networks represent a large collection of linear combinations and use of linear activation functions, their true power lies in their ability to model them to solve complex nonlinear behaviors. "Sigmoids and ReLU" are two examples of two nonlinear activation functions. But there exists many other popular nonlinear activation functions. To name a few of those are ELU, Leaky ReLU, TanH, Maxout and a newer state-of-the-art 'Mish' (Misra, 2019). There is no general rule for which function to use for the output unit. The linear activation function is used for the regression task because there is a single unit output, but if there are $n$ output nodes then the softmax activation function are used to classify tasks having $n$ classes. Using the softmax function, the network is forced to give the output a probability between 0 and 1. A sigmoid activation function can be used to output probability for a binary classification using a single output node. It is important to choose the correct activation function for hidden units. In the backward pass, the update counts on the activation functions derivative. Deep neural networks face a "vanishing gradient problem", which means that the gradient for updated weights is zero in the first few layers. Another problem faced by deep neural networks is the "explosion gradient problem", which can be steep and exponentially large. This is especially true for activation functions such as sigmoids, whose derivatives take smaller or greater values. These problems can be prevented by using activation functions such as the ReLU which produces a derivative of 1 for positive output and 0 otherwise. Therefore using a ReLU activation function for sparse networks having a small number of activated connections, the loss that is passed through the network seems more useful, but in some other cases, ReLU can make many neurons inactive (Den Bakker, 2017).

### 12.1.6 Loss Function

The main purpose of training neural networks for supervised learning problems is to minimize the loss function, which, during the forward pass, compares prediction vis-a-vis the ground truth. The loss function output is then used to optimize the weights towards more real values, during the backward pass. Thus loss function is a very important component used for network training. Therefore, the accuracy of the loss function is the key to force the network to optimize for the desired prediction. For example, an unbalanced dataset requires a separate loss function.

Mean Squared Error (MSE), Mean Absolute Error (MAE), and Category Cross Entropy can be used as loss functions. There are also some other common loss functions available or alternatively, customized loss functions can be created, which improves the ability to optimize for the desired output (Den Bakker, 2017).

### 12.1.7 Optimization Methods

The well known optimizer 'Stochastic Gradient Descent(SGD)' is also the most popular one among researchers. The SGD technique is widely used in many other machine learning models as well. SGD optimizer uses SGD method to find minima/maxima by iterations. SGD is available in many variants which easily adapt to learning rate and speed up the convergence and do not need much of a tuning. Some other available optimizer methods are ADAM, ADAGRAD, CONJUGATE GRADIENT, LBFGS, LINE GRADIENT DESCENT, HESSIAN FREE, RMSPROP etc (Den Bakker, 2017).

## 12.2 Transfer Learning

Transfer learning is a research area in machine learning where a model that is already trained for a task is reused to train a model on another task. It is a very popular technique in deep learning, where such architectures or models are utilized as a starting point - typically for computer vision and natural language processing tasks. Basically, the learned model weights are used as initial weights - or one can say the learnings are transferred on a dataset that is of interest to us and training starts from there. This is particularly useful, when we have less dataset or dont have much time to do training. ImageNet is one such image database which is used to train the pre-trained models. (Deng et al., 2009) which contains millions of images with different class labels. We have used such pre-trained models like VGG-16, ResNet-18 and ResNet-50 in our proposed deep neural network architecture.

### 12.2.1 Convolutional Neural Network as feature extractor

It is a transfer learning technique where weights in all the layers of the Convolutional Neural Network are frozen and these layers are not trainable. These layers are generally known as feature extraction layers. The fully connected layer or usually the head of the network is replaced with the newer one, initialized with random weights. Only this fully connected layer is then trained on custom dataset (Takhirov, 2017).

### 12.2.2 Fine-tuning

This is also a transfer learning technique. In this case the head or some part of the head is changed to address the different number of outputs that are required to train custom dataset. The training does not start at random, as the model is pre-trained and the weights are already initialized. The training is carried out normally and is possible to either fine-tune the weights across all the layers or some of them. This is usually carried out when very small changes are required and to generalize the model to fit to the custom dataset (Takhirov, 2017).

### 12.2.3 Combining fine-tuning and CNN as feature extractor

When the above two methods are combined i.e, first freeze the feature extractor and then training is done on fully connected layer. After the network learns the weights and is able to generalize on new dataset, the feature extractor or some part of it is unfrozen and set to trainable. The model training will then continue as usual. In our research this hybrid approach is being used to fine-tune the selected model to predict acceptability index of a guava (Takhirov, 2017).

## 12.3 Multi-Task Learning

Multi-task learning is a field of machine learning where multiple tasks are learned from one shared model. One or various inputs can be used for many different outputs. However, these outputs are interconnected. Example would be predicting stop sign, pedestrians, trees on the road at the same time (Peng et al., 2020).

## 12.4 Convolutional Neural Networks

### 12.4.1 Introduction

A convolutional neural network is a particular type of feed-forward neural network also known as ‘ConvNet’. In this type of network one or multiple convolutional layers can be complemented with fully connected layers. The network architecture is known as a Fully Convolutional Network, if the network consists of only convolutional layers.

### 12.4.2 Filters and Parameter Sharing

The significant part of convolutional network is convolutional layers. It has blocks (such as sliding windows) that convolve the input data so that it can be detected within the block throughout the input data function. The parameters of each block are shared in this technique, and the kernel size or filter size is the size of the block.

This method can be applied multiple times because of several different features in the input data, and number of features for learning. The number of filters are determined by the number of features also known as “filter depth”. In other words for a convolutional layer with five filters on the image data, the convolutional block attempts to learn five different features in the images. For example, various facial features such as nose, eyes, mouth, ears, and eyebrows are never explicitly specified, but the network itself tries to learn the worthwhile features it needs.

### 12.4.3 Optimizing With Pooling Layers

Another common optimization technique for CNNs is the “pooling layer”, which is a smart way to reduce the number of trainable parameters. Two widely used pooling layer techniques are:

1. Average pooling averaged and extracted the inputs for a specified block size
2. Maximum (max) pooling extracts the maximum value within a block.

Therefore, both of these pooling layers provide translational invariance. In other words, the pooling layer method is not much effected by the feature location. Hence, this would reduce the number of parameters to be trained which in turn reduces the network complexity and also prevent overfitting. Another advantage is that training and inference time are significantly reduced (Den Bakker, 2017).

### 12.4.4 Batch Normalization

Batch normalization is a common CNN optimization technique, which normalizes the input of the current batch and then feeds it to the next layer. Thus for each batch average activation is near zero and standard deviation is about 1. This technique avoids internal covariate shifts, thus making this model more generalized and faster to train because the input distribution of data per batch has less impact on the network (Den Bakker, 2017).

### 12.4.5 Padding and Strides

Previously, the default strides were used for networks, that means the model contains one input on each axis- that is step size of one. However, if the information available at the pixel level of dataset is less granular, a larger stride can be tried. As the stride size increases, the convolutional layer skips more input variables on each axis, reducing the number of trainable parameters. Therefore, this method speeds up convergence without significant reduction in performance.

Another parameter that can be adjusted is padding, which sees how the boundaries of the input data are processed. If no padding has been added, only border pixels are included as in the case of images. Therefore, if the boundaries are expected to contain valuable information, try to add padding to data. Therefore, this adds a boundary for dummy data that can be used during convolution. The main idea to use padding is that the data present in each convolutional layer will have same dimension and hence stacking up more convolutional layer on top of each other is simple. The size and data complexity makes stride and padding highly dependable, when combined with the pre-processing techniques that could be used. There are no general rules for choosing padding and stride values (Den Bakker, 2017).

### 12.4.6 Different Types of Initializations

Weight and bias initialization is very important for CNNs. Gradient can be reduced by some initialization techniques due to the gradient magnitude in the final layer for very deep neural networks. Therefore, choosing the right initialization can speed up network convergence (Den Bakker, 2017).

### 12.4.7 Fully Connected Layers

A Fully Connected Layer (FCL) connects all the nodes in one layer to the output of the next layer. The activation function in the output of FCL is named as Softmax function. FCL produce the output which is a class of probabilities, a probability is assigned to each class and the summation of all these probabilities is 1 (Den Bakker, 2017).

## 12.5 Object Detection and Localization

### 12.5.1 Region Proposals

The image of interest always has two primary regions, which are the foreground region consisting of the image itself and a vast background region. We can always make the assumption that there is a little change in the intensity of a pixel which belongs to the background region as compared to the intensity of pixels which belong to the foreground, that is to the image itself. Moreover the pixels belonging to the image will have different intensity values depending upon the part of the image they belong to, thus we can conclude that the region image itself can have multiple sub regions, and can lead to identify Regions of Interest (ROI). Therefore Region proposal is a useful technique for identifying islands in regions where pixels are similar to each other, that helps in object localization and detection by building a bounding box that fits exactly around the object in the image. The algorithm 'SelectiveSearch' is widely used for object localization. It groups similar pixels of regions based on their pixel intensities, which in turn, take advantage of the color, texture, size, and shape compatibility of the content in the image. First, SelectiveSearch over segments the image by grouping pixels based on the attributes mentioned above, and then iterates over these groups to further group those on the basis of similarity between them. At each iteration, small areas are combined to form a large region.

The above diagram clearly shows that the pixels that belong to the same group have similar pixel values, and the preceding diagrams grids represent the candidate regions (region proposals) coming from the SelectiveSearch method (Kumar et al., 2019).

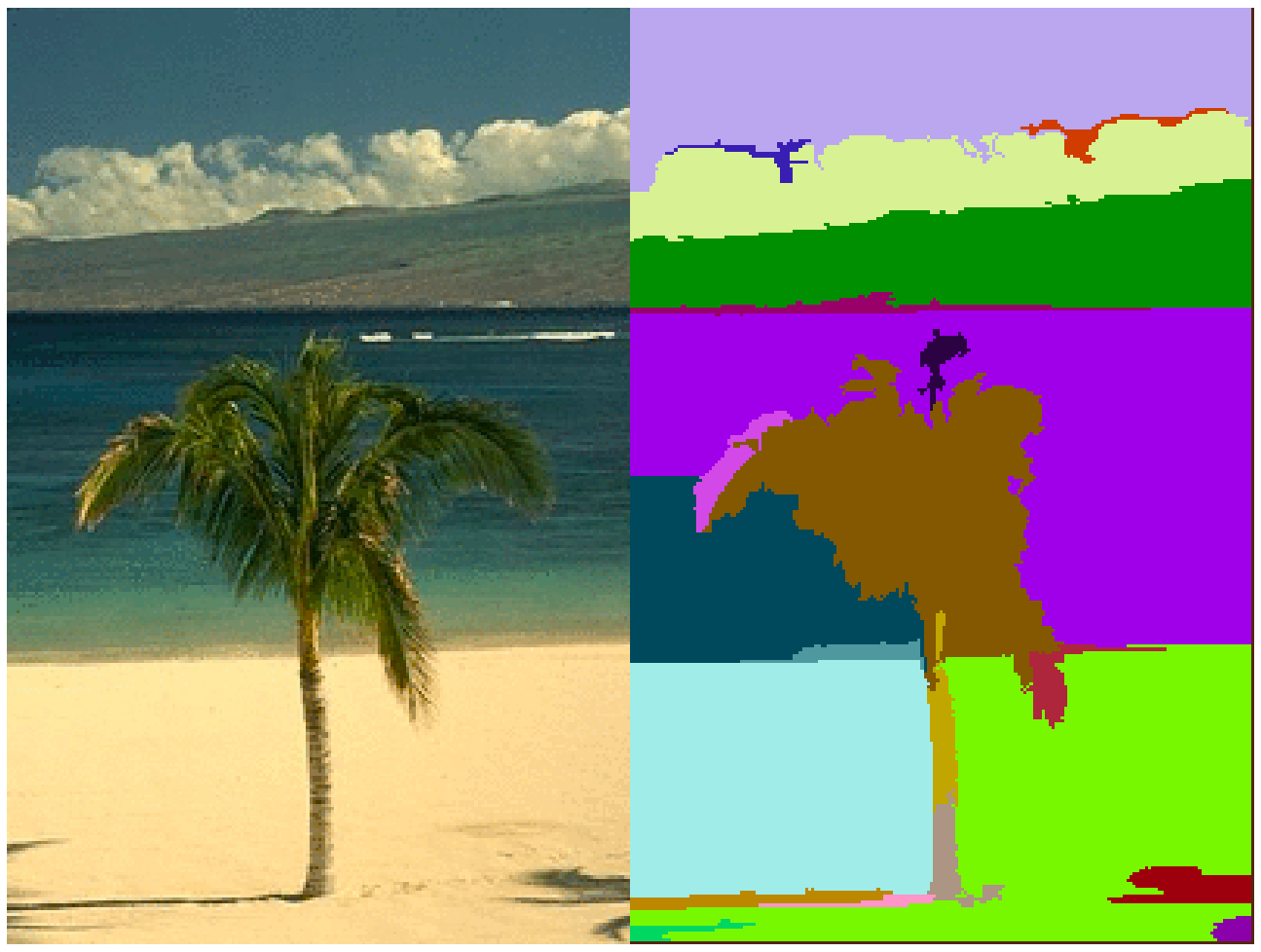

Figure 8. Selective Search, from Rosebrock, 2020

Now the next step is object detection and localization by making use of these identified region proposals. The region proposals having a high intersection with the position of the object (ground truth) in the image are labeled as containing the object, and those having a low intersection are labeled as the background (Ren et al., 2015).

### 12.5.2 Intersection over Union(IoU)

Imagining a prediction of a bounding box for an object. IoU is the ratio of the overlapping area between the bounding boxes of 'intersection' (that is overlap of predicting boundary box and the actual boundary box) and the 'union' (which measures the overall space possible for overlap) to the combined area of both bounding boxes. The diagram below represents the ratio IoU. The left rectangle represents the left bounding box, that is the 'ground truth', and the right rectangle represents the right bounding box, that is the 'predicted location' of the object.

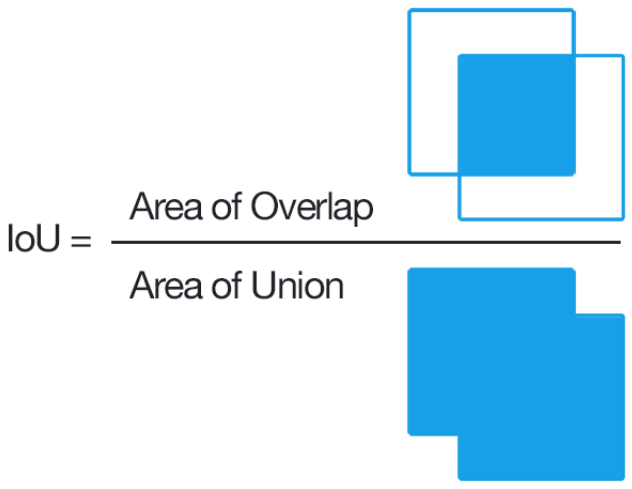

Figure 9. Intersection over Union, from Rosebrock, 2016

Another diagram below, shows variation in the IoU metric as the overlap between bounding boxes varies.

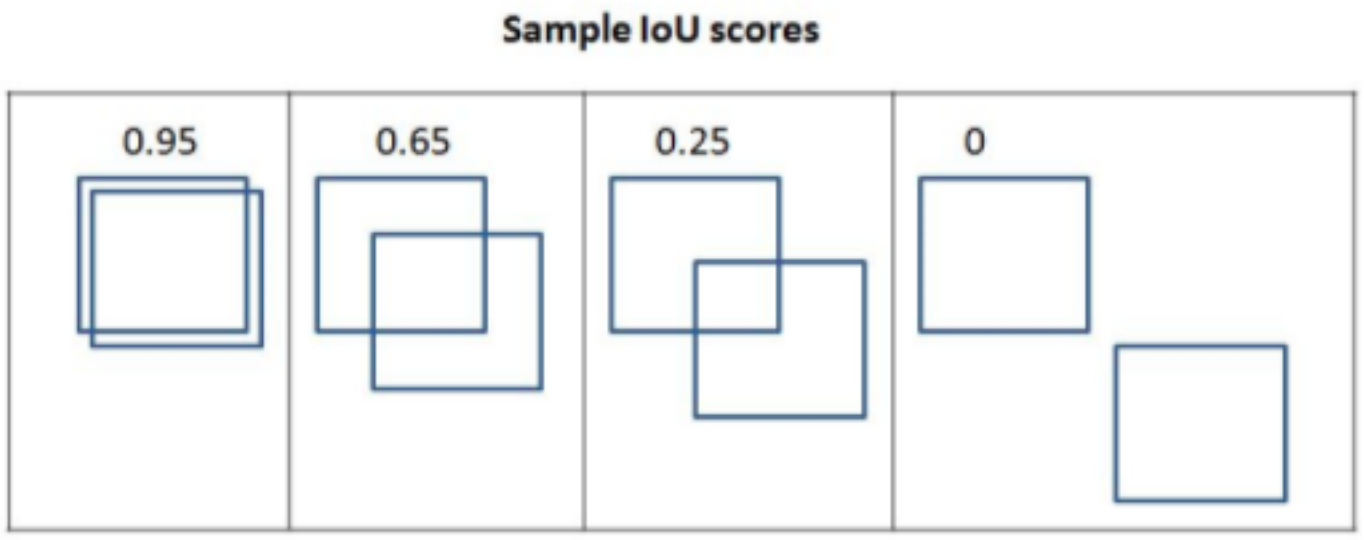

Figure 10. Intersection over Union Metrics

As shown in the above diagram, when the overlap decreases, IoU decreases and where there is no overlap, the IoU metric is 0.

Mean average precision mAP is defined as the mean precision value. The precision values are the different IoU threshold values calculated across all the classes of objects present within the dataset. It helps to quantify the accuracy of the predictions coming from our model. Where the precision is represented by the below ratio formula.

$$\text{Precision} = \frac{\text{True positives}}{(\text{ True positives } + \text{ False positives })} \tag{13}$$

The average of precision values calculated at various IoU thresholds is defined as the 'Average precision'. Where a true positive is a bounding box with an IoU that predicts the correct class of objects and has a ground truth greater than a certain threshold, and false positives are bounding boxes that mispredict a class or have an overlap that is less than the defined threshold with ground truth. In addition, if there are multiple bounding boxes identified for the same ground truth bounding box, only one box will be true positive and all others will be false positives (Rezatofighi et al., 2019).

### 12.5.3 Anchor boxes

Anchor boxes provide a convenient alternative to the SelectiveSearch algorithm. Typically for an object, a majority will have a similar shape and a good idea is available for the height and width of the bounding box for an image even before training the model, though some images still can be scaled to create heights and widths either much smaller or much larger than the average. As the ground truth values of the aspect ratio, height/width ratio define the anchor boxes that represent most of the objects in the dataset. Typically, the size of the anchor box is obtained by employing K-means clustering on top of the ground truth bounding boxes of objects present in images.
The two steps involved in this process are:

1. Each anchor box is slided over the selected image from top left to bottom right.

2. The anchor box with high IoU with the object will be labeled mentioning it contains an object, whereas all others will be labeled 0: The labeling is achieved by defining the threshold of the IoU in such a way that if IoU value is greater than threshold, the object class is 1, and if less than threshold, the object class is 0, and otherwise unknown.

Once the defined ground truths obtained, the model to predict the location and the offset of an object to match it to the anchor box can be built.

To understand anchor boxes represented in the image, see the example image below.

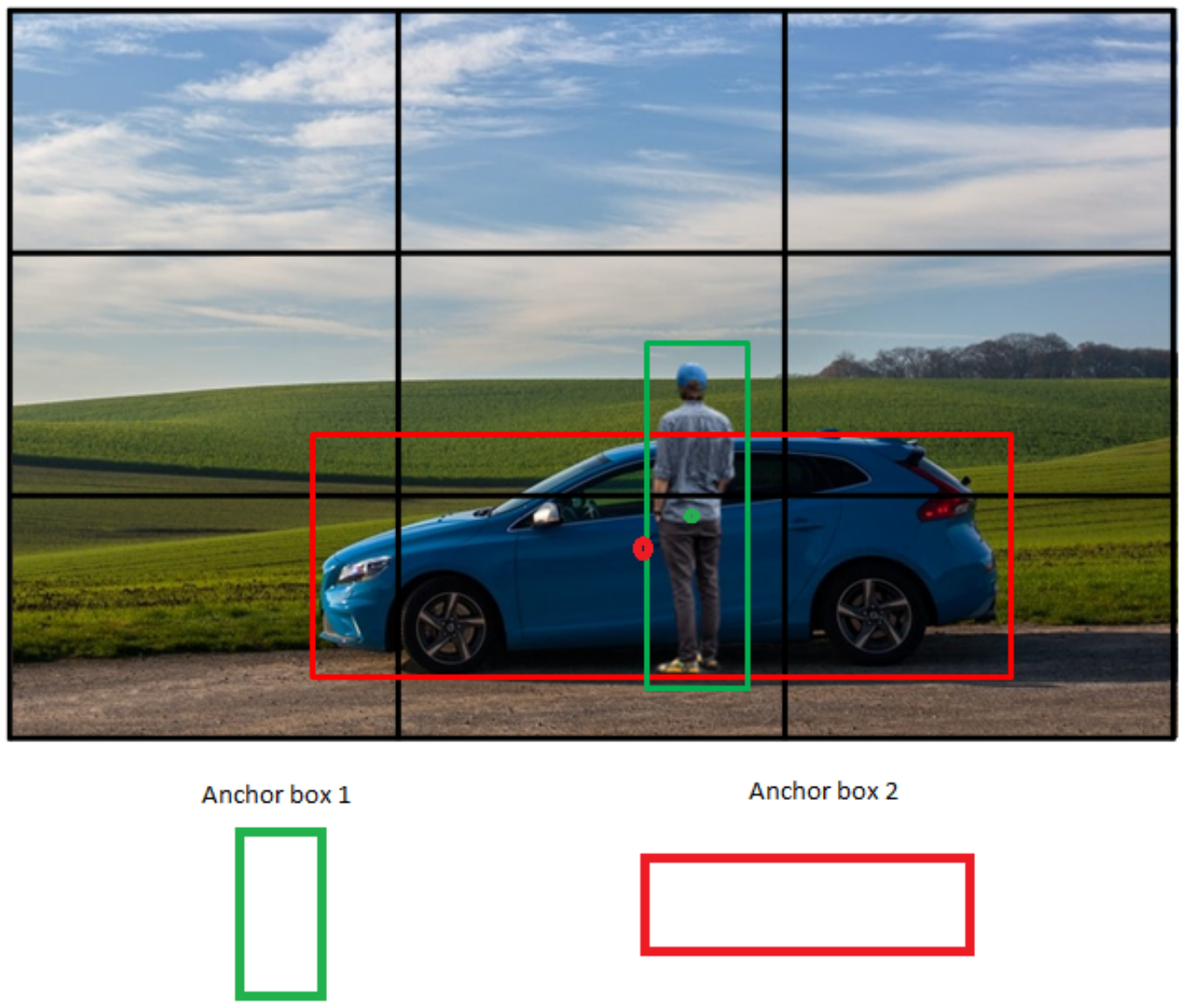

Figure 11. Anchor Boxes

The below example image, which corresponds to two objects (classes) a person and a car, shows two anchor boxes, one with height greater than width and the other with width greater than height (Ren et al., 2015). The two anchor boxes are slided over the image and locations noted for the highest value of IoU vis-a-vis the anchor box and the ground truth, which denotes that the particular location contains an object while no object exists in the rest of the locations.

The anchor boxes can also be created with varying scales to accommodate different scales to present an object within the image. Here is an example:

The point to note here is that all anchor boxes have the same center but different aspect ratios.

### 12.5.4 Region Proposal Network

Assuming there is an image of dimensions 224 x 224 x 3, and the available anchor box 8 x 8. Then for a stride of 8 pixels, 224/8 = 28 crops of a picture are fetched for every row that is 28 x 28 = 576 crops fetched in total from the picture. Now when each of these crops is passed through a Region Proposal Network (RPN) model to identify whether the crop contains an image or not, that is RPN suggests the likelihood of a crop containing an object (Ren et al., 2015).
The steps performed by an RPN trained model to identify region proposals with a high likelihood of containing an object are summarised hereunder:

1. To fetch the image crops, Anchor boxes having different sizes and aspect ratio are slided across an image.

2. The IoU calculated vis-a-vis the ground truth and bounding boxes in the image, and the crops obtained in step 1.

3. The model trained with threshold such that crops having IoU greater than the threshold, marked as containing object and crops having IoU less than the threshold marked as devoid of object.

4. The model is trained to identify regions containing an object.

5. Non-max suppression performed to identify the region candidate with highest probability of containing an object and other region candidates that have a high overlap are eliminated.

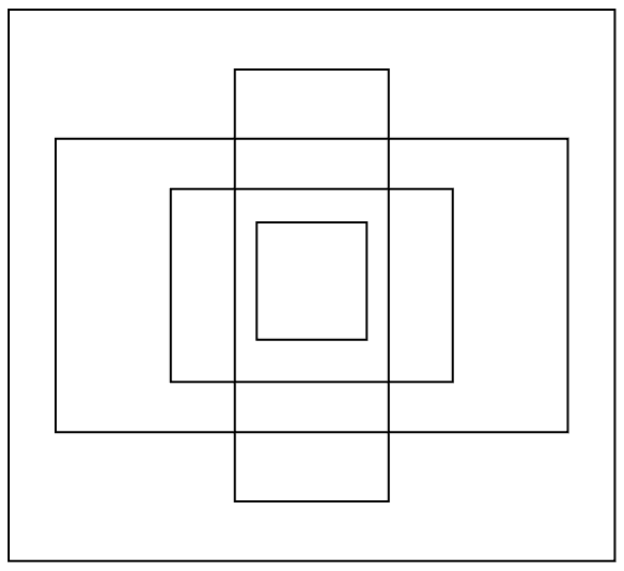

Figure 12. Aspect Ratio

Table 6. Comparison of RPN and SelectiveSearch

| **SelectiveSearch** | **Region Proposal Network (RPN)** |
|---|---|
| Provides region candidates based on computations on top of pixel values. | In an RPN the region candidates generated based on the strides with which the given anchor boxes are slid over the image. |
| The region proposal generation is done outside of the neural network. | It can be part of the object detection network. No unnecessary computations needed to detect region proposals outside of the network. A single model can identify regions, classes of objects in the image, and their corresponding bounding box locations. |

### 12.5.5 R-CNN

Proposed by (Girshick et al., 2014) R-CNN is the abbreviation for "Region-based Convolutional Neural Networks". This network functioning is based on the idea that has two steps. The first step is to identify region proposals. The selective search algorithm finds around 2000 region proposals for a single image. The second step is for classification. In this step these region proposals are passed into the CNN, the CNN then extracts the features to be used for classification from each region proposal independently.

### 12.5.6 Fast R-CNN

Proposed by (Girshick, 2015) in 2015. Fast R-CNN came into being due to some problems in R-CNN such as: a) R-CNN takes a lot of time to train the network as it needs to classify 2000 region proposals per image. b) As it takes almost 47 second for each test image, therefore, it is not feasible to implement in real time. c) Since no learning is possible in case of selective search as it is a fixed algorithm. This results in generation of bad candidates for Region of Interest (ROI). Thus, fast-RCNNs has come out to be the one building faster object detection algorithm and solving the short- comings and issues of R-CNNs, the approach is similar to the R-CNN algorithms. In fast R-CNN the input image is fed to the CNN instead of the region proposal candidates to generate a convolutional feature map, and then from this map, region proposals are detected and warped as squares, and reshaped into a fixed size using the ROI pooling layer to be fed into connected layers. Now to predict the offset value for the bounding box and class of the proposed region, softmax layer is used from the ROI feature vector.

### 12.5.7 Faster R-CNN

Introduced in 2016 by (Ren et al., 2016). In Faster R-CNN, selective search was replaced by Region Proposal Network which tells us about those regions, that are likely to have the object. Regions Proposal Network does this by using anchor boxes of different scales and aspect ratios. The anchors then are fed into ROI pooling layers and from there it is divided into two outputs. One is classification output, which classifies the object and then second one is regression output, which is used to regress the bounding box.

### 12.5.8 YOLOv4

The YOLOv4 algorithm was developed from YOLOv1, YOLOv2, YOLOv3, and YOLO networks can generally be divided into three parts: backbone networks, neck networks, and forecasts. Backbone networks are primarily responsible for extracting image features. Though further evolution of deep learning enables more layered networks, and extraction of more feature information, it will add to the cost of training. In addition, after passing a certain number of layers, the training effect is reduced instead. Therefore, lightweight layered networks are becoming popular with academia, students and the research community. Neck networks enhance image features, process and enhance shallow features extracted from the backbone network, and merge shallow features with deep features to enhance network robustness, thereby able to fetch out more effective features. The head network further classifies and regresses the functionality acquired by the backbone and the neck networks (Orac, 2020).

#### 12.5.8.1 Backbone

ResNet-50, EfficientNet, ResNet-101, Darknet53, and lightweight networks such as Mobilenet V1, V2, ShuffleNet1, 2 are some of the popular backbone networks for YOLO. The CSPDarknet53, an improved version of Darknet53 of YOLOv3, is a popular backbone network of YOLOv4. CSPDarknet53 uses DenseNet and Cross-Stage-Partial-connection (CSP) for enhancement of the learning capabilities of convolutional networks. Accuracy is not compromised, but significant reductions in network model calculations and memory costs have been achieved. To split the input feature map channel layer in two, Darknet and CSPDarknet53 use a 1x1 convolution before each residual network and add a CSP after each residual module. The Mish function is the main activation function of CSPDarknet53, avoids saturation due to capping, and has no hard zero boundaries as present in the ReLU function. Its smooth performance allows passage of better information. After traversing CSPDarknet53, the input image has three outputs (C.-Y. Wang et al., 2020).

#### 12.5.8.2 Neck

The main use of the neck is to generate the feature pyramid, which enhances the model's detection of objects of different scales, thus allowing it to recognize the same objects of different sizes and scales. FPN was always at the forefront of the functional aggregation layer of object detection frameworks until the advent of PANet (Path Aggregation Network). SSP (Spatial Pyramid Pooling) and PANet are used by the YOLOv4 neck network. SSP, which allows the spatial size of each candidate map to be preserved, convolves candidate images using four different sized sliding kernels, 1x1, 5x5, 9x9, and 13x13 to get feature maps of the same dimensions, and applies multiscale Max-Pooling to connect these feature maps to form a fixed size feature map as output (He et al., 2015b). FPN also had a limitation of extracting information only from high-level feature layers. Therefore, PANet is developed to overcome this limitation of FPN. Several changes have been incorporated in FPN and Mask R-CNN to develop PANet, which provides a more flexible ROIPooling (Region of interested Pooling) that extracts and integrates features of different dimensions (S. Liu et al., 2018).

#### 12.5.8.3 Prediction

The final detection part uses predictions generated by applying the anchor box to the feature map to generate the final output vector containing the class probabilities, object scores, and bounding boxes. Feature maps of different dimensions are provided as inputs to PANet, and PANet splits those into 3 sets with different dimensions as output of the convolution operation. The obtained 3 head sizes are (76 x 76 * 3 * (4 + 1 + class)), (38 x 38 x 3 * (4 + 1 + class)), (19 x 19 x 3 * (4 + 1 + class_num)) ), where 4 depicts the coordinate value of the calibration box, 1 depicts the confidence level of the calibration box, and classes depicts the number of categories. Each head has three bounding boxes that enclose a particular dimension object taken out from three sets of objects with different dimensions. The observations confirmed that YOLOv4 predictions are compatible with YOLOv3.

#### 12.5.8.4 Loss function

The YOLO series loss function calculations are based on CIoU-Loss – bounding box regression loss and DIoU-NMS- classification loss. Bounding box calculations range from smooth L1 losses, IoU losses, GIoU losses to today's CIoU losses. Taking into account the overlap area between the prediction box and the target box, the distance between the center points, and the aspect ratio, the. CIoU loss improves the speed and accuracy of the regression of prediction box. DIoU NMS is used to exclude prediction boxes with higher reliability, which allows retaining only highest value prediction boxes. The formula used to calculate YOLO loss function is given here under, along with some other formulae:

$$\begin{aligned}
\text{Loss} = {} & \lambda_{\text{cord}} \sum_{i=0}^{s^2} \sum_{j=0}^{B} 1_{ij}^{\text{obj}} \left[ \left(x_{ij} - \bar{x}_{ij}\right)^2 + \left(y_{ij} - \bar{y}_{ij}\right)^2 \right] \\
& + \lambda_{\text{cord}} \sum_{i=0}^{s^2} \sum_{j=0}^{B} 1_{ij}^{\text{obj}} \left[ \left(\sqrt{w_{ij}} - \sqrt{\bar{w}_{ij}}\right)^2 + \left(\sqrt{h_{ij}} - \sqrt{\bar{h}_{ij}}\right)^2 \right] \\
& + \lambda_{\text{cord}} \sum_{i=0}^{s^2} \sum_{j=0}^{B} 1_{ij}^{\text{obj}} \left[ \left(c_{ij} - \bar{c}_{ij}\right)^2 \right] + \lambda_{\text{cord}} \sum_{l=0}^{s^2} \sum_{j=0}^{B} 1_{ij}^{\text{obj}} \left[ \left(p_{ij} - \bar{p}_{ij}(c)\right)^2 \right]
\end{aligned} \tag{14}$$

$$
\begin{aligned}
IoU &= \frac{Prediction \cup GroundTruth}{Prediction \cap GroundTruth} \\
CIoU &= IoU - \left(\frac{\rho^2(b, b^s t)}{c^2}\right) - \alpha v \\
v &= \frac{4}{\pi^2}\left(\arctan\frac{w^{gt}}{h^{8t}} - \arctan\frac{w}{h}\right)^2 \\
\alpha &= \frac{v}{(1 - IoU) + v}
\end{aligned}
\tag{15}
$$

The explanation of the terms used in the above formulae: $A_c$: The minimum closure area of the prediction box and the ground truth box, U: The sum of the area of the prediction box and the real ground truth. Prediction means prediction box, Ground Truth means ground truth box, $\partial$ : The newly added weight coefficient based on GIoU, v: The similarity of the aspect ratio between the predicted frame and the real frame, c: is the length of the diagonal of the box, $\rho^2$ means the square of the distance between the center of the real frame and the predicted frame (De Carolis et al., 2020).

## 12.6 Image Classifcation and Regression

### 12.6.1 AlexNet

As mentioned in the description of CNN, AlexNet, introduced in 2012 by (Alex Krizhevsky, 2012) is a very important CNN architecture consisting of eight layers. The five of which are convolutional layers followed by other three, which are fully-connected layers. Due to training via GPU AlexNet has developed a depth capability that has great impact on its computational efficiency, the reason its excellent performance over several image recognition competitions, won it awards. This network comprises of 60 million parameters. It achieved top 5% test error of 16.4% in ImageNet Large Scale Visual Recognition Challenge (ILSVRV).

### 12.6.2 VGG

In the year 2014, the Visual Geometry Group (VGG) (Oxford University) invented it, thus the name is VGGNet or simply VGG. It is a classic CNN architecture and the main idea is to increase the depth of such networks. The network is characterized by its simplicity, an innovative object recognition model that supports up to 19 layers. ((Karen Simonyan, 2014) used for ILSVRV, and has the distinction of achieving high accuracy in many competitions. It comprises of 138 million parameters and achieved a top 5% test error of 7.3% in ILSVRV. Noticeably, VGG has very large receptive fields.

#### 12.6.3 GoogleNet

It is another important CNN architecture which is 22 layers deep. Introduced in 2014 by (Szegedy et al., 2014). The idea of GoogleNet is to use an inception module which consists of kernels of different granularity that is: 1x1 to reduce the dimension and then applying 3x3 and 5x5 kernels. Finally these are concatenated to form an inception module. These modules are introduced in between the convolutional layers. It achieved a top 5% test error of 6.7%.

#### 12.6.4 ResNet

ResNet as discussed by (He et al., 2015a) is a short for Residual Network. This network was introduced in 2015 at the ILSVRC, after observing that with the increase in depth of the network the accuracy gets saturated and the gets degraded very fast. Core idea is to be able to use shallow network to create a deeper network. The network goes upto 152 layers deep. The architecture introduces a shortcut connection also known as skip connection, which without any modification fits unmodified input obtained from the previous layer into the next layer and allows the network to be trained in case the flow of gradient stops at any layer. This achieved a top 5% test error of 3.5%.

# 13 Methodology

## 13.1 Data Collection and Preparation

### 13.1.1 Object detection localization

To train YOLOv4 for the purpose of object detection and localization, we curated the dataset consisting of 576 images, consisting of 452 images for training and 124 for validation approximately (80:20 split). Due to the novelty of this project, images of guava at different acceptability index are difficult to find. Therefore, most of the images contained are our own images. The images consists of guavas with noisy background collected through the internet, by clicking pictures of the guava using a high resolution mobile phone camera at different time intervals and with various backgrounds.

Further images were created from the validation set used by CNN to predict the acceptability index to provide a proper distribution of data. These images were edited, the background removed, and then mixed with images of other objects using photo / image-editor. This dataset trained YOLOv4 on guavas with different acceptability indexes and provided a better trained model for the purpose.

These images were then annotated using image annotation tool labelImg (Tzutalin, 2015) to train YOLOv4. This tool was used to create bounding boxes around guavas within an image. A bounding box is rectangular/square in shape, and surrounds the object involved, which is identified with the help of the algorithm used. The position coordinates of the bounding box are then saved in a text file, and are used by the algorithm for further training of the model.

### 13.1.2 Predicting acceptability index

Dataset to predict acceptability index consists of 299 photos of guavas clicked with high resolution cellphone camera at different time intervals for the purpose of training the VGG-16 + Custom FCN, ResNet-18 + Custom FCN, ResNet-50 + Custom FCN. The dataset consists of 239 images for training and 60 for validation (80:20) split.

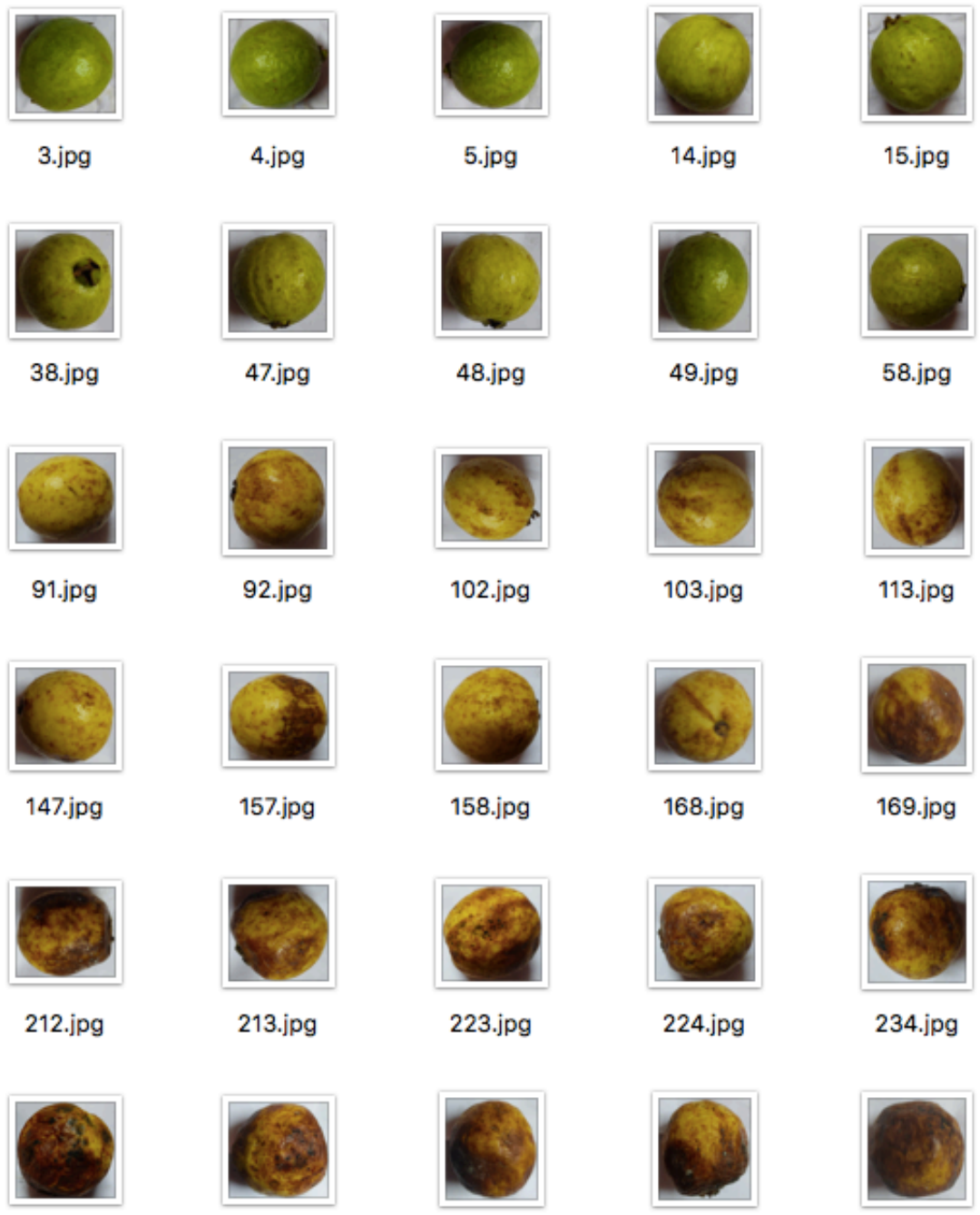

Figure 13. Training Dataset

Since there is no available dataset for the purpose of carrying out any kind of sensory evaluation on fruits - particularly guava, the dataset for guava has been curated over the course of three months. Each day, 1 to 4 photos of guavas were taken at different time intervals until the guavas became rotten. These photos are then assigned points using a nine-point Hedonic scale (Lim, 2011) for each of these sensory attributes: shape, color and texture, where the 9 is for extreme like and 1 is for extreme dislike. Furthermore, to make sure the numbers are correct, got a few samples reviewed from my peers for sensory evaluation and from food technologist to have their expert opinion on this. The reason for using hedonic scale is that it has been the most common scale and is widely accepted when it comes to testing how well the consumers will accept a particular food item or to test the preference of a consumer. Secondly, being a discretized nine-point scale it is widely accepted by researchers and participants of a research survey as is easy to use and implement.

Furthermore, hedonic scale provides us with degree of acceptance and with such score carrying out sensory evaluation is a easy.

The final mean hedonic score is applied based on the fact that most of the times consumers reject fruits based on their color and other times based on their texture. Shape is of importance as well, however most of the times its texture and color consumers are on a lookout for, whenever selecting a fruit. This was further evident after an anonymous survey was put up about - What properties/attributes are people generally looking for whenever hand-picking fruits/vegetables of the shelf? And for 71.2% its Texture, 31.5% its Color and 72.6% its Shape, which supports the argument of how color and texture is important. Based on these two attributes, a weighted scheme is designed and a weighted mean of the hedonic values for each of the photos of the guava is calculated, where the ratio of weights is 2:1:2 for color:shape:texture respectively and is calculated as:

$$W = \frac{\sum_{i=1}^{n} w_i X_i}{\sum_{i=1}^{n} w_i} \tag{16}$$

$W$ = weighted average $n$ = number of terms to be averaged $w_i$ = weights applied to $x$ values $X_i$ = data values to be averaged. $w = [2, 1, 2]$, $X$ = [Color, Shape, Texture]

## 13.2 Object detection and localization

For detection and localization of guavas pre-trained YOLOv4 on Microsoft COCO dataset (Lin et al., 2015) is used. COCO dataset includes common objects such as commonly found fruits, bikes, cars, people, cats, dogs etc. As YOLOv4 has fruit features, the higher level features would be transfer learned and training on our custom guava dataset would therefore converge faster as the training would not start with random initial weights. Further, using transfer learning technique would be highly beneficial for cases such as ours with less data. The head of the YOLOv4 predicts the bounding box of the object, and outputs the center coordinates, width, and height $b_x$, $b_y$, $b_w$, $b_h$ respectively. The bounding box is cropped from the output image - which contains the guava localized and detected by YOLOv4 is fed into the CNN for the purpose of acceptability index prediction.

## 13.3 Acceptability index prediction

Acceptability index being a numeric value between 1-9 our problem of predicting this number turns out to be a regression task. The acceptability index is predicted alongside a label telling the likeability of a consumer as can be seen in the table below. For each of the acceptability index range there is a level attached to it. Based on the predicted regressed value of acceptability index, a level is assigned.

For the purpose of transfer learning using CNN as feature extractor and fine tuning, custom fully connected network is added onto its head to customize the outputs according to what is required. Each of the pre-trained architectures such as VGG-16, ResNet-18 and ResNet-50 are used as base and over that our custom Fully Connected Network (FCN) is added to output regression predictions.

Table 7. Acceptability index levels

| Condition | Level |
|---|---|
| Acceptability Index > 6.5 | Like extremely |
| Acceptability Index >= 6 and < 6.5 | Like moderately |
| Acceptability Index >= 5.5 and < 6 | Like slightly |
| Acceptability Index >= 5 and < 5.5 | Neither like nor dislike |
| Acceptability Index >= 4.5 and < 5 | Dislike slightly |
| Acceptability Index >= 4 and < 4.5 | Dislike moderately |
| Acceptability Index >= 3.5 and < 4 | Dislike very much |
| Acceptability Index < 3.5 | Dislike extremely |

### 13.3.1 VGG-16 + Custom Fully Connected Network (FCN)

Pre-trained VGG-16 on ImageNet has been used as a base layer. Average pooling layer and fully connected layer of the network is customized. Furthermore, dropout layers have been used to prevent over-fitting. The configuration of the custom FCN can be viewed inside the green box as shown in the figure below.

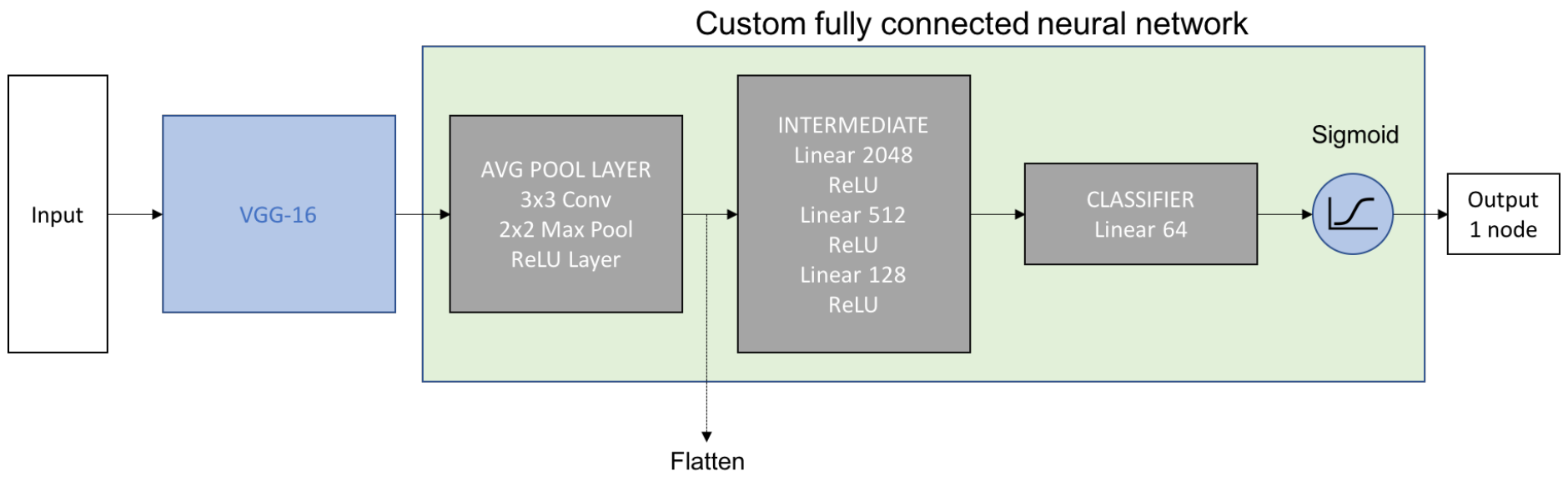

Figure 14. VGG-16 Base + Custom Average Pooling and Custom FCN

### 13.3.2 ResNet-18 + Custom Fully Connected Network (FCN)

Pre-trained ResNet-18 on ImageNet has been used as a base layer. Fully connected layer of the network is customized. Furthermore, dropout layers have been used to prevent over-fitting. The configuration of the custom FCN can be viewed inside the green box as shown in the figure below.

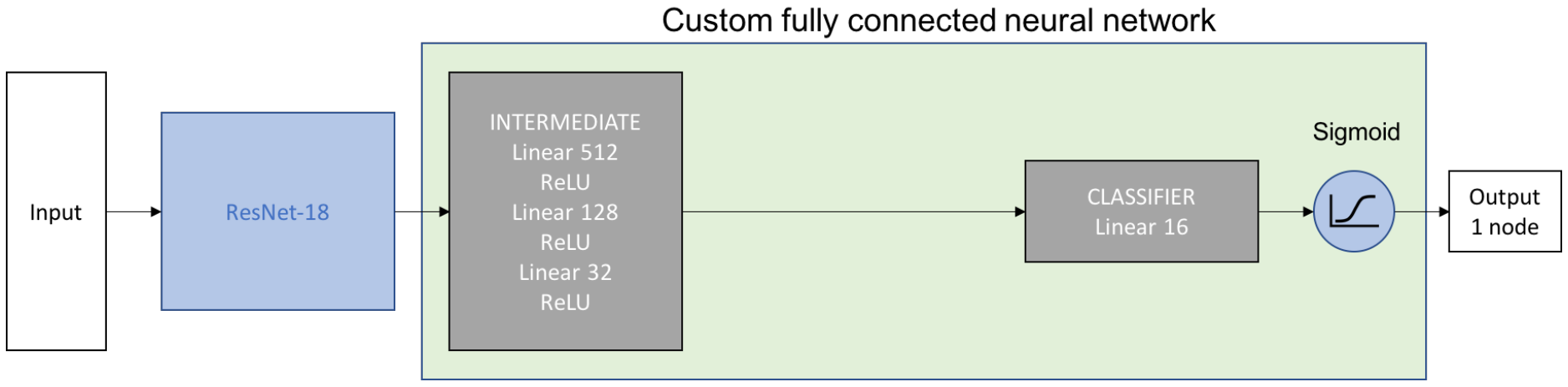

Figure 15. ResNet-18 Base + Custom FCN

### 13.3.3 ResNet-50 + Custom Fully Connected Network (FCN)

Pre-trained ResNet-50 on ImageNet has been used as a base layer. Fully connected layer of the network is customized. Furthermore, dropout layers have been used to prevent overfitting. The configuration of the custom FCN can be viewed inside the green box as shown in the figure below.

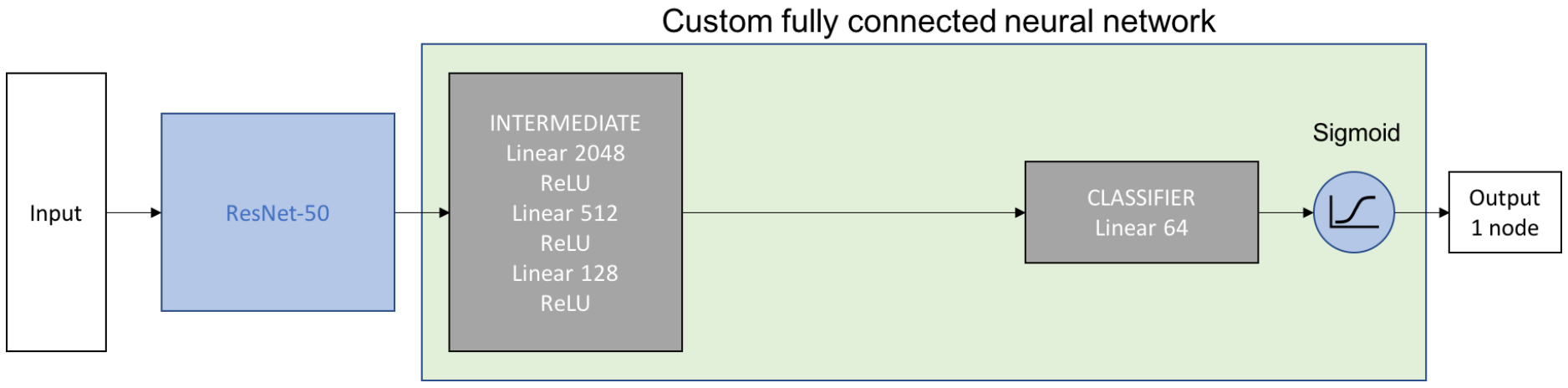

Figure 16. ResNet-50 Base + Custom FCN

### 13.3.4 Model selection for acceptability index prediction

To make predictions we make use of transfer learning utilizing pre-trained deep neural network architectures such as VGG-16, ResNet-18 and ResNet-50. The feature layers of these architectures are frozen, and parameters are learned in the last trainable custom fully connected neural network layer attached to the head. This transfer learning approach is known as using CNN as feature extractor. After finding out the model which has the best MAE, the layers of the feature extractor are unfrozen and then is trained on the whole architecture. This is a combination of using CNN as fixed feature extractor and fine-tuning the CNN. The training or fine-tuning is continued after unfreezing the layers of the feature extractor. The previous step helps in initializing the model and generalizes the model to be able to learn the features of the custom dataset, such as ours without touching the weights in other layers, as initialized after training on ImageNet. Once, this model is able to generalize for new dataset, the training is then started from scratch, and the weights are initialized as per the weights of the CNN model used as feature extractor instead of random initialization. This aids in better generalization of the model to the custom dataset and thus performs better. Moreover, considering the novelty of this project and our dataset, that is not available anywhere, and makes it hard to use the pre-trained models out-of-the-box. However, the pre-trained models converge faster, as they are trained on various images with similar higher level features, and is always better than starting with random initial weights. The figure below should provide a good intuition in regard to fine-tuning and using CNN as a feature extractor.

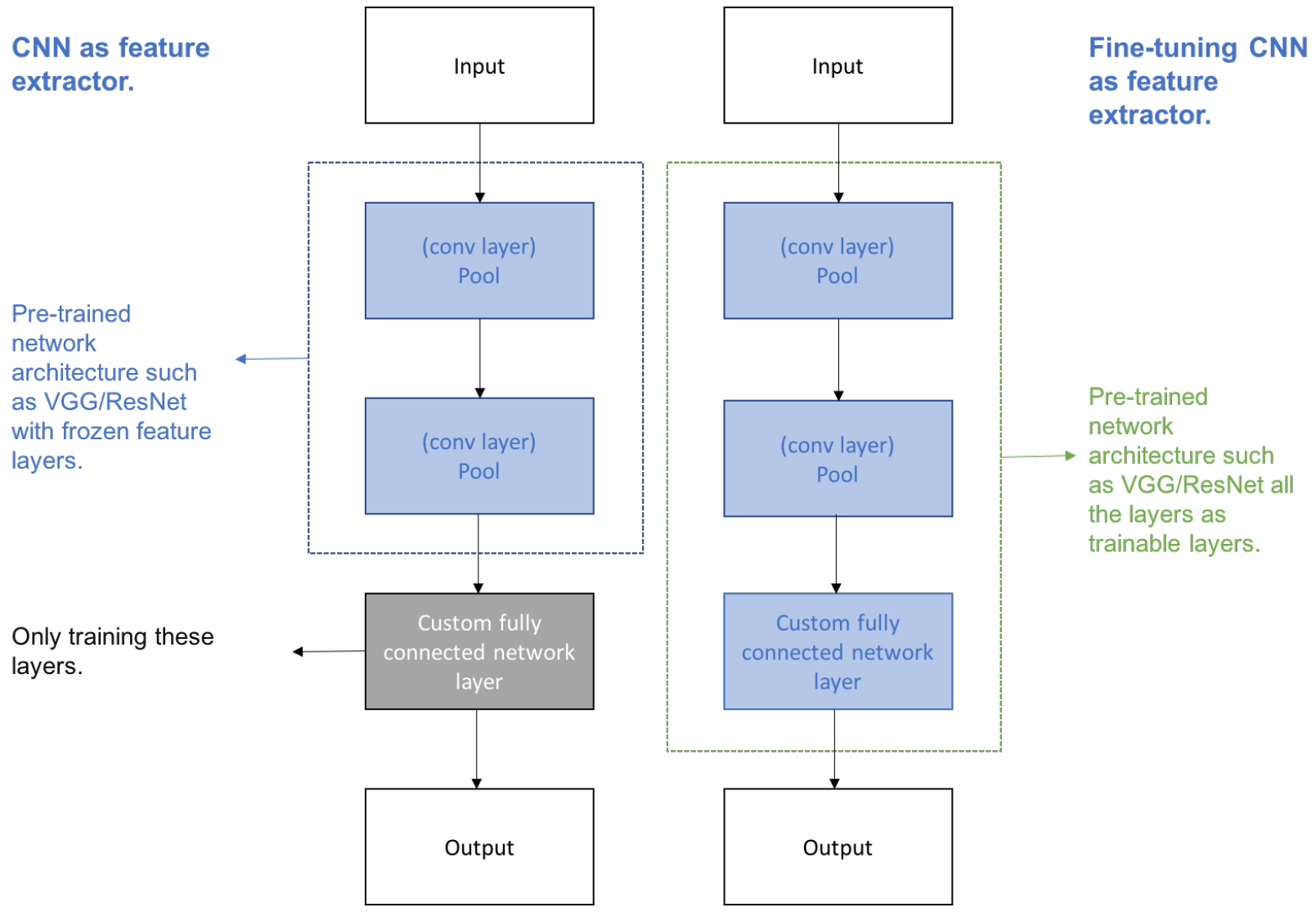

Figure 17. Combining transfer learning techniques of using CNN as feature extractor (left) and Fine-tuning CNN as feature extractor (right). The architecture on the left side of the figure after training is fine-tuned as shown in the figure (right). (Rosebrock, 2020)

### 13.3.5 Hyper-parameters selection for the models

Learning rate for the architecture has been selected manually after starting with a high learning rate and then keeping a track of model convergence, lowering the loss and at the same time keeping the training time fast. For each one of the VGG-16, ResNet-18 and ResNet-50 we chose the learning rate as **0.0001** or **1e-4**. At this learning rate training was fast and was reducing the loss.

Batch size was also selected manually in accordance to the size of the data. After, trying different batch sizes for training and validation data like 10, 16, 32. Batch size of **32** gave best results for all the three architectures (VGG-16 + Custom FCN, ResNet-18 + Custom FCN, ResNet-50 + Custom FCN).

Furthermore, to be able to generalize the models we ran training for **25**, **50**, **50** epochs for VGG-16 + Custom FCN, ResNet-18 + Custom FCN, ResNet-50 + Custom FCN respectively.

The decision to halt the training after these many epochs were based on either test loss not changing for more than 10 iterations or we got the desired MAE.

After, fixing on the desired model, in our case VGG-16 + Custom FCN on the basis of MAE and being able to generalize our custom dataset better, the feature layers were unfrozen and the network was trained on **15** epochs with the same configuration of learning rate i.e **0.0001** or **1e-4** and batch size of **32**. The reason for running **15** iterations was that the training halt became stagnant beyond these many iterations and we got best MAE of **0.26** which was an improvement on the model used as feature extractor.

### 13.3.6 Choosing a loss function

Given below are the formulae for Mean Absolute Error (MAE), Mean Squared Error(MSE), and Root Mean-Square Error(RMSE).

$$\text{MAE} = \frac{1}{N}\sum_{i=1}^{N} \mid y_i - \hat{y}_i \mid \tag{17}$$

$$\text{MSE} = \frac{1}{n}\sum_{i=1,\ \text{L}}^{n} \left(y_i - \hat{y}_i\right)^2$$

$$\text{RMSE} = \sqrt{\frac{7}{n}\sum_{i=1}^{n} \left(Y_1 - \hat{Y}_1\right)^2} \tag{18}$$

Taking arbitrary data a table is created below and Mean Residual Error value is calculated for MAE/MSE/RMSE.

The observations made from the above table are given below:

1. Variation of MAE of 2.4 from the minimum error value 1 is 140% and from maximum error value 4 is 40%. The variation from other values like from 3 is 20% and from error value 2 is also 20%.

Table 8. MAE/MSE/RMSE Calculation

| Observations | $y$ | $\hat{y}$ | $(y-\hat{y})$ | $\lvert y-\hat{y} \rvert$ | $(y-\hat{y})^2$ | |
|---|---|---|---|---|---|---|
| 1 | 10 | 13 | -3 | 3 | 9 | |
| 2 | 15 | 17 | -2 | 2 | 4 | |
| 3 | 20 | 22 | -2 | 2 | 4 | |
| 4 | 25 | 26 | -1 | 1 | 1 | |
| 5 | 30 | 34 | -4 | 4 | 16 | |
| $\sum$ | | | | 12 | 34 | |
| **Mean Residual Error** | | | | MAE : **2.4** | MSE : **6.8** | RMSE : **2.61** |

2. In case of MSE the variation of MSE of 6.8 from minimum value 1 is 580%, from max value 16 is 57.5% and from error square value 4 is 70%, and from error square value 9 is 24.44%.

3. In case of RMSE the variation of RMSE 2.61 from the minimum error value 1 is 161% and from maximum error value 4 is 34.75%. The variation from other values like from value 3 is 13% and from error value 2 is 30.5%.

4. The comparison clearly shows leaving aside the outliers, that is error value 1 and 4, the result of MAE are the best, 20%, the lowest of three in majority of cases.

Also for the high value of absolute error that is 4 in the above example the variation of 40% in MAE is very reasonable, that is high value of error doesn't affect much the performance of the regression model using MAE. Therefore, if our model is not having many outliers, and if we can ignore them, then MAE is much better option for our regression model and that is the reason we have chosen MAE as loss function for our regression model (DataCourses, 2020) (DataQuest, 2018).

A comparison between MAE/MSE/RMSE is tabulated in the below table.

Table 9. MAE/MSE/RMSE comparison

| Sno | MAE | MSE | RMSE |
|---|---|---|---|
| 1. | Being absolute error (-ve or +ve), always +ve for calculations. | Consider both positive or negative values. | Consider both positive or negative values. |
| 2. | Bias is less when the error value is high. Thus not very suitable for large value errors. | Better suited than MAE for high error values due to higher bias. | Also better than MAE for high value of errors. |
| 3. | Though not suitable to deal with high value errors, MAE doesn't hamper the output. | The high value errors hamper the output to a great extent. | RMSE also hampers the output greatly for high error values. |

## 13.4 GuavaNet construction

A hierarchical ensemble of deep learning neural network is constructed using two deep convolutional neural network (DCNN) architectures one for object detection and localization and the other one to predict acceptability index. For the purpose of object detection pre-trained YOLOv4 on COCO dataset Lin et al., 2015 is used. Head of YOLOv4 as an output predicts a bounding box $b_x, b_y, b_w, b_h$ where $b_x$ and $b_y$ are center points of the bounding box that is the ground truth. Let us say input to YOLOv4 is $Y$. The output will be $boundingBox(Y, b_x, b_y, b_w, b_h)$. The resultant output image is cropped as per the bounding box around the image as $I_{c_r} = crop(Y, b_x, b_y, b_w, b_h)$. This cropped image $I_{c_r}$ is then fed into the trained VGG-16 + Custom FCN architecture that has been fine-tuned after being trained as a feature extractor. As VGG-16 + Custom FCN as feature extractor performed better than ResNet-18 + Custom FCN and ResNet-50 + Custom FCN as feature extractors therefore this VGG-16 + Custom FCN model is further fine-tuned and is used to predict the acceptability index. Construction of such a hierarchical ensemble of deep convolutional neural network (DCNN) involves training YOLOv4 and training VGG-16 + Custom FCN as a feature extractor. After, the training is complete for the feature extractor, its layers were unfrozen and then again, the whole architecture was trained for another 15 epochs. This process involved combining the transfer learning approach of using CNN as feature extractor and fine-tuning the feature extractor by unfreezing all the layers.

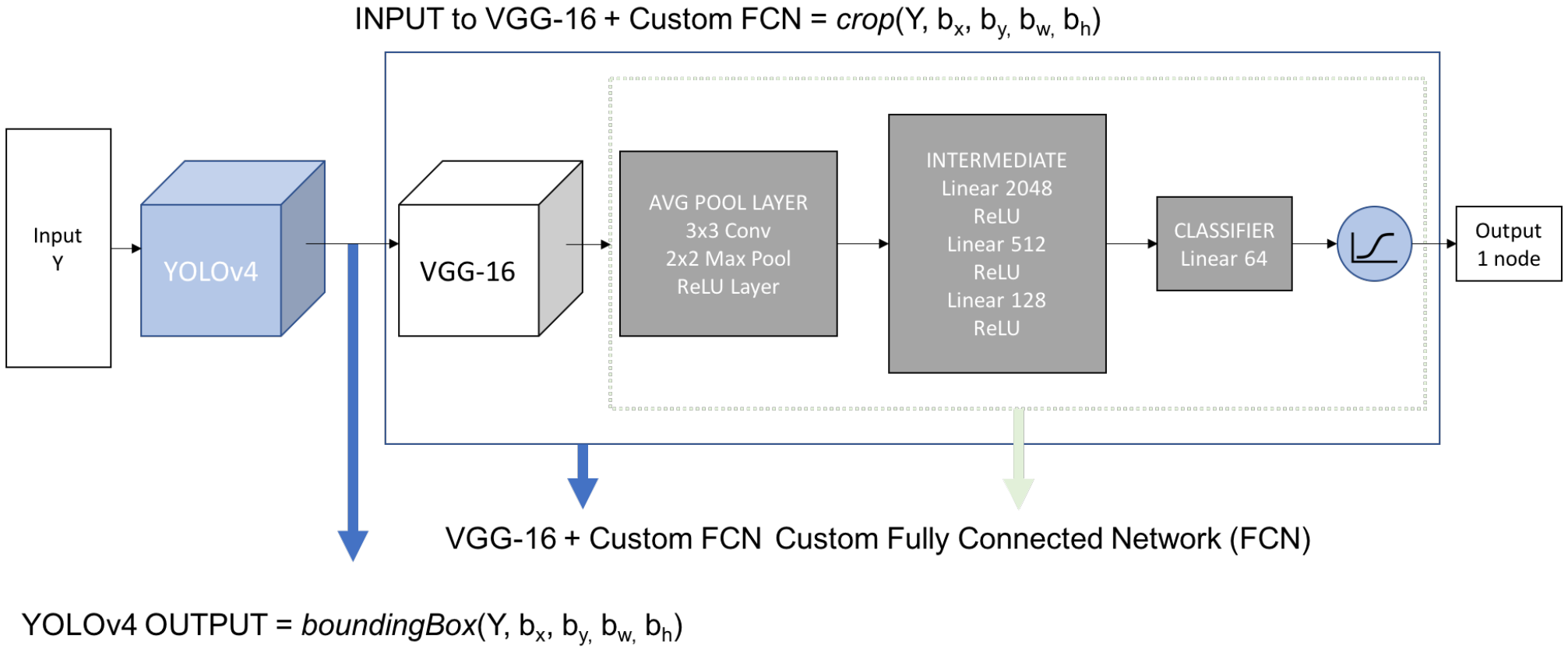

Figure 18. GuavaNet: A Hierarchical ensemble architecture to predict acceptability index

# 14 Results and discussion

## 14.1 Object detection and localization

YOLOv4 pre-trained on COCO dataset (Lin et al., 2015) has been used for the purpose of object detection and localization. YOLOv4 was trained on 452 images and validated on 124 images. It is able to generalize guavas with mean average precision of **mAP@0.5** of **97.56%**.

### 14.1.1 YOLOv4 report

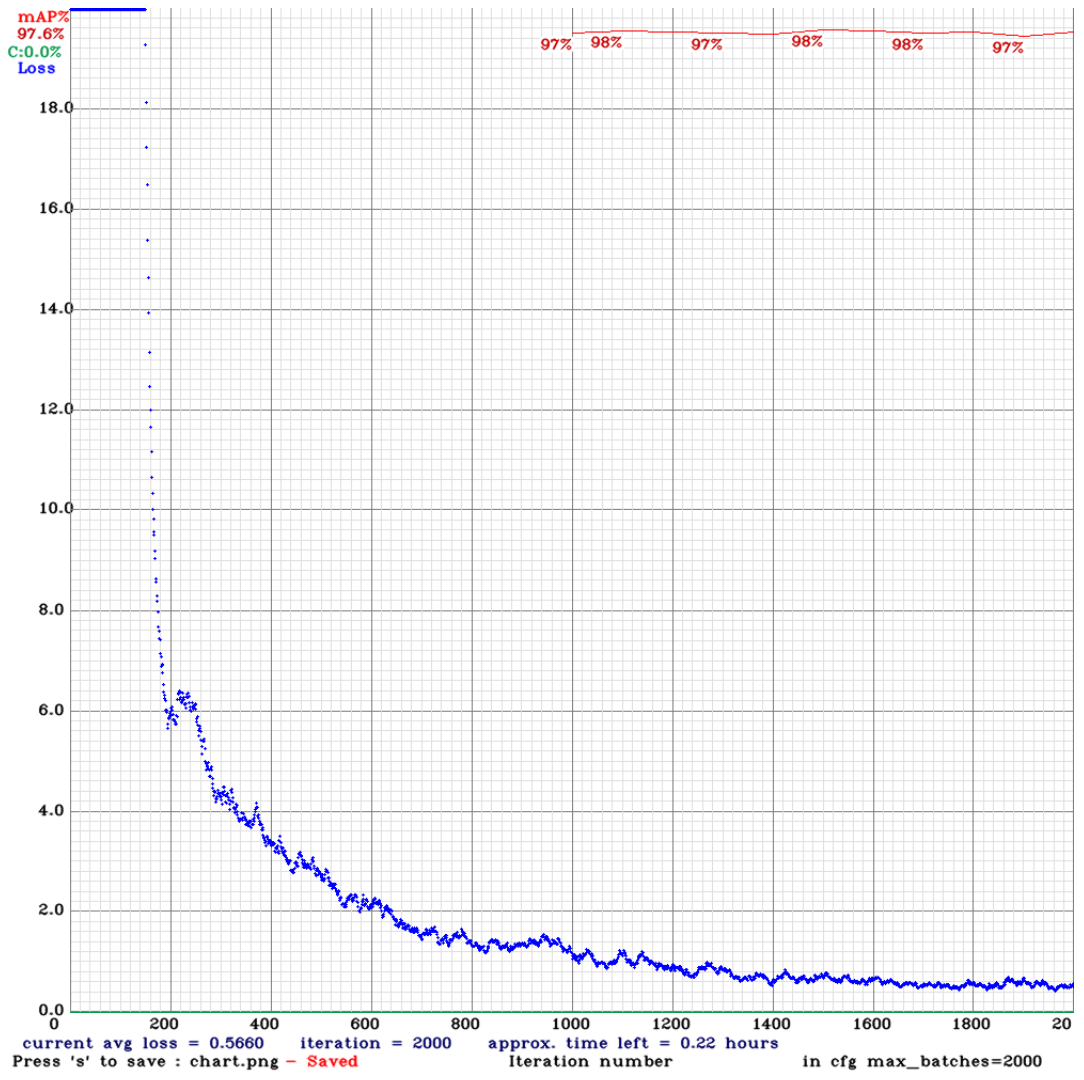

Figure 19. YOLOv4 Report

### 14.1.2 YOLOv4 results

Table 10. YOLOv4 - Evaluation metrics

| TP | FP | FN | Average IoU | mAP@0.5 | Precision | Recall | F1 Score |
|---|---|---|---|---|---|---|---|
| 229 | 17 | 5 | 83.41% | 97.56% | 0.93 | 0.98 | 0.95 |

## 14.2 Predicting the acceptability index

To predict the regressed variable ‘Acceptability Index’ to carry out sensory evaluation, pre-trained VGG-16, ResNet-18 and ResNet-50 networks on ImageNet (Deng et al., 2009) were analyzed. These were first used as a fixed feature extractor with a custom Fully Connected Neural Network (FCN) replacing the last layer. All the other layers except the last layer weights were frozen and were set to being not trainable. Only the last layer was then trained on our custom dataset consisting of 299 images - where 60 images were used for validation and 239 for training. For each one the training ran until a suitable mean absolute error (MAE) was found or the test loss stopped changing. After 25 epochs it was found VGG-16 had a MAE of **0.29**, whereas for ResNet-18 and ResNet-50 after 50 epochs MAE was **0.35** and **0.33** respectively. VGG-16 was able to generalize the dataset better. As VGG-16 performed best, this was selected for the purpose of predicting the acceptability index. This trained model’s feature extractor was then unfrozen and the whole model was then trained on this dataset. We combined the transfer learning approaches of using pre-trained network - fine-tuning and as a fixed feature extractor. The training happened for 15 epochs. After 15 epochs the change in test loss came to a halt. The MAE improved by **10.34%** from **0.29** initially to **0.26** and the model was able to generalize better.

### 14.2.1 Results - Convolution neural network as a fixed feature extractor

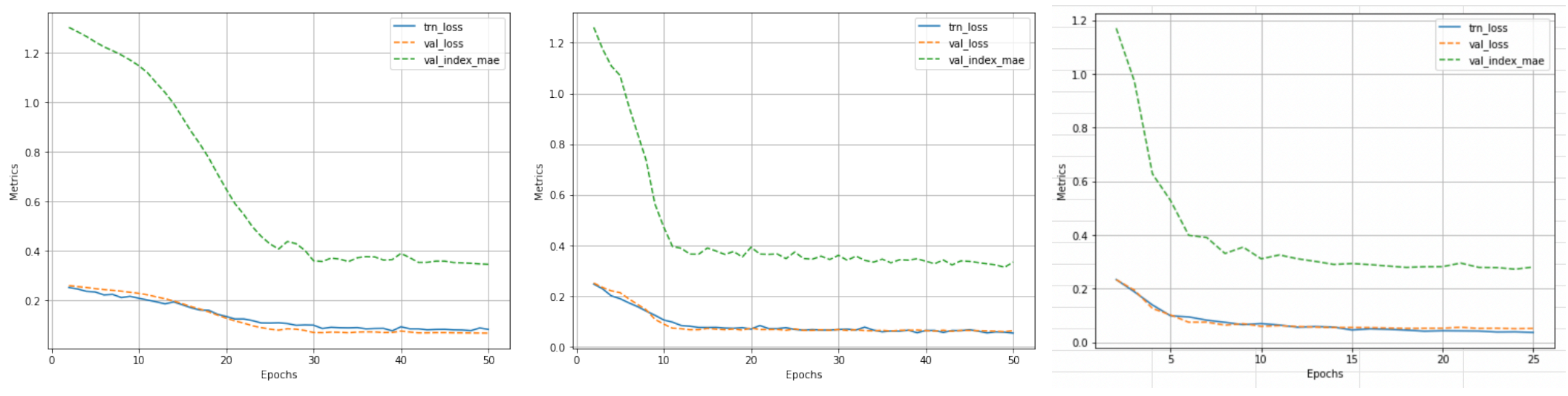

(a) (b) (c)

Figure 20. (a) ResNet-18 + Custom FCN report (b) ResNet-50 + Custom FCN report (c) VGG-16 + Custom FCN report

Table 11. Convolutional neural network as a fixed feature extractor - Evaluation metrics

| Network | Epochs | Train loss | Test loss | Best test loss | MAE |
|---|---|---|---|---|---|
| **VGG-16 + Custom FCN** | 25 | **0.036** | **0.053** | **0.051** | **0.29** |
| ResNet-18 + Custom FCN | 50 | 0.075 | 0.069 | 0.068 | 0.35 |
| ResNet-50 + Custom FCN | 50 | 0.065 | 0.064 | 0.060 | 0.33 |

#### 14.2.2 Results - Fine-tuning trained fixed feature extractor VGG-16 model

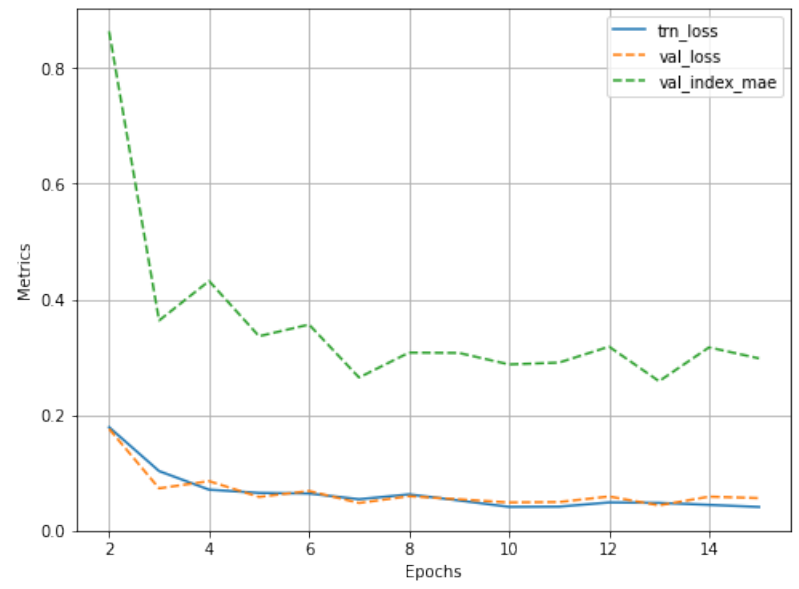

Figure 21. VGG-16 + Custom FCN report

Table 12. Fine-tuning CNN as fixed feature extractor - Evaluation metrics

| Network | Epochs | Train loss | Test loss | Best test loss | MAE |
|---|---|---|---|---|---|
| VGG-16 + Custom FCN | 15 | 0.041 | 0.045 | 0.044 | **0.26** |

#### 14.2.3 Performance evaluation

Initially, when these three (VGG-16 + Custom FCN, ResNet-18 + Custom FCN, ResNet-50 + Custom FCN) models used as feature extractors were trained, then the sensory evaluation was done manually testing on hold-out dataset unseen by these architectures, it was found that all the three models predicted almost similar acceptability index for the guavas with lower acceptability index, but for the guavas with higher acceptability index ResNet-18 + Custom FCN and ResNet-50 + Custom FCN predictions of acceptability index were much lower compared to the ground. However, VGG-16 + Custom FCN model predicted acceptability index was very near to the ground truth. This can be seen in the below figures containing the ground truth and prediction values for acceptability index. VGG-16 is able to transfer learn features of guavas at differently acceptability index better than ResNet-18 + Custom FCN and ResNet-50 + Custom FCN when dealing with small datasets such as ours.

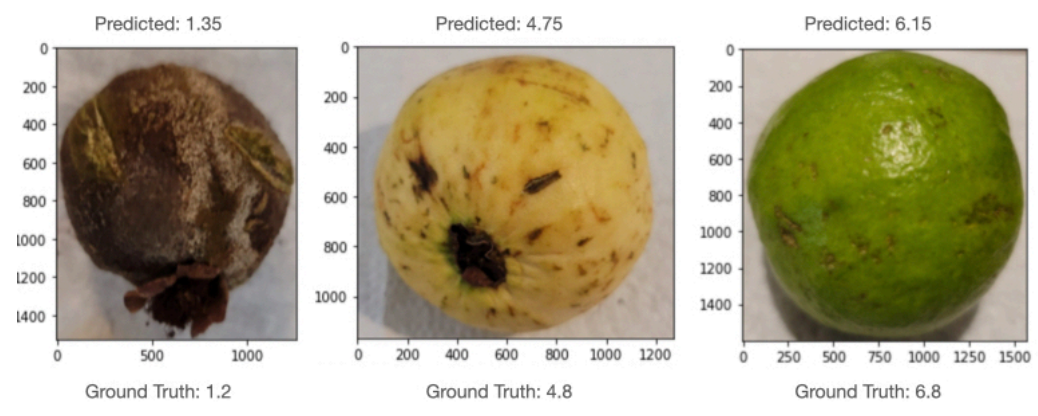

(a)

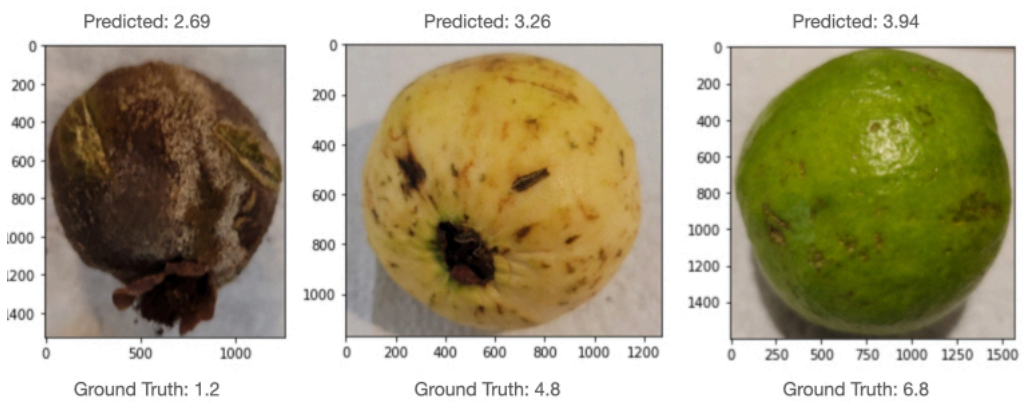

(b)

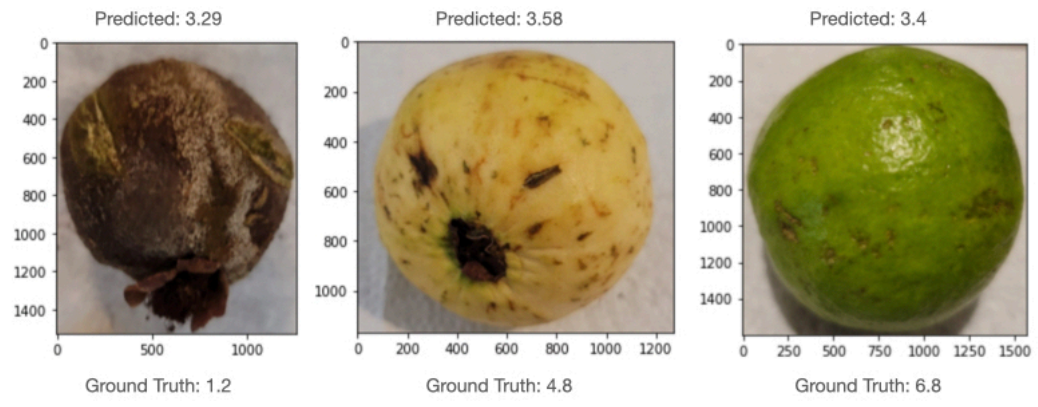

(c)

Figure 22. (a) Predictions from VGG-16 + Custom FCN as Feature Extractor (b) Predictions from ResNet-18 + Custom FCN as Feature Extractor (c) Predictions from ResNet-50 + Custom FCN as Feature Extractor

ResNet-18 + Custom FCN, ResNet-50 + Custom FCN as feature extractors were discarded, and VGG-16 was trained combining the transfer learning approach of fine-tuning the fixed feature extractor model. This further improved the MAE and is able to predict very near to the ground truth as can be seen in the below figure.

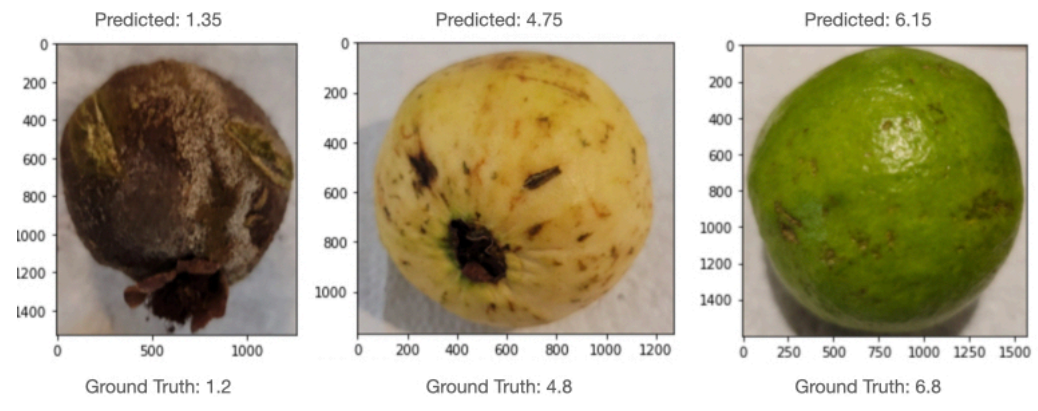

(a)

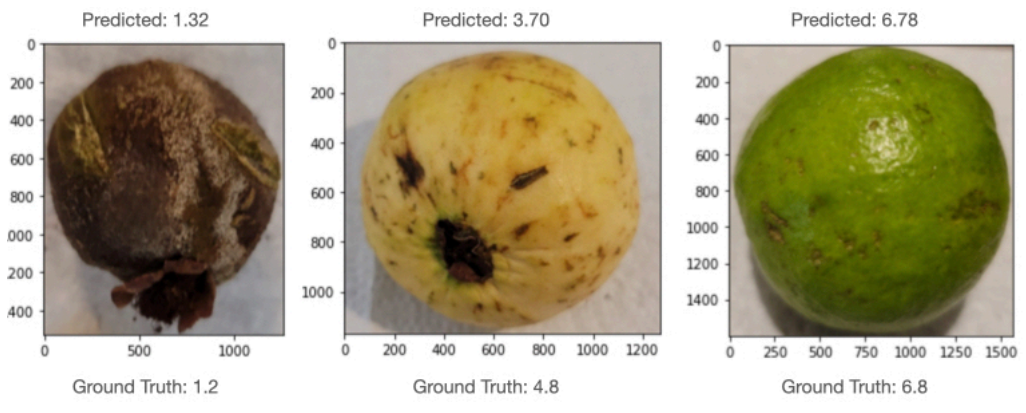

(b)

Figure 23. (a) Predictions from VGG-16 + Custom FCN as Feature Extractor (b) Predictions from Fine-tuned VGG-16 + Custom FCN as Feature Extractor

## 14.3 GuavaNet Output

The above image is taken in real-time through computer camera. These guavas and apples are unseen by our model and have not been in the validation or training set.

As can be seen from the above figures, guavas are correctly classified by YOLOv4 architecture. The predicted values of acceptability index for guavas in figure (b) and (d) are **3.12**, **7.63**, **7.14**. Their corresponding ground truth values are **1.4**, **7.2** and **7.14** and have been assigned a label for sensory evaluation on the basis of acceptability index as "Dislike extremely", "Like extremely" and "Like extremely" respectively. The results are very much in line with their ground truth values, which suggests that GuavaNet is successfully able to carry out sensory evaluation and predict the degree of acceptability for guava by a consumer.

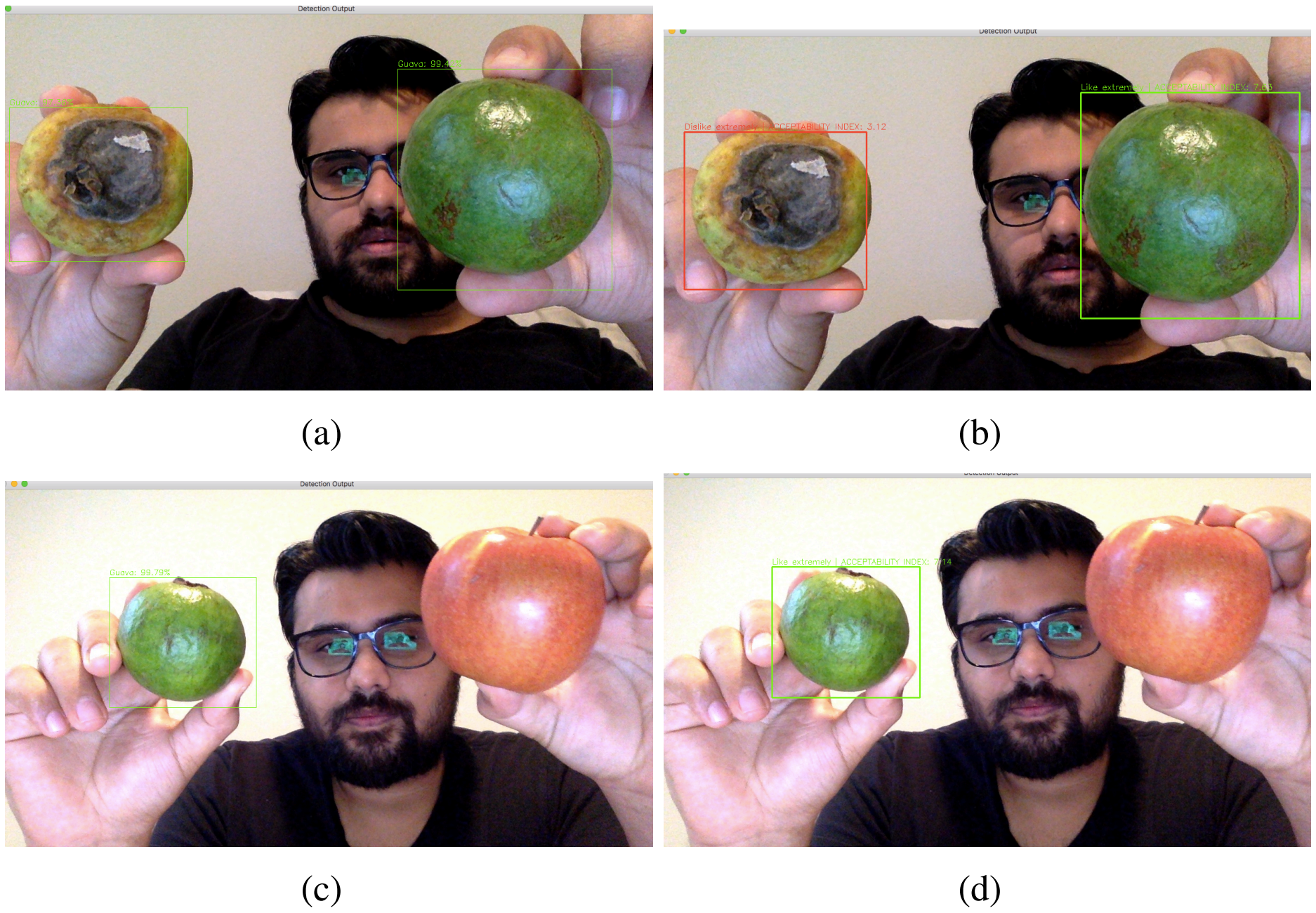

Figure 24. (a) Object Localization - 2 Guavas (b) Predicted acceptability index - 2 Guavas (c) Object Localization - 1 apple, 1 guava (d) Predicted acceptability index - 1 Guava

# 15 Conclusion

In this thesis we propose a novel hierarchical deep learning architecture based on YOLOv4 and VGG-16 + Custom FCN that is able to predict the acceptability index for a guava by a consumer.

Various pre-trained architectures like VGG-16 + Custom FCN, ResNet-18 (Custom ) and Resnet-50 were trained on the custom dataset to predict the acceptability index. VGG-16 + Custom FCN was able to generalize the dataset best and model was further improved after combining the transfer learning techniques of using CNN as feature extractor and fine-tuning. This improved the MAE by **10.34%** from **0.29** in **0.26**. Furthermore, for object detection and localization YOLOv4 pre-trained on COCO dataset (Lin et al., 2015) performed really well with an overall mAP@0.5 of **97.56%**, and is able to decipher between guava and other fruits with a high degree of accuracy.

A hierarchical ensemble of these two trained models is constructed to provide final output, that is detection and localization of guava and then predict the degree of acceptability by a consumer for the guavas detected in the picture. The predicted bounding box as an output from the YOLOv4 is used to crop the input image and then that cropped image is fed into the CNN to predict the acceptability index. And end-to-end pipeline has been created to be able to do this simultaneously.

The overall architecture proposed in this research work is in reality very close to the thought process of the common people and the predictions are fairly good.
By using the nine-point hedonic scale and feature weighting based on texture, color and shape to label and classify the images, the system is able to answer the long time pending question of, "Whether these guavas will be picked up from the shelf or will be discarded by the consumer?". The overall weighting of shape, color and texture is based on an anonymous survey that was carried out to find out between shape, color and texture what is more important. It was concluded from the survey that when looking to buy fruits, vegetables or any such items people are looking for color and texture over shape. Further, the literature review also talks about the importance of shape and texture over color. Based on this weighted scheme we labelled the score for each of the 299 guavas, where the ratio of weights is 2:1:2 for color:shape:texture respectively.

The model has developed the intelligence of food technologists, harvesters, food experts and commoners observations for selecting the best and healthy guavas that are likely to be accepted and picked up by a consumer in a store. The system in a way is an improved alternative for the way the people would inspect a guava through sensory evaluations via evoking, measuring, analyzing and comprehending the reaction to those characteristics of guavas as they are perceived by their eyes, smell, taste and touch to choose fresh guavas at a market place. The system can also be helpful to the fruit, vegetable sellers and supermarkets. The simplicity of the developed system has the potential to revolutionize the food and vegetable selling markets, and enable them to stock things that are likely to be accepted by the buyers. Thus, reducing waste and providing fresh products.

# 16 Limitations

The hierarchical deep neural network architecture is limited to guavas that are not cut or sliced. It can predict degree of acceptability for guava by a consumer for whole guavas. The model is limited to predict the degree of acceptability of those guavas that have been captured over time in ambient temperature and conditions.

The architecture is not trained on dehydrated or diseased guava, hence might not be able to predict the acceptability index with a very high degree of accuracy. Since, pictures are two dimensional in nature, the system is not be able to capture the texture, shape and color from all the possible angles. For example: consider a guava whose one side is diseased, and the other side is free of any issues. If the image consists of only the side, that has no issues, the prediction for acceptability index would be positive. On the other hand, if the side shown in the picture or on the camera is the diseased one, the predictions will be exactly opposite.

The current system in place is not so robust to be able capture 360 degree details in one go and therefore is limited to what is being provided as an input to the model. Object detection and localization using YOLOv4 though performs really well, however sometimes it predicts images of green avocados, lemons and pears as a green guava, though the percentage of such detections is very low. It is happening due to their close resemblance to guavas in terms of texture, shape and color. This anomaly can be rectified by adding some more relevant data in the training set.

# 17 Future Work

In the future this can be extended to other whole, cut and sliced guavas and other fruits and vegetables. Object localization and detection could be further improved by adding in more relevant samples in the dataset, that would improve the classification accuracy and reduce the misclassification.